\documentclass{article}

\usepackage[main, final]{neurips_2026}

\usepackage[utf8]{inputenc} 
\usepackage[T1]{fontenc}    
\usepackage{hyperref}       
\usepackage{url}            
\usepackage{booktabs}       
\usepackage{amsfonts}       
\usepackage{nicefrac}       
\usepackage{microtype}      
\usepackage{xcolor}         
\usepackage{amsmath, amssymb, amssymb, cancel, graphicx, wrapfig}
\usepackage{tocloft, enumitem, amsthm}
\newtheorem{theorem}{Theorem}
\newtheorem{defn}{Definition}
\newtheorem{lemma}[theorem]{Lemma}
\newtheorem{claim}[theorem]{Claim}
\newtheorem{proposition}[theorem]{Proposition}
\newtheorem{coro}[theorem]{Corollary}

\newtheorem{remark}[theorem]{Remark}
\newtheorem{example}[theorem]{Example}
\newcommand{\bs}{\boldsymbol}
\newcommand{\M}{\mathcal{M}}

\newcommand{\y}{\mathbf{y}}
\newcommand{\Y}{\mathbf{Y}}
\newcommand{\x}{\mathbf{x}}
\newcommand{\z}{\mathbf{z}}
\newcommand{\X}{\mathbf{X}}
\allowdisplaybreaks

\title{On Inherent Privacy of Posterior Sampling: A Unified R\'enyi-Divergence Framework}

\author{%
  Ruyu Zhou \\ Department of Computer Science and Statistics \\ University of Rhode Island\\
  \texttt{ruyu.zhou@uri.edu}
  \And
  Fang Liu \\ Department of ACMS \\ University of Notre Dame\\
  \texttt{fliu2@nd.edu}
}

\begin{document}

\maketitle

\begin{abstract}
Differential Privacy (DP) provides a rigorous framework for quantifying privacy guarantees. While most methods achieve DP by injecting calibrated random noise, the intrinsic randomness of certain procedures such as sampling can yield DP guarantees ``for free''. In this work, we develop a unified R\'enyi divergence framework to characterize the inherent DP guarantees of sampling from a Bayesian posterior distribution. We show that the privacy loss incurred by One-Posterior-Sampling (OPS) is governed by posterior exponential moments of the single-record log-likelihood ratio, and establish explicit DP guarantees under uniformly bounded, sub-Gaussian, or sub-exponential tail regimes. We apply the framework to representative Bayesian models, including categorical likelihoods with arbitrary priors, Gaussian-Gaussian conjugate pair, generalized linear models from the exponential family and linear regression with Gaussian likelihood, both with Gaussian priors on the regression coefficients. Our results elucidate sample size, model structure, neighboring relations in DP (bounded or unbounded), and how prior if applicable, jointly determine the inherent DP guarantees of OPS. Furthermore, our results recover the existing DP guarantees for OPS as special cases -- often with tighter privacy loss bounds -- and substantially broaden the class of Bayesian models for which the inherent DP guarantees of OPS can be rigorously characterized.
\end{abstract}

\section{Introduction}\label{sec:intro}

\paragraph{Background and Motivation.} Differential Privacy (DP) has emerged as the gold standard for providing rigorous privacy guarantees when releasing information from sensitive databases since its introduction in 2006 \citep{dwork2006our, dwork2006calibrating}. By bounding the change in a randomized algorithm's output distribution induced by any single individual's data, classical DP notion ensures that no individual can exert a disproportionate influence on the outcome. However, the worst-case nature of pure DP can be overly conservative for practical applications. This has motivated several influential relaxations that enable more refined privacy accounting, such as R\'enyi DP \citep{mironov2017renyi}, zero-concentrated DP \citep{bun2016concentrated}, Gaussian DP \citep{dong2022gaussian}, among others.

While a target DP level is typically achieved by explicitly injecting calibrated noise -- a paradigm underpinning most DP methods --  
many learning and analytical procedures are intrinsically stochastic. 
Accounting for the privacy protection provided by such inherent stochasticity would potentially help meet DP requirements while reducing, or even eliminating, the need for external perturbation, thereby better preserving model utility. Bayesian posterior sampling naturally falls into this category and offers a principled setting for exploiting this form of ``free'' privacy. In particular, a single posterior draw $\bs\theta^*$ defines a complete randomized predictor or synthetic-data generator; any number of predictions or synthetic records computed from $\bs\theta^*$ inherit its privacy guarantee by post-processing, without composition.

Despite growing interest in Bayesian methods under DP, existing guarantees for One-Posterior-Sampling (OPS) remain limited (see Related Work): many require uniformly bounded likelihood ratios, bounded sample or parameter spaces, or modified (pseudo-)posterior mechanisms; or yield loose privacy loss bounds. Recent work on posterior sampling under contamination and pseudo-posterior mechanisms shows that this problem remains active and practically relevant. We revisit OPS and develop sharper DP guarantees under weaker and more broadly applicable conditions.

%
\paragraph{Our Contributions.} %
\begin{itemize}[leftmargin = *, itemsep = 0pt]
    
    \item We establish an \textit{exact characterization} of the R\'enyi divergence between the posteriors formulated over two neighboring datasets, showing that the privacy loss of OPS is fully determined by two specific \textit{exponential moments} of the single-record log-likelihood ratio.
    \item We derive explicit DP guarantees for OPS under three log-likelihood ratio regimes: \textit{uniformly bounded}, \textit{sub-Gaussian}, and \textit{sub-exponential}. These results recover existing OPS DP guarantees as special cases and are compatible with various DP notions.
    \item We apply the framework to a broader class of Bayesian models than previously studied. The resulting bounds show how the prior, sample size, model structure, and neighboring relation jointly determine the inherent DP guarantees of OPS; in particular, under unbounded neighboring, the two directions of the R\'enyi divergence can differ even in their range of finite orders.
\end{itemize}

%
\paragraph{Related Work.}\cite{chaudhuri2006random} initiated the idea that random sampling itself can provide DP guarantees. Building on this insight, \citet{li2012sampling} showed that random sampling before $k$-anonymization \citep{sweeney2002k} yields approximate DP. Subsequent work has mainly followed two directions: 1) studying DP guarantee induced by algorithmic randomness, including uncertainty and noise in control systems \citep{giraldo2017security} and stochastic generative models such as GANs \citep{lin2021privacy}; 2) studying the privacy amplification effect, where subsampling randomness amplifies the privacy guarantees of a base mechanism. Foundational results include \citet{balle2018privacy} and survey by \citet{steinke2022composition}, with recent extensions to multistage sampling schemes \citep{zhao2024enhancing}.

Privacy guarantees from posterior sampling in the Bayesian framework have also garnered significant research attention, especially for OPS. Early work \cite{mir2013differential,dimitrakakis2014robust, dimitrakakis2017differential, wang2015privacy} established pure DP guarantees for general OPS by assuming uniformly bounded log-likelihood ratio between neighboring datasets -- an assumption that is often too restrictive. The resulting DP guarantees scale with the assumed uniform bound and can be conservative due to triangle-inequality-based arguments.
Subsequent work examined OPS in specific Bayesian models, including discrete Bayesian networks \citep{zhang2016differential} and exponential family models with conjugate priors \citep{foulds2016theory}. The work most closely related to ours is \citet{geumlek2017renyi}, which studies OPS under R\'enyi DP \citep{mironov2017renyi}, but is limited to exponential family and Bayesian logistic regression. In contrast, our work develops a unified R\'enyi DP characterization for OPS that applies beyond these settings, and recover their results as special cases.

Beyond OPS, intrinsic DP guarantees have also been studied for related posterior-sampling-based inference procedures, including collapsed Gibbs sampling for latent Dirichlet allocation \citep{zhao2020latent, blei2003latent} and stochastic gradient MCMC \citep{li2019connecting}. Related directions include combining OPS with $\beta$ divergences to obtain bounded sensitivity \citep{jewson2023differentially}, per-instance DP analyses for linear regression \citep{minami2016differential, wang2019per}, and posterior sampling under contamination models without boundedness assumptions \citep{hu2024privacy}. Our work differs by focusing on unmodified OPS and deriving sharper privacy-loss guarantees under weaker assumptions.

%
\section{Preliminaries}\label{sec:prelim}
\begin{defn}[Differential Privacy (DP)] \citep{dwork2006our}
    A randomized mechanism $\M$ is said to satisfy $(\varepsilon,\delta)$-DP if for all pairs of neighboring datasets $(\Y,\Y')$ differing by one record, denoted by  $d(\Y,\Y')=1$, and for all subsets $\mathcal{S}\subseteq$ image$(\M)$, it holds that
    \begin{equation}\label{eqn:DP}
    \Pr(\mathcal{M}(\Y)\in \mathcal{S}) \leq e^{\varepsilon} \Pr(\mathcal{M}(\Y')\in \mathcal{S})+\delta.
    \end{equation}
\end{defn}\label{defn:dp}
$d(\Y,\Y')\!=\!1$ refers to either \textit{bounded} neighboring \citep{Kifer2011}, where $\Y$ becomes  $\Y'$ by substituting exactly one record, or \textit{unbounded} neighboring, where $\Y$ becomes $\Y'$ by adding or removing a single record. $\varepsilon>0$ and $\delta\ge0$ in Eq.~\eqref{eqn:DP} are privacy loss parameters, with smaller values indicating stronger privacy protection.  When $\delta=0$, $(\varepsilon,\delta)$-DP becomes pure  $\varepsilon$-DP. 

DP can also be characterized using a privacy loss random variable that quantifies the privacy loss incurred by releasing a particular output (Definition \ref{def:loss}).
\begin{defn}[Privacy Loss Random Variable] \citep{bun2016concentrated, sommer2018privacy}\label{def:loss}
For a randomized mechanism $\M$, let $o\in \text{image}(\M)$ be any potential output of $\M$ and let $\X_0, \X_1 \in \mathcal{X}$ be two inputs. The privacy loss random variable $\mathcal{L}$ associated with releasing $o$ is defined as
\begin{equation}\label{eqn:PrivLoss}
    \mathcal{L}_{\M(\X_0)\|\M(\X_1)}(o)\!=\log\!\left(\Pr(\M(\X_0) = o)/\Pr(\M(\X_1) = o)\right).
\end{equation}
\end{defn}
For any two neighboring datasets $\Y$ and $\Y'$, a mechanism $\M$ satisfies pure $\varepsilon$-DP if and only if $\Pr(\mathcal{L}_{\M(\Y)\|\M(\Y')}\!>\!\varepsilon)\!=\!0$, whereas $(\varepsilon, \delta)$-DP is equivalent, up to a small loss in parameters, to the requirement that $\Pr(\mathcal{L}_{\M(\Y)\|\M(\Y')}\!>\! \varepsilon)\!\le\! \delta$. A relaxation of DP based on the R\'enyi divergence is: 

\begin{defn}[R\'enyi Differential Privacy (RDP)] \citep{mironov2017renyi}\label{def:rdp}
A randomized mechanism $\M$ is said to have $(\alpha, \varepsilon)$-RDP if for all pairs of neighboring datasets $\Y,\Y'$, it holds that
\begin{equation}
    D_{\alpha}(\M(\Y) \|\M(\Y'))\le \varepsilon,
\end{equation}
where $D_{\alpha}(P\|Q)$ is the R\'enyi divergence \citep{renyi1961measures} of order $\alpha > 1$ between distributions $P, Q$:
\begin{equation}\label{eqn:renyi_div}
    D_{\alpha}(P\|Q) =(\alpha-1)^{-1}\log \mathbb{E}_{x\sim Q}\left(P/Q\right)^\alpha = (\alpha-1)^{-1}\log \mathbb{E}_{x\sim P}\left(P/Q\right)^{\alpha-1}.
\end{equation}
\end{defn}
Beyond defining RDP, R\'enyi divergence is also linked to several popular DP notions, such as $(\varepsilon, \delta)$-DP, $\rho$ zero-concentrated DP (zCDP), and Gaussian DP (GDP), in the sense that they can be expressed via R\'enyi divergence. The following proposition summarizes these relationships.
\begin{proposition}[Relations between R\'enyi divergence and DP notions]\label{prop:DPconversion}
Let $\M$ be a randomized mechanism, and $\Y,\Y'$ be any pair of neighboring datasets satisfying $d(\Y, \Y') = 1$. Then,
\begin{itemize}[leftmargin =*, itemsep=0pt]
    \item $\M$ is of $\varepsilon$-DP if and only if $D_{\infty}(\M(\Y)\|\M(\Y'))\le \varepsilon$ \citep{mironov2017renyi}\footnote{$D_{\infty}(P\|Q) =\sup_{x\in\text{support }P}\log(P/Q)$ is the max divergence between $P$ and $Q$ when $\alpha=\infty$.}.
    \item $\M$ is of $\rho$-zCDP if and only if $\;\forall \alpha \!\in\! (1, \infty)$, $D_{\alpha}(\M(\Y) \|\M(\Y'))\!\le\!\rho\alpha$ \citep{bun2016concentrated}.
    \item If $\M$ is of $(\alpha, \varepsilon)$-RDP, it is also of $(\varepsilon\!+(\alpha\!-\!1)^{-1}\log 1/\delta, \delta)$-DP for $\forall \;0 < \!\delta\! < 1$ \citep{mironov2017renyi}.
    \item If $\M$ is of $\rho$-zCDP, it is also of
    $\sqrt{2\rho}$-GDP. If $\M$ is of $\mu$-GDP,
    it is also of $\mu^2/2$-zCDP \citep{dong2022gaussian}.
\end{itemize}
\end{proposition}

\section{A Unified R\'enyi Framework for Privacy Loss from Posterior Sampling}

We develop a unified R\'enyi-divergence framework for characterizing the inherent privacy loss of posterior sampling, based on a decomposition of the privacy loss random variable between the posterior distributions formulated over two neighboring datasets.


\subsection{Privacy Loss Decomposition for Neighboring Posteriors}

Let $\y \in \mathcal{Y}$ denote a single observation, its distribution is $p(\y|\bs\theta)$ with parameter $\bs\theta \in \bs\Theta$, and  $\pi(\bs\theta)$ is the prior distribution. Given a dataset $\Y = \{\y_i\}_{i=1}^n$ and assuming independence across the $n$ observations,  the posterior distribution of $\bs\theta$ is
{
\begin{align}
\pi(\bs\theta|\Y) = \frac{\pi(\bs\theta)\prod_{i=1}^{n}p(\y_i|\bs\theta)}{\int \prod_{i=1}^{n}p(\y_i|\bs\theta)\pi(\bs\theta) d\bs\theta}
= \frac{\pi(\bs\theta)\ell(\bs\theta;\Y)}{p(\Y)},\label{eqn:posterior}
\end{align}}

where $\ell(\bs\theta;\Y) \triangleq \prod_{i=1}^n p(\y_i|\bs\theta)$ denotes the likelihood function and $p(\Y) \!=\! \int \ell(\bs\theta;\Y)\pi(\bs\theta) d\bs\theta$ is the normalizing constant. Consider any pair of neighboring datasets  $d(\Y,\Y')=1$ with corresponding posterior distributions, $\pi(\bs\theta|\Y)$ and $ \pi(\bs\theta|\Y')$, which we refer to as \textbf{neighboring posteriors}. 
The  privacy loss random variable $c(\bs\theta)$ associated with releasing one posterior sample $\bs\theta$ is
{
\begin{align}
c(\bs\theta)
=\log\left(\frac{\pi(\bs\theta|\Y)}{\pi(\bs\theta|\Y')}\right)
&=\log\left(\frac{\cancel{\pi(\bs\theta)}\ell(\bs\theta;\Y)}{p(\Y)}\cdot \frac{p(\Y')}{\cancel{\pi(\bs\theta)}\ell(\bs\theta;\Y')}\right) \text{ per Eq.~\eqref{eqn:posterior}}\notag\\
&=\underbrace{\log\left(\frac{\ell(\bs\theta;\Y)}{\ell(\bs\theta;\Y')}\right)}_{\text{$\triangleq\; c_1(\bs\theta)$}}+\underbrace{
\log\left(\frac{p(\Y')}{p(\Y)}\right)}_{\text{$\triangleq\; c_2$}}=c_1(\bs\theta) + c_2.
\label{eqn:decomp}
\end{align}}\vspace{-9pt}

Here, $c_1(\bs\theta)$ is the pointwise log-likelihood ratio evaluated at $\bs\theta$, while $c_2$ represents the marginal log-likelihood ratio obtained after integrating out $\bs\theta$. $c_1(\bs\theta)$ and $c_2$ are related; in particular, $c_2$ can be expressed as the logarithm of an exponential moment of $c_1(\bs\theta)$ under the posterior $\pi(\bs\theta|\Y)$. This relation is formalized in Lemma \ref{lemma:c2_in_c1}, with the proof is provided in Appendix~\ref{ape:c2_in_c1}. 
\begin{lemma}[Marginal-Pointwise identity]\label{lemma:c2_in_c1}
     $c_2 = \log\mathbb{E}_{\bs\theta \sim \pi(\bs\theta|\Y)}\left[\exp(-c_1(\bs\theta))\right].$
     Specifically, if there exist finite constants $a, b \in \mathbb R$ with $a<b$ such that $c_1(\bs\theta) \in [a, b]$ for $\forall \bs\theta\in\bs\Theta$, then $c_2\in [-b, -a]$.
\end{lemma}

\subsection{R\'enyi Divergence Characterization}
Building on the decomposition in Eq.~\eqref{eqn:decomp}, we establish an exact expression for the R\'enyi divergence of any order $\alpha\!>\!1$ between neighboring posteriors in Theorem \ref{thm:D_alpha_general}. The proof is provided in Appendix~\ref{ape:D_alpha_general}. 
\begin{theorem}[Exact  characterization of R\'enyi divergence between neighboring posteriors]\label{thm:D_alpha_general}
Let $P=\pi(\bs\theta|\Y)$ and $Q=\pi(\bs\theta|\Y')$ be a pair of neighboring posteriors. With $c_1(\bs\theta)$ and $c_2$ defined  in Eq.~\eqref{eqn:decomp}, the R\'enyi divergence of order $\alpha>1$ between $P$ and $Q$ is 
{
\begin{align}
    D_{\alpha}(P\|Q) &=c_2 +(\alpha-1)^{-1}\log \mathbb{E}_{\bs\theta\sim P}\left[e^{(\alpha-1)c_1(\bs\theta)}\right]\label{eqn:D_alpha_general_c2}\\
    &= \log\Big[\!\left(\mathbb{E}_{\bs\theta \sim P}\left[e^{-c_1(\bs\theta)}\right]\right)\cdot\left(\mathbb{E}_{\bs\theta\sim P}\left[e^{(\alpha-1)c_1(\bs\theta)}\right]\right)^{\frac{1}{\alpha-1}}\Big] \text{ per Lemma~\ref{lemma:c2_in_c1}}.\label{eqn:D_alpha_general}
\end{align}}
\end{theorem}
Eq.~\eqref{eqn:D_alpha_general} suggests that the R\'enyi divergence of order $\alpha$ -- $D_{\alpha}(P\|Q)$ --  between neighboring posteriors is fully determined by exponential moments $\mathbb E_{\bs\theta \sim P} [e^{t c_1(\bs\theta)}]$, evaluated at $t=-1$ and $t=\alpha-1$, of the log-likelihood ratio $c_1(\bs\theta)$ under $P=\pi(\bs\theta|\Y)$. This establishes Theorem \ref{thm:D_alpha_general} as the foundational result to build a unified framework for analyzing the inherent privacy guarantees of OPS.

\begin{remark}[Reverse direction]\label{rem:reverse}
Theorem~\ref{thm:D_alpha_general} holds for any ordered pair of neighboring posteriors. Swapping the roles of $\Y$ and $\Y'$, let $\tilde c_1(\bs\theta)\triangleq\log\frac{\ell(\bs\theta;\Y')}{\ell(\bs\theta;\Y)}=-c_1(\bs\theta)$; then
\begin{align}
D_\alpha(Q\|P)=\log\mathbb{E}_{\bs\theta\sim Q}\big[e^{-\tilde c_1(\bs\theta)}\big]+\tfrac{1}{\alpha-1}\log\mathbb{E}_{\bs\theta\sim Q}\big[e^{(\alpha-1)\tilde c_1(\bs\theta)}\big],\label{eqn:D_alpha_reverse}
\end{align}
with expectations taken under $Q=\pi(\bs\theta|\Y')$. Since $d(\Y,\Y')=1 \Leftrightarrow d(\Y',\Y)=1$, $(\alpha, \varepsilon)$-RDP requires $\sup_{(\Y,\Y'):d(\Y,\Y')=1}\max\{D_\alpha(P\|Q),D_\alpha(Q\|P)\}\le\varepsilon$.
\end{remark}

As Eqs.~\eqref{eqn:D_alpha_general_c2}, \eqref{eqn:D_alpha_general} and \eqref{eqn:D_alpha_reverse} depend on the specific neighboring pair $\Y, \Y'$ under consideration, the task of deriving DP guarantees for OPS reduces to obtaining data-independent upper bounds for these exponential moments: of $c_1(\bs\theta)$ under $\pi(\bs\theta|\Y)$ and of $\tilde c_1(\bs\theta)$ under $\pi(\bs\theta|\Y')$.
To this end, we derive a series of results in Sections~\ref{sec:uniform} to \ref{sec:discuss} under progressively weaker regularity conditions on $c_1(\bs\theta)$. These results significantly expand the class of Bayesian models for which explicit upper bounds on $D_{\alpha}$ in both directions can be obtained, while also tightening the bounds available in the existing literature. Furthermore, the derived upper bounds on $D_{\alpha}$ can be interpreted under various DP notions, either through definitions or via well-established conversion results summarized in Proposition~\ref{prop:DPconversion}.


\section{DP Guarantees of OPS with Uniformly Bounded Log-Likelihood Ratio}\label{sec:uniform}
Under uniform boundedness of $c_1(\bs\theta)$, we derive explicit DP guarantees for OPS in Corollaries~\ref{coro:uniform} and~\ref{coro:pureDP} based on Theorem~\ref{thm:D_alpha_general}, and demonstrate their application to categorical likelihoods in Example~\ref{eg:categorical}.

\begin{coro}[Attainable RDP guarantees of OPS with uniformly bounded log-likelihood ratio]\label{coro:uniform}
Suppose that $c_1(\bs\theta)$ is uniformly bounded, that is, there exist finite constants $a, b \in \mathbb R$ with $a<b$ such that $c_1(\bs\theta) \in [a, b]$ for all $\bs\theta\in\bs\Theta$. 
Define $\Delta \triangleq  b-a$ and $A \triangleq e^{-\Delta}\in (0,1), B \triangleq  e^{(\alpha-1)\Delta}>1$. 
Then, for every $\alpha>1$,
{\small
\begin{align}
\max\{D_{\alpha}(P||Q),D_{\alpha}(Q||P)\} \le \varepsilon_{\mathrm{ub}}(\alpha, \Delta) \triangleq \log\left(\frac{(\alpha-1)(B-A)}{\alpha(B-1)}\right)
+\frac{1}{\alpha-1}\log\!\left(\frac{B-A}{\alpha(1-A)}\right),\label{eqn:D_alpha_ub}
\end{align}}
and releasing OPS satisfies $(\alpha, \varepsilon_{\mathrm{ub}}(\alpha, \Delta))$-RDP.
\end{coro}
The full proof is in Appendix~\ref{ape:uniform}. We provide the proof sketch below. First, we show that $D_{\alpha}(P||Q)$ is monotonically increasing with respect to the convex order of $c_1(\bs\theta)$; Second, among all random variables supported on $[a,b]$, the endpoint-supported binary random variable $X^*$ -- defined by $\Pr(X^*=a) = 1-p$ and $\Pr(X^*=b)=p$, with $\mathbb{E}(X^*) = \mathbb{E}[c_1(\bs\theta)]$ -- attains the maximal convex order. Evaluating $D_{\alpha}(P||Q)$ at $X^*$ thus yields an analytical upper bound for a given pair of neighboring datasets, parameterized by $p$. Finally, by maximizing this neighboring pair-specific upper bound over $p\in [0,1]$, we obtain the data-independent upper bound $\varepsilon_{\mathrm{ub}}(\alpha, \Delta)$. 

The bound is sharp over all random variables supported on $[a,b]$, though not necessarily realized by a valid neighboring posterior pair. Since $\tilde c_1(\bs\theta)=-c_1(\bs\theta)\in[-b,-a]$ has the same range $\Delta$, the same bound holds for $D_\alpha(Q\|P)$ by Remark~\ref{rem:reverse}. Corollary~\ref{coro:uniform} reveals two insightful properties of $\varepsilon_{\mathrm{ub}}(\alpha, \Delta)$ in Eq.~\eqref{eqn:D_alpha_ub} for a fixed $\alpha>1$. First, \textbf{shift-invariance} property: $\varepsilon_{\mathrm{ub}}(\alpha, \Delta)$ depends solely on the range $\Delta=b-a$, rather than the specific values of $a$ and $b$, implying that a reparameterization that only shifts $c_1(\bs\theta)$ does not affect the inherent DP guarantee offered by OPS. Second, \textbf{monotonic increase} of $\varepsilon_{\mathrm{ub}}(\alpha, \Delta)$ with $\Delta$: larger values of $\Delta$ lead to weaker privacy guarantees, which aligns with the intuition that greater dissimilarity between the two neighboring posteriors implies weaker intrinsic privacy protection provided by OPS.

Corollary~\ref{coro:pureDP} below recovers the existing pure-DP guarantees for the symmetric case $a=-b$ (e.g., Thm.~2 in \cite{dimitrakakis2014robust}, Thm.~8 in \cite{dimitrakakis2017differential}, and Thm.~1 in \cite{wang2015privacy}) as a special case; the proof is in Appendix~\ref{ape:pure}.
\begin{coro}[Pure-DP guarantee]\label{coro:pureDP}
Under the conditions in Corollary~\ref{coro:uniform}, for every $\alpha>1$,
{
\begin{align}
\max\{D_{\alpha}(P\|Q),D_{\alpha}(Q\|P)\}\le b-a = \Delta = \sup_{\alpha>1}\varepsilon_{\mathrm{ub}}(\alpha,\Delta),
\end{align}}
implying OPS satisfies pure $\varepsilon$-DP with $\varepsilon = \Delta$. 
\end{coro}
While Corollary~\ref{coro:pureDP} recovers the pure-DP guarantee, Corollary~\ref{coro:uniform} is strictly more informative in two respects. \textit{First}, it addresses the attainability gap in the non-degenerate case of $a < b$. While $\Delta$ is a valid upper bound on $D_{\alpha}(P||Q)$, it is generally unattainable at finite $\alpha$ except $a=b$. Specifically, attaining $\Delta = b-a$ would require simultaneously that $(\alpha-1)^{-1}\log \mathbb{E}_{\bs\theta\sim P}[e^{(\alpha-1)c_1(\bs\theta)}] = b$ and $c_2=-a$. However, per Lemma~\ref{lemma:c2_in_c1}, $c_2$ decreases monotonically with $c_1(\bs\theta)$: the condition $c_2=-a$ implies  $c_1(\bs\theta)=a$ almost surely, whereas the condition $(\alpha-1)^{-1}\log \mathbb{E}_{\bs\theta\sim P}[e^{(\alpha-1)c_1(\bs\theta)}] \!=\! b$ requires $c_1(\bs\theta)\!=\!b$ almost surely. These two conditions cannot hold simultaneously unless $b=a$. Corollary \ref{coro:uniform} resolves this attainability gap by explicitly accounting for the interdependence of $c_1(\bs\theta)$ and $c_2$.

\textit{Second}, $\varepsilon_{\mathrm{ub}}(\alpha, \Delta)$ in Eq.~\eqref{eqn:D_alpha_ub} captures the full spectrum of R\'enyi divergence across all $\alpha>1$, whereas Corollary \ref{coro:pureDP} and prior works provide only the static, worst-case bound corresponding to the limit $\alpha \to \infty$. This order-dependent profile in Corollary~\ref{coro:uniform} thus enables sharper RDP accounting for composition. For fixed $\Delta$, expanding $\varepsilon_{\mathrm{ub}}(\alpha, \Delta)$  via Taylor expansion  as $\alpha\rightarrow\infty$ gives
$\varepsilon_{\mathrm{ub}}(\alpha, \Delta)\!=\! \Delta-\alpha^{-1}\log(\alpha(1-e^{-\Delta})) \!+\!\mathcal{O}(\alpha^{-1})$, that is, $\varepsilon_{\mathrm{ub}}(\alpha,\Delta)$ converges to the pure-DP bound $\Delta$ at the rate of $\Theta(\log \alpha / \alpha)$ (see Appendix~\ref{ape:pure} for the proof).


\begin{example}[Categorical likelihood with arbitrary prior]\label{eg:categorical}
    Let $Y \!=\! \{y_i\}_{i=1}^n$ be a dataset, where each $y_i\!\in\! \{1, \ldots, K\}$ is a categorical outcome. Suppose $y_i|\bs\theta\overset{\text{\tiny i.i.d}}{\sim}\text{Categorical}(\bs\theta)$, where $\bs\theta\!=\!(\theta_1,\dots,\theta_K)$ with $\sum_{j=1}^K \theta_j \!=\! 1$ and $\theta_j\!=\!\Pr(y_i\!=\!j)$. Let $s_j(Y)\!=\! \sum_{i=1}^n \mathbf{1}\{y_i=j\}$ denote the $j$-th category count.\vspace{-3pt}
    \begin{enumerate}[label=(\arabic*), leftmargin = *, itemsep=-3pt]
    \item Bounded neighboring ($Y, Y'$): there exist distinct indices $j\neq k$ for $j, k \in \{1, \ldots, K\}$ such that
    $s_j(Y)=s_j(Y')+1,
    s_k(Y)=s_k(Y')-1$, and $s_\ell(Y)=s_\ell(Y')$ for all $\ell\notin\{j,k\}$. 
    {\small
    \begin{align}
    &c_{1, \mathrm{bounded}}(\bs\theta) = \log\frac{\ell(\bs{\theta};Y)}{\ell(\bs{\theta};Y')} = \sum_{\ell=1}^K \big(s_\ell(Y)-s_\ell(Y')\big)\log\theta_\ell = \log\frac{\theta_j}{\theta_k},\notag\\[-3pt]
    &D_{\alpha, \mathrm{bounded}}\big(P\|Q\big) \le \max_{j\ne k}\Big\{\!\log \mathbb{E}_{\bs\theta\sim P}\!\left[\theta_k/\theta_j\right]+
    (\alpha\!-\!1)^{-1}
    \log \mathbb{E}_{\bs\theta\sim P}\left[\left(\theta_j/\theta_k\right)^{\alpha-1}\right]\!\Big\}.\label{eqn:cate_bounded}
    \end{align}}
    \item Unbounded neighboring ($Y, Y'$): 
    there exists a unique index $j\in \{1, \ldots, K\}$ such that
    $s_j(Y)=s_j(Y')+1$, and $s_\ell(Y)=s_\ell(Y')$ for all $\ell\ne j$. 
    {\small
    \begin{align}
        &c_{1, \mathrm{unbounded}}(\bs\theta) = \log\frac{\ell(\bs{\theta};Y)}{\ell(\bs{\theta};Y')} = \sum_{\ell=1}^K \big(s_\ell(Y)-s_\ell(Y')\big)\log\theta_\ell = \log\theta_j,\notag\\[-3pt]
        &D_{\alpha,\mathrm{unbounded}}(P\|Q)\le\max_{j\in\{1,\ldots,K\}}\Big\{\log\mathbb{E}_{\bs\theta\sim P}\big[\theta_j^{-1}\big]+(\alpha-1)^{-1}\log\mathbb{E}_{\bs\theta\sim P}\big[\theta_j^{\alpha-1}\big]\Big\}.\label{eqn:cate_unbounded}
    \end{align}}
    \end{enumerate}
\end{example}
Eqs.~\eqref{eqn:cate_bounded} and ~\eqref{eqn:cate_unbounded} are direct applications of Theorem~\ref{thm:D_alpha_general}; the proof is in Appendix~\ref{ape:multinomial}. Since $c_{1,\text{bounded}}(\bs\theta)$ 
and $c_{1,\text{unbounded}}(\bs\theta)$ diverge as $\theta_j \to 0$, we restrict the parameter space to obtain explicit DP guarantees. Specifically, suppose $\theta_j\in[\eta,1-\eta]$ for some $\eta\in(0,1/2)$ and all $j$; then $c_{1,\text{bounded}}(\bs\theta)=\log(\theta_j/\theta_k)\in[-\log\frac{1-\eta}{\eta},\log\frac{1-\eta}{\eta}]$ and $c_{1,\text{unbounded}}(\bs\theta)=\log\theta_j\in[\log\eta,\log(1-\eta)]$, with ranges $\Delta_{\text{bounded}}=2\log\frac{1-\eta}{\eta}$ and $\Delta_{\text{unbounded}}=\log\frac{1-\eta}{\eta}$, respectively. Hence the bounded neighboring dominates with $\Delta=2\log\frac{1-\eta}{\eta}$. Some concrete examples of the inherent RDP guarantees for OPS when  $\alpha=2$ (where  Eq.~\eqref{eqn:D_alpha_ub} reduces to $\varepsilon_{\mathrm{ub}}(2,\Delta)=\log([1+\cosh(\Delta)]/2)$ with $\Delta=2\log\frac{1-\eta}{\eta}$) and $\alpha = \infty$ are given below:
\begin{itemize}[leftmargin = *, itemsep = 0pt]
    \item If $\eta=0.01$, Corollary~\ref{coro:uniform}
    yields $(2,\sim\!7.804)$-RDP,  and Corollary~\ref{coro:pureDP} yields
    $(\infty,\sim\!9.190)$-RDP.
    \item If $\eta=0.1$, Corollary~\ref{coro:uniform} yields 
    $(2,\sim\!3.033)$-RDP, and Corollary~\ref{coro:pureDP} yields
    $(\infty,\sim\!4.394)$-RDP.
    \item If $\eta=0.3$, Corollary~\ref{coro:uniform} yields 
    $(2,\sim\!0.646)$-RDP, and Corollary~\ref{coro:pureDP} yields
    $(\infty,\sim\!1.695)$-RDP.
\end{itemize}
We remark that Eqs.~\eqref{eqn:cate_bounded} and \eqref{eqn:cate_unbounded} can be evaluated exactly for specific choices of the prior to produce distribution-specific bounds. Examples include the Bernoulli-Beta pair (Appendix~\ref{ape:bernoulli}) and Categorical-Dirichlet pair (Appendix~\ref{ape:multinomial}). 


\section{DP Guarantees of OPS with sub-Gaussian Log-Likelihood Ratio}\label{sec:subgaussian}
In this section, we establish explicit DP guarantees for OPS under a weaker assumption than the uniform boundedness assumption in Section \ref{sec:uniform}, that is, when $c_1(\bs\theta)$ is sub-Gaussian. The main result (Corollary~\ref{coro:subgaussian}) applies to a broader class of Bayesian models than those covered in Corollary~\ref{coro:uniform}. 

\begin{coro}[RDP guarantees of OPS with sub-Gaussian log-likelihood ratio]\label{coro:subgaussian}
Suppose that there exists a variance proxy $\sigma_{\Y,\Y'}^2\in(0,\infty)$ such that for $\forall t\in\mathbb R$, it holds that $\log\mathbb{E}_{\bs\theta\sim P}(e^{t(c_1(\bs\theta)-\mu_{\Y,\Y'})})\le t^2\sigma_{\Y,\Y'}^2/2$
where $\mu_{\Y,\Y'}\!=\!\mathbb{E}_{\bs\theta\sim P}\left[c_1(\bs\theta)\right]$.  Define
\begin{align}\label{eqn:scale}
\textstyle \bar{\sigma}^{2} \triangleq \sup_{(\Y,\Y'): d(\Y, \Y')=1}\sigma_{\Y,\Y'}^2,
\end{align}
where the supremum is over all ordered neighboring pairs.
Then, for every $\alpha>1$,
\begin{align}\label{eqn:eps_subG}
\max\{D_{\alpha}(P||Q), D_{\alpha}(Q||P)\}  \le  \varepsilon_{\mathrm{subGaus}}(\alpha, \bar{\sigma}^2)\triangleq\alpha\bar{\sigma}^2/2,
\end{align}
and releasing OPS satisfies $(\alpha, \varepsilon_{\mathrm{subGaus}}(\alpha, \bar{\sigma}^2))$-RDP.
\end{coro}

The proof is in Appendix~\ref{ape:subgaussian}. Unlike the uniformly bounded case in Corollary~\ref{coro:uniform}, $\varepsilon_{\mathrm{subGaus}}(\alpha, \bar{\sigma}^2)$ in Eq.~\eqref{eqn:eps_subG} scales linearly with $\alpha$ and diverges as $\alpha\!\to\!\infty$. Therefore, the sub-Gaussian regime does not yield pure DP guarantees for OPS. Nevertheless, this linear dependence on $\alpha$ aligns naturally with other relaxed DP notions. Specifically, since Eq.~\eqref{eqn:eps_subG} holds for all $\alpha \!\in\! (1,\infty)$, OPS satisfies $\rho$-zCDP  with $\rho = \bar{\sigma}^2/2$, $\mu$-GDP with $\mu=\bar{\sigma}$ and $(\alpha\bar{\sigma}^2/2 + \tfrac{\log(1/\delta)}{\alpha-1}, \delta)$-DP for $\forall\;0\!<\!\delta \!<\! 1$ via Proposition~\ref{prop:DPconversion}. Furthermore, by Corollary 3 in \cite{mironov2017renyi}, the Gaussian mechanism with scale parameter $\sigma^2$ satisfies $(\alpha, \alpha/(2\sigma^2))$-RDP. This facilitates an intuitive interpretation of Corollary~\ref{coro:subgaussian}:

\vspace{1.5pt}
\textbf{Claim:} When $c_1(\bs\theta)$ is sub-Gaussian, the inherent DP guarantees of OPS are conceptually equivalent to applying the Gaussian mechanism with scale parameter $\bar{\sigma}^{-2}$ defined in Eq.~\eqref{eqn:scale}.
\vspace{1.5pt}

We apply Corollary~\ref{coro:subgaussian} and Theorem~\ref{thm:D_alpha_general} to Generalized Linear Models (GLMs) with Gaussian priors (Section~\ref{sec:GLM}), and to Gaussian likelihood (Section~\ref{sec:GG}), including the Gaussian-Gaussian conjugate model and Bayesian linear regression with Gaussian priors on regression coefficients. Our results show that the choice of neighboring relation can fundamentally affect the tail behavior of $c_1(\bs\theta)$, leading to qualitatively different inherent privacy guarantees.

\subsection{Generalized Linear Models with Gaussian Prior}\label{sec:GLM}
Let each data record be $(\x_i,y_i)$, where $\x_i\in\mathbb R^d$ is the covariate vector and $y_i\in\mathcal{Y}$ is the response.
Let $\Y=\{(\x_i,y_i)\}_{i=1}^n$ denote the dataset.
Conditioned on $\x_i$ and the parameter $\bs\beta \in\mathbb R^d$, the response $y_i$ follows an exponential family distribution: $p(y_i|\x_i,\bs\beta)=
\exp(y_i\,\x_i^\top\bs\beta-\psi(\x_i^\top\bs\beta)+h(y_i))$, where $\psi(\cdot)$ is the log-partition function and $\exp(h(y_i))$ is the base measure (which may include a known dispersion parameter, such as $\sigma^2$ in Gaussian likelihood). 
We show that, under a bounded record-level score condition, $c_1(\bs\beta)$ for GLMs with a Gaussian prior on $\bs\beta$ is sub-Gaussian under the posterior $\pi(\cdot |\Y)$. This yields explicit OPS privacy guarantees for both bounded and unbounded neighboring relations, as formalized in Proposition~\ref{thm:GLM}.  The proof is in Appendix~\ref{ape:GLM}.

\begin{proposition}[GLMs with Gaussian prior]\label{thm:GLM}
Suppose there exists a finite constant $L>0$ such that for every \((\x_i,y_i)\) and every \(\bs\beta\), $\left\|\nabla_{\bs\beta}\log p(y_i| \x_i,\bs\beta)\right\|\le L$, and impose a Gaussian prior on $\bs\beta$, i.e. $\pi(\bs\beta)\!=\! \mathcal N(0,\tau^2 \mathbf{I}_d)$. Then:
\begin{enumerate}[label=(\arabic*), leftmargin = *, itemsep=-3pt]
    \item Bounded neighboring $(\Y, \Y')$: $c_1(\bs\beta)$ is sub-Gaussian under $ \pi(\bs\beta|\Y)$ with variance proxy  $\sigma_{\Y, \Y'}^2\le 4L^2\tau^2$ for every ordered neighboring pair and, per Corollary~\ref{coro:subgaussian}, 
    \begin{align}
    \max\{D_{\alpha}(P\|Q),D_{\alpha}(Q\|P)\}  \le \varepsilon_{\mathrm{GLM, bounded}}(\alpha) \triangleq \alpha\bar{\sigma}^2/2 \le 2\alpha L^2\tau^2.\label{eqn:eps_GLM_bounded}
    \end{align}
    \item Unbounded neighboring $(\Y, \Y')$: $c_1(\bs\beta)$ is sub-Gaussian under $ \pi(\bs\beta|\Y)$ with variance proxy $\sigma_{\Y,\Y'}^2\le L^2\tau^2$ for every ordered neighboring pair and, per Corollary~\ref{coro:subgaussian}, 
    \begin{align}
    \max\{D_{\alpha}(P\|Q),D_{\alpha}(Q\|P)\}  \le\varepsilon_{\mathrm{GLM, unbounded}}(\alpha)\triangleq\alpha\bar{\sigma}^2/2\le\alpha L^2\tau^2/2.\label{eqn:eps_GLM_unbounded}
    \end{align}
\end{enumerate}
Eqs.~\eqref{eqn:eps_GLM_bounded} and \eqref{eqn:eps_GLM_unbounded} taken together, releasing OPS satisfies $(\alpha, 2\alpha L^2\tau^2)$-RDP.
\end{proposition}
 For GLMs with Gaussian priors, as in Example~\ref{eg:categorical}, the bounded neighboring relation yields a larger worst-case R\'enyi divergence, specifically incurring a 4-fold increase relative to the unbounded case. Under bounded neighboring, OPS in this setting can also be viewed as a regularized exponential mechanism, and the bound in Eq.~\eqref{eqn:eps_GLM_bounded} matches the strongly log-concave Lipschitz-perturbation analysis of \cite{gopi2022private}. However, our formulation distinguishes itself by further accommodating unbounded neighboring -- yielding Eq.\eqref{eqn:eps_GLM_unbounded} -- and by embedding the GLM case within a broader framework that supports the non-log-concave-tail analyses developed in Section~\ref{sec:GG}.

Proposition~\ref{thm:GLM} shows how the bounded-score constant $L$, prior variance $\tau^2$, and neighboring relations jointly determine the inherent DP guarantee of OPS from GLMs with Gaussian prior. First, $\varepsilon_{\text{GLM}}$ in Eqs.~\eqref{eqn:eps_GLM_bounded} and \eqref{eqn:eps_GLM_unbounded} scale quadratically with $L$. Thus, records with large score norm can substantially weaken the intrinsic DP guarantee offered by OPS, consistent  with the common practice of data clipping and normalization in DP procedures. Second, $\varepsilon_{\text{GLM}}$ grows linearly with the prior variance $\tau^2$. A larger $\tau^2$ corresponds to a less informative prior, making the posterior more dominated by the likelihood formulated from the private data and thus weakening the inherent DP guarantees of OPS.

The bound $\varepsilon_{\text{GLM}}$ does not explicitly depend on the data size $n$,  as a consequence of the worst-case uniform bound used in our proof. Specifically, we control the variance proxy $\sigma^2_{\Y, \Y'}$ via a uniform score bound $L$ and the strong log-concavity of the posterior induced by the Gaussian prior, whose curvature governed by $\tau^2$. Under these conditions, the only curvature guaranteed uniformly over all datasets and $\bs\beta$ values comes from the prior. Deriving a sharper explicit $n$-dependent guarantee is possible but would require showing that the likelihood contributes a nontrivial curvature, such as by establishing an $n$-dependent lower bound on the Hessian of the negative log-posterior $H(\bs\beta)\!=\!\sum_{i=1}^n \!\psi''(\mathbf{x}_i^\top\bs{\beta})\mathbf{x}_i\mathbf{x}_i^\top \!+\!\tau^{-2}I_d$. However, the likelihood curvature $\psi''(\mathbf{x}_i^\top\bs{\beta})$ can be arbitrarily close to $0$ in the worst case: in logistic regression, $\psi''(\eta) \!=\! \sigma(\eta)(1-\sigma(\eta))$ with $\sigma(t) \!=\! (1\!+\!e^{-t})^{-1}$ vanishes as $|\mathbf{x}_i^\top\bs{\beta}|\!\rightarrow\!\infty$. The likelihood thus provides no guaranteed curvature in general, leaving the Gaussian prior as the only universal source of strong convexity $H(\boldsymbol{\beta}) \!\succeq\! \tau^{-2}I_d$. For instance, if $\sum_{i=1}^n\psi''(\x_i^\top\bs\beta)\x_i\x_i^\top\succeq k(n)\mathbf I_d$ uniformly over all admissible datasets and $\bs\beta$ for some $k(n)$ increasing in $n$, then $\tau^{-2}$ can be replaced by $\tau^{-2}+k(n)$, yielding an $n$-dependent guarantee for a more restricted model class. We leave developing sharper, $n$-dependent DP guarantees under additional curvature or design assumptions for future investigation.

\begin{example}[Logistic regression with Gaussian prior]\label{eg:logistic}
Logistic regression is a special case of the GLM above with binary response $y_i\in\{0,1\}$, log-partition function $\psi(t) = \log(1+e^t)$ and $h(\cdot) = 0$. Suppose \(\|\x_i\|\le R<\infty\) for \(\forall \;i\), then the condition in Proposition~\ref{thm:GLM} holds with $L=R$.
\begin{enumerate}[label=(\arabic*), leftmargin = *,itemsep=-3pt]
    \item Bounded neighboring $(\Y, \Y')$: WLOG, assume ($\Y, \Y'$) differ in the $n$-th observation. 
    \begin{align}
    \max\{D_{\alpha}(P\|Q),D_{\alpha}(Q\|P)\}&\le\varepsilon_{\mathrm{logistic, bounded}}(\alpha)\notag\\
    &=
    \begin{cases}
        \alpha R^2\tau^2/2 &\text{ if }\x_n=\x'_n (\mbox{and thus } y_n\ne y'_n) \\
        2\alpha R^2\tau^2 &\text{ if }\x_n\ne\x'_n (y_n\ne y'_n \mbox{ or } y_n= y'_n) \\
    \end{cases}\label{eqn:eps_logi_bounded}.
    \end{align}
    \item Unbounded neighboring $(\Y, \Y')$: Letting $L=R$ in Eq.~\eqref{eqn:eps_GLM_unbounded} yields 
    \begin{align}
    \max\{D_{\alpha}(P\|Q),D_{\alpha}(Q\|P)\}\le\varepsilon_{\mathrm{logistic, unbounded}}(\alpha)= \alpha R^2\tau^2/2\label{eqn:eps_logi_unbounded}.
    \end{align}
\end{enumerate}
\end{example}
Example~\ref{eg:logistic} suggests that replacing a binary label $y_n$ incurs the same worst-case privacy loss as removing an entire record $(\x_n, y_n)$ while replacing the covariate vector $\x_n$ results in a variance proxy 4-fold that of a record removal. It is worth noting that Theorem 16 in \cite{geumlek2017renyi} examines the same Bayesian logistic regression setting. However, their $(\alpha, 2\alpha R^2\tau^2)$-RDP guarantee for OPS is restricted to the bounded setting with \(\x_n \!=\! \x_n'\), whereas our results cover all neighboring scenarios; in their setting ($\x_n=\x'_n$), our bound $\alpha R^2\tau^2/2$ reduces theirs by 75\%, because label replacement cancels the log-partition terms so that $c_1(\bs\beta)$ is $R$-Lipschitz rather than $2R$-Lipschitz.

\subsection{Gaussian likelihood with Gaussian prior \& linear regression with Gaussian prior}\label{sec:GG}
We analyze two foundational Bayesian models not previously studied for the inherent DP of OPS: (1) univariate Gaussian likelihood with a Gaussian prior for its mean and known variance (Proposition~\ref{thm:GG}); (2) linear regression with a multivariate Gaussian prior for its coefficients and known variance (Proposition~\ref{eg:LR}). 
Notably, both models exhibit fundamentally different privacy profiles depending on neighboring relation: $c_1$ is sub-Gaussian under bounded neighboring, but transitions to a non-sub-Gaussian regime under unbounded neighboring; moreover, the two directions of the R\'enyi divergence also differ in their range of finiteness.

\begin{proposition}[Gaussian-Gaussian conjugate pair with known variance]\label{thm:GG}
    Let $\Y \!=\! \{y_i\}_{i=1}^n$ be the dataset, where $y_i|\theta \overset{\text{\tiny i.i.d}}{\sim}\mathcal{N}(\theta, \sigma^2)$ with known variance $\sigma^2$. With Gaussian prior on $\theta$: $\pi(\theta)= \mathcal N(\mu_0,\tau^2)$, 
    the posterior is $\theta|Y \sim \mathcal{N}(\mu_Y, v_Y^2)$, where $v^2_Y = \sigma^2/\omega$ and $\mu_Y=\mu_0 + \sum_{i=1}^n (y_i-\mu_0)/\omega$, with $\omega = n + \sigma^2/\tau^2$.
    \begin{enumerate}[label=(\arabic*), leftmargin = *, itemsep=-3pt]
    \item Bounded neighboring $(\Y, \Y')$: WLOG, assume $\Y, \Y'$  differ in their $n$-th record $y_n$ and $y'_n$. Then $c_1(\theta)$ is Gaussian with variance $\sigma^2_{Y, Y'} = (y_n-y'_n)^2/(\sigma^2\omega)$. If all records lie in an interval of length $2k\sigma$ for some $k>0$, then $\sigma^2_{Y, Y'} \le \bar{\sigma}^2 = 4k^2/\omega$, and per Corollary~\ref{coro:subgaussian}\vspace{-3pt}
    \begin{align}
    \max\{D_\alpha(P\|Q),D_\alpha(Q\|P)\}\le\varepsilon_{\mathrm{GG, bounded}}(\alpha,\omega) \triangleq \alpha\bar{\sigma}^2/2= 2k^2\alpha/\omega.\label{eqn:eps_GG_bounded}
    \end{align}
    \item  Unbounded neighboring $(\Y, \Y')$: WLOG, assume $\Y'$ is obtained by removing the $n$-th record $y_n$ in $\Y$. Then $c_1(\theta)$ is not sub-Gaussian; nevertheless, each term in the exact expression of the R\'enyi divergence $D_{\alpha}(P\|Q)$ in Eq.~\eqref{eqn:D_alpha_general} can be analytically derived using the moment-generating functions (MGFs) of the non-central $\chi^2$ distribution with one degree of freedom and non-centrality parameter $(\mu_Y\!-\!y_n)^2/v^2_Y$. Specifically,
    {\small
    \begin{align}
        D_{\alpha}(P\|Q) 
        & = A(\alpha, \omega) + \frac{(\mu_Y\!-\!y_n)^2}{2\sigma^2} B(\alpha, \omega)\label{eqn:DPQ_GG_unbounded} \\
        &A(\alpha, \omega)=-\frac{1}{2}\log\left(1\!-\! \omega^{-1}\right)-\frac{\log\left(1\!+\!\frac{\alpha-1}{\omega}\right)}{2(\alpha-1)}, \quad B(\alpha, \omega) = \left[\frac{1}{1-\omega^{-1}}-\frac{1}{1\!+\!\frac{\alpha-1}{\omega}}\right],\notag\\
        D_{\alpha}(Q\|P)
        &=
        \begin{cases}
            \widetilde A(\alpha,\omega)+\frac{(\mu_Y-y_n)^2}{2\sigma^2}\widetilde B(\alpha,\omega) &\text{if } 1<\alpha<\omega,\\
            \infty & \text{if } \alpha\ge\omega,
        \end{cases}\label{eqn:DQP_GG_unbounded}\\
        &\widetilde A(\alpha,\omega)=-\frac12\log\frac{\omega}{\omega-1}-\frac{1}{2(\alpha-1)}\log\frac{\omega-\alpha}{\omega-1},\quad \widetilde B(\alpha,\omega)=\frac{\alpha\omega}{(\omega-1)(\omega-\alpha)}.\notag
    \end{align}}
    If for all $i$, $|y_i\!-\!\mu_Y|\le k^*\sigma$ for some $k^*>0$, then for $1<\alpha<\omega$,
    {\small
    \begin{align}
    \max\{D_\alpha(P\|Q),D_\alpha(Q\|P)\}&\le\varepsilon_{\mathrm{GG, unbounded}}(\alpha, \omega)\notag\\
    &=\max\Big\{A(\alpha,\omega)+\frac{k^{*2}}{2}B(\alpha,\omega),\ \widetilde A(\alpha,\omega)+\frac{k^{*2}}{2}\widetilde B(\alpha,\omega)\Big\}.\label{eqn:eps_GG}
    \end{align}}
    \end{enumerate}
\end{proposition}
The proof is in Appendix~\ref{ape:GG_known}. 
To  our knowledge, Proposition~\ref{thm:GG} gives the first quantitative characterization of how sample size $n$, prior variance $\tau^2$, and neighboring relation jointly determine the inherent DP guarantees of OPS in the Gaussian-Gaussian models. The key takeaways are:

\textbf{Prior regularization.} In both neighboring relations, the influence of the prior is characterized by the precision term $\omega = n +\sigma^2/\tau^2$, where the ratio $\sigma^2/\tau^2$ can be interpreted as the ``effective sample size'' contributed by the prior. A more informative prior (smaller $\tau^2$) increases $\omega$, leading to smaller values for both $\varepsilon_{\text{GG, bounded}}$ and  $\varepsilon_{\text{GG, unbounded}}$, thereby tightening the privacy loss bounds. 

\textbf{Bounded neighboring.} The worst-case privacy loss of OPS is controlled by the maximum difference between possible replacements, as shown in Eq.~\eqref{eqn:eps_GG_bounded}. Specifically, when the substituted records are identical (i.e., $y_n \!=\! y'_n$), both directions vanish. Moreover, for fixed $\alpha, k$ and  $\sigma^2$, $\varepsilon_{\text{GG, bounded}}(\alpha,\omega) \!=\!  2k^2\alpha/\omega \!\rightarrow\! 0$ as $n\!\rightarrow\! \infty$ or $\tau^2 \!\rightarrow\! 0$. This indicates that OPS can achieve arbitrarily strong privacy through increasing the data size or decreasing the prior variance.  

\textbf{Unbounded neighboring.} $D_{\alpha}(P\|Q)$ in Eq.~\eqref{eqn:DPQ_GG_unbounded} decomposes into a structural scale-shift term $A(\alpha, \omega)$, and a data-dependent location-shift term $(\mu_Y\!-\!y_n)^2B(\alpha, \omega)/(2\sigma^2)$.  The lower bound of $D_{\alpha}(P\|Q)$ is attained when the removed record $y_n$ coincides with the original posterior mean $\mu_Y$. In this case, while the removal does not shift the posterior mean, it reduces the posterior precision. This change in the "sharpness" of the distribution is a detectable signal of the record's presence. Thus, $D_{\alpha}(P\|Q)$ reduces to a strictly positive baseline $A(\alpha, \omega)$ -- the minimum unavoidable privacy loss incurred by removal; likewise, $D_\alpha(Q\|P)$ reduces to $\widetilde A(\alpha,\omega)>0$ for addition (Claim~\ref{claim:reverse-AB}).

The two-sided upper bound is in Eq.~\eqref{eqn:eps_GG}, where $k^*$ represents the maximum deviation of a potentially removed record from $\mu_Y$, scaled by $\sigma$. As $n\!\rightarrow\!\infty$, $\omega\to\infty$; $A,\widetilde A=\mathcal{O}(\omega^{-2})$ and $B,\widetilde B=\mathcal{O}(\omega^{-1})$, so $\varepsilon_{\text{GG, unbounded}}(\alpha,\omega)$ vanishes and the admissible range $1<\alpha<\omega$ expands. However, the two directions behave differently: $D_\alpha(P\|Q)$ (removal) is finite for all $\alpha>1$, whereas $D_\alpha(Q\|P)$ (addition) diverges for $\alpha\ge\omega$, since removing a record inflates the posterior variance so that $\pi(\theta|\Y')$ has heavier tails than $\pi(\theta|\Y)$. Consequently, OPS does not satisfy pure DP under unbounded neighboring, and the neighboring relation affects not only the magnitude of the privacy loss but also the range of R\'enyi orders over which a finite guarantee exists.

\begin{proposition}[Linear regression with Gaussian prior]\label{eg:LR}
Let each data record be $(\x_i,y_i)$, where $\x_i\!\in\!\mathbb R^d$ is the covariate vector,  $y_i\!\in\! \mathbb{R}$ is the response, and $y_i|\x_i,\bs\beta \sim \mathcal N(\x_i^\top\bs\beta,\sigma^2)$ with known variance $\sigma^2$. Let \(\Y\!=\! \{(\x_i,y_i)\}_{i=1}^n\) be the dataset, $\x \in\mathbb{R}^{n\times d}$ be the design matrix, $\y\in \mathbb{R}^n$ be the response vector. With Gaussian prior $\pi(\bs\beta)\!=\! \mathcal N(0,\tau^2 \mathbf{I}_d)$, the posterior $\bs\beta|\Y \!\sim\! \mathcal N(\bs\mu_\Y,\Sigma_\Y)$, where $\Sigma_\Y^{-1}\!=\!\sigma^{-2}\x^\top \x\!+\!\tau^{-2}\mathbf{I}_d$ and $\boldsymbol{\mu}_\Y \!=\! \Sigma_\Y \left( \sigma^{-2} \x^\top \mathbf{y} \right)$. Assume $\|\x_i\|\le R<\infty$ for all $i$.
\begin{enumerate}[label=(\arabic*), leftmargin = *, itemsep=-3pt]
    \item Bounded neighboring $(\Y, \Y')$ with $\x_n \!=\! \x'_n$: WLOG, suppose the $n$-th record $(\x_n,y_n)$ in $\Y$ is replaced by $(\x_n,y'_n)$ in $\Y'$, where $y_n\ne y'_n$. Then $c_1(\bs\beta)$ is Gaussian under $\pi(\bs\beta|\Y)$ with variance $\sigma_{\Y,\Y'}^2=(y_n-y_n')^2\x_n^\top\Sigma_\Y\x_n/\sigma^4$. If \(|y_n-y_n'|\le 2k\sigma\) for some \(k>0\), then \(\sigma_{\Y,\Y'}^2\le\bar\sigma^2=4k^2\tau^2R^2/\sigma^2\), and per Corollary~\ref{coro:subgaussian}
    \begin{align}
    \max\{D_\alpha(P\|Q),D_\alpha(Q\|P)\}\le\varepsilon_{\mathrm{LR, bounded}}(\alpha) \triangleq \alpha\bar{\sigma}^2/2= 2\alpha k^2\tau^2 R^2/\sigma^2.\label{eqn:eps_LR_bounded}
    \end{align}
    \item Unbounded neighboring $(\Y, \Y')$: WLOG, suppose $\Y'$ is obtained by removing the $n$-th record $(\x_n,y_n)$ in $\Y$. $c_1(\bs\beta)$ is not sub-Gaussian under $ \pi(\bs\beta|\Y)$. Define $\omega_n =\sigma^2/(\x_n^\top \Sigma_\Y \x_n)$. With $A,B,\widetilde A,\widetilde B$ defined in Proposition~\ref{thm:GG},
    {\small
    \begin{align}
        D_{\alpha}(P\|Q) &= A(\alpha, \omega_n) +\frac{(\x_n^\top \bs\mu_\Y - y_n)^2}{2\sigma^2}B(\alpha, \omega_n),\label{eqn:DPQ_LR_unbounded}\\
        D_{\alpha}(Q\|P) &=
        \begin{cases}
            \widetilde A(\alpha,\omega_n)+\frac{(\x_n^\top \bs\mu_\Y-y_n)^2}{2\sigma^2}\widetilde B(\alpha,\omega_n), &\text{if } 1<\alpha<\omega_n,\\
            \infty, & \text{if } \alpha\ge\omega_n.
        \end{cases}\label{eqn:DQP_LR_unbounded}
    \end{align}}
    Since $\omega_n\ge\omega_{\min}\triangleq1+\sigma^2/(\tau^2R^2)$ for every admissible pair, if for all $i$, $|\x_i^\top \bs\mu_\Y - y_i|\le k^*\sigma$ for some $k^*>0$, then for $1<\alpha<\omega_{\min}$,
    {\small
    \begin{align}
    \max\{D_\alpha(P\|Q),D_\alpha(Q\|P)\}\le\varepsilon_{\mathrm{LR, unbounded}}(\alpha)=\varepsilon_{\mathrm{GG, unbounded}}(\alpha,\omega_{\min}).\label{eqn:eps_LR}
    \end{align}}
\end{enumerate}
\end{proposition}
Proposition~\ref{eg:LR} extends the results in Proposition~\ref{thm:GG} to high-dimensional settings. In the bounded case, the privacy loss is determined by the difference between the replacement records and the posterior precision, and scales linearly with $\alpha$. In the unbounded case,  the data geometry determines the inherent DP guarantees of OPS through $\omega_n = \sigma^2/(\x_n^\top\Sigma_\Y\x_n)$. Specifically, $A(\alpha, \omega_n)$ is governed by the influence of the removed record on the estimate; high-influential points have low $\omega_n$ and may substantially reduce the posterior precision, creating a detectable signal even when the posterior mean is unchanged. As in Proposition~\ref{thm:GG}, the reverse direction is finite only for $\alpha<\omega_n$; thus high-influence points, with small $\omega_n$, both enlarge the privacy loss and shrink the range of R\'enyi orders admitting a finite guarantee.

\section{Discussion}\label{sec:discuss}
This work advances our understanding of the inherent privacy guarantees of Bayesian posterior sampling in a unified R\'enyi-divergence framework. Our results demonstrate that the intrinsic stochasticity of OPS provides meaningful privacy protection -- shaped by the interplay among prior information, data size, model structure, and neighboring relationships -- even without explicit noise injection. By offering explicit and interpretable privacy characterizations across diverse Bayesian models, our analysis clarifies when and how OPS can serve as a reliable DP mechanism.  Our results offer a principled basis for reducing external noise injection and thereby improving the utility of released statistical outputs while maintaining formal privacy protection. The work also offers insights on designing more transparent and effective private data analysis methods in sensitive applications.

There are several interesting directions for future work. For example, the sub-Gaussian regime examined in Section~\ref{sec:subgaussian} can be further relaxed to sub-exponential log-likelihood ratios, thereby extending the framework to settings where both uniform boundedness and sub-Gaussianity fail.
\begin{coro}[RDP guarantees of OPS with sub-exponential log-likelihood ratio]\label{coro:subexponential}
Suppose that there exist $\nu_{\Y,\Y'}^2\in(0,\infty)$ and $b_{\Y,\Y'}\in(0,\infty)$ such that for $\forall\;t\in\mathbb{R}$ satisfying $|t|<b_{\Y,\Y'}^{-1}$, it holds that $\log\mathbb{E}_{\bs\theta\sim P}\!\left(e^{t(c_1(\bs\theta)-\mu_{\Y,\Y'})}\right)\le \frac{t^2\nu^2_{\Y,\Y'}}{2(1-b_{\Y,\Y'}|t|)}$
where $\mu_{\Y,\Y'}=\mathbb{E}_{\bs\theta\sim P}\left[c_1(\bs\theta)\right]$. Define
\begin{align}\label{eqn:eps_subexp}
\varepsilon_{\mathrm{subExp}}(\alpha, b_{\Y,\Y'}, \nu_{\Y,\Y'}^2) \triangleq \frac{\nu_{\Y,\Y'}^2}{2}\cdot
    \frac{\alpha(1-2b_{\Y,\Y'})+2b_{\Y,\Y'}}{(1-b_{\Y,\Y'})\bigl(1-b_{\Y,\Y'}(\alpha-1)\bigr)}.
\end{align}
When $b_{\Y,\Y'}<\min \{(\alpha-1)^{-1}, 1\}$ for all ordered neighboring pairs, 
{
\begin{align}
\max\{D_{\alpha}(P||Q), D_{\alpha}(Q||P)\}  \le \sup_{(\Y,\Y'): d(\Y,\Y')=1}\varepsilon_{\mathrm{subExp}}(\alpha, b_{\Y,\Y'}, \nu_{\Y,\Y'}^2),
\end{align}}
and releasing OPS satisfies $(\alpha,\sup_{(\Y,\Y')}\varepsilon_{\mathrm{subExp}}(\alpha,b_{\Y,\Y'},\nu^2_{\Y,\Y'}))$-RDP.
\end{coro}

The proof is in Appendix~\ref{ape:subexp}.
While $\varepsilon_{\text{subExp}}(\alpha, b_{\Y,\Y'}, \nu_{\Y,\Y'}^2) $ is tight for fixed sub-exponential parameters $(\nu_{\mathbf{Y},\Y'}^2, b_{\mathbf{Y},\Y'})$, deriving these parameters for a specific model often introduces looseness. For example, in the unbounded case of the Gaussian-Gaussian pair (Proposition~\ref{thm:GG}), the log-likelihood ratio satisfies the sub-exponential condition as shown in Appendix~\ref{ape:GG_known}, but the derived parameters rely on upper bounds and are thus sub-optimal. In contrast, directly exploiting the exact MGFs of a non-central $\chi^2$ distribution and calculating $D_\alpha(P\|Q)$ via Theorem~\ref{thm:D_alpha_general} yields sharper bounds. This illustrates the potential gap between general tail-based bounds and model-specific exact characterizations. 

Despite this limitation, Corollary~\ref{coro:subexponential} remains a broadly applicable theoretical tool for analyzing inherent DP guarantees of OPS when both uniform boundedness and sub-Gaussianity fail. Remark~\ref{rem:LR_case2} demonstrates this for linear regression under bounded-neighboring with covariate replacement, a case excluded from Proposition~\ref{eg:LR} because \(c_1(\bs\beta)\) contains a non-sub-Gaussian quadratic term.
\begin{remark}[Linear regression with Gaussian prior, bounded neighboring with $\x_n \ne \x'_n$]
\label{rem:LR_case2}
When $\x_n \neq \x_n'$, $c_1(\bs\beta)$ contains both affine 
and quadratic terms in $\bs\beta$ and is generally not 
sub-Gaussian under $\pi(\bs\beta|\Y)$. Suppose $\|\x_i\|\le R < \infty$ and $|y_i| \leq M_y< \infty$ for $\forall \;i$, and assume the scale 
restriction $\sigma^2 > 4\tau^2R^2$, then $c_1(\bs\beta)$ is sub-exponential with parameters
{
\begin{align}\label{eqn:para_LR_ne}
b_{\Y, \Y'} = \frac{4\tau^2R^2}{\sigma^2}, \quad 
\nu^2_{\Y, \Y'} \leq \frac{8M_y^2R^2\tau^2}{\sigma^4} + 
\frac{16\tau^4R^4}{\sigma^4} + 
\frac{8n^2\tau^6M_y^2R^6}{\sigma^8},
\end{align}}
which can be plugged into Corollary~\ref{coro:subexponential} for explicit RDP bound.
\end{remark}
The proof is in Appendix~\ref{ape:LR_bounded}. The parameters in Eq.~\eqref{eqn:para_LR_ne} also illustrate the conservatism that can arise when deriving model-specific sub-exponential parameters for Corollary~\ref{coro:subexponential}: the Cauchy-Schwarz decomposition, operator norm bounds, and row-wise boundedness assumptions all introduce looseness, leading to a restricted valid range of $\alpha$ and an $n^2$-dependent variance proxy. Developing tighter techniques for estimating $(\nu_{\Y, \Y'}^2, b_{\Y, \Y'})$, especially for complex models where exact MGFs are analytically intractable, is a promising direction for future work. Other future directions include leveraging OPS as a principled DP mechanism in applications such as privacy-preserving synthetic data generation, Markov chain Monte Carlo algorithms, and variational inference, with the goal of optimizing the trade-off between privacy guarantees and statistical utility.



\clearpage
\bibliographystyle{unsrtnat}
\bibliography{ref}

\newpage
\appendix

\setcounter{tocdepth}{0} 
\addtocontents{toc}{\protect\setcounter{tocdepth}{2}} 
\renewcommand{\thefigure}{A.\arabic{figure}}
\renewcommand{\thetable}{A.\arabic{table}}
\renewcommand{\thelemma}{A.\arabic{lemma}}
\renewcommand{\thetheorem}{A.\arabic{theorem}}
\renewcommand{\thecon}{A.\arabic{condition}}
\renewcommand{\theequation}{A.\arabic{equation}}
\renewcommand{\thesection}{A.\arabic{section}}
\renewcommand{\thesubsection}{A.\arabic{section}.\arabic{subsection}}
\renewcommand{\thesubsubsection}{A.\arabic{section}.\arabic{subsection}.\arabic{subsubsection}}

\setcounter{section}{0}
\setcounter{subsection}{0}
\setcounter{figure}{0}
\setcounter{table}{0}
\setcounter{theorem}{0}
\setcounter{equation}{0}

\setlength{\cftsecindent}{0pt}
\setlength{\cftsecnumwidth}{2.5em}

\setlength{\cftsubsecindent}{24pt}
\setlength{\cftsubsecnumwidth}{3.5em}

\setlength{\cftsubsubsecindent}{60pt}
\setlength{\cftsubsubsecnumwidth}{6em}
\newcommand{\appendixtoc}{
  \begingroup
  \renewcommand{\contentsname}{Appendix Contents}
  \let\clearpage\relax
  \setlength{\cftbeforesubsecskip}{0pt}
  \tableofcontents
  \endgroup
}
\appendixtoc 


\section{Proof of Lemma \ref{lemma:c2_in_c1}}\label{ape:c2_in_c1}
\begin{proof}
The posterior of parameter $\bs\theta$ given data $\Y$ is
\begin{align}\label{eqn:BayesRule}
    \pi(\bs\theta|\Y) &= \frac{\pi(\bs\theta)\prod_{i=1}^{n}p(\y_i|\bs\theta)}{\int \prod_{i=1}^{n}p(\y_i|\bs\theta)\pi(\bs\theta) d\bs\theta}
    \overset{\triangle}{=}\frac{\pi(\bs\theta)\ell(\bs\theta;\Y)}{\int \ell(\bs\theta;\Y)\pi(\bs\theta) d\bs\theta} 
    = \frac{\pi(\bs\theta)\ell(\bs\theta;\Y)}{p(\Y)}.
\end{align}
For any pair of neighboring datasets $\Y$ and $\Y'$ where $d(\Y, \Y')=1$, the privacy loss for releasing posterior sample $c(\bs\theta)$ per Eq~\eqref{eqn:PrivLoss} can be decomposed as:
\begin{align}
    c(\bs\theta)
    &=
    \log\left(\frac{\pi(\bs\theta|\Y)}{\pi(\bs\theta|\Y')}\right)=\underbrace{\log\left(\frac{\ell(\bs\theta;\Y)}{\ell(\bs\theta;\Y')}\right)}_{\text{\normalsize $\triangleq\; c_1(\bs\theta)$}}+\underbrace{
    \log\left(\frac{p(\Y')}{p(\Y)}\right)}_{\text{\normalsize $\triangleq\; c_2$}}. \label{eqn:decomp_ape}
\end{align}
Note that $p(\Y')$
\begin{align}
=\;&\!\!\int\! \ell(\bs\theta;\Y')\pi(\bs\theta) d\bs\theta  = \!\!\int\!\frac{\ell(\bs\theta;\Y')}{\ell(\bs\theta;\Y)}\ell(\bs\theta;\Y)\pi(\bs\theta) d\bs\theta =\!\! \int\!\frac{\ell(\bs\theta;\Y')}{\ell(\bs\theta;\Y)}\pi(\bs\theta|\Y)p(\Y) d\bs\theta\;\;\mbox{ per Eq.~\eqref {eqn:BayesRule}} \notag\\
=\;&p(\Y)  \int\!\frac{\ell(\bs\theta;\Y')}{\ell(\bs\theta;\Y)}\pi(\bs\theta|\Y)d\bs\theta\notag\\
\Rightarrow \;\;&\frac{p(\Y')}{p(\Y)} 
=\int\frac{\ell(\bs\theta;\Y')}{\ell(\bs\theta;\Y)}\pi(\bs\theta|\Y) d\bs\theta = \int e^{-c_1(\bs\theta)}\pi(\bs\theta|\Y)d\bs\theta = \mathbb{E}_{\bs\theta \sim \pi(\cdot|\Y)}\left[e^{-c_1(\bs\theta)}\right].
\end{align}
Since $c_2 = \log(p(\Y')/p(\Y))$,  we have
\begin{equation}
c_2 
= \log \left(\frac{p(\Y')}{p(\Y)}\right) = \log\left(\mathbb{E}_{\bs\theta \sim \pi(\cdot|\Y)}\left[e^{-c_1(\bs\theta)}\right]\right). \label{eqn:c2_in_c1}
\end{equation}
\end{proof}


\section{Proof of Theorem~\ref{thm:D_alpha_general}}\label{ape:D_alpha_general}
\begin{proof}
Per the definition of R\'enyi divergence in Eq~\eqref{eqn:renyi_div},
\begin{align}
    D_{\alpha}(P || Q) =\frac{1}{\alpha-1}\log \mathbb{E}_{x\sim Q}\left(\frac{P(x)}{Q(x)}\right)^\alpha
    = \frac{1}{\alpha-1}\log \mathbb{E}_{x\sim P}\left(\frac{P(x)}{Q(x)}\right)^{\alpha-1}.
\end{align}
For any neighboring datasets $\mathbf{Y}$ and $\mathbf{Y}'$, we can calculate the R\'enyi divergence between the corresponding neighboring posterior distributions $\pi(\bs\theta|\Y)$ and $\pi(\bs\theta|\Y')$ as below
\begin{align}
    D_{\alpha}(\pi(\bs\theta|\mathbf{Y})\|\pi(\bs\theta|\mathbf{Y}')) &= \frac{1}{\alpha-1}\log \mathbb{E}_{\bs\theta\sim \pi(\cdot|\mathbf{Y})}\left[\left(\frac{\pi(\bs\theta|\Y)}{\pi(\bs\theta|\Y')}\right)^{\alpha-1}\right]\notag\\
    &=\frac{1}{\alpha-1}\log \mathbb{E}_{\bs\theta\sim \pi(\cdot|\mathbf{Y})}\left[e^{(\alpha-1)c(\bs\theta)}\right]\notag\\
    &=\frac{1}{\alpha-1}\log \mathbb{E}_{\bs\theta\sim \pi(\cdot|\mathbf{Y})}\left[e^{(\alpha-1) (c_1(\bs\theta) + c_2)}\right]\quad\text{ per Eq~\eqref{eqn:decomp_ape}}\notag\\
    &=c_2 +\frac{1}{\alpha-1}\log \mathbb{E}_{\bs\theta\sim \pi(\cdot|\mathbf{Y})}\left[e^{(\alpha-1) c_1(\bs\theta)}\right]\label{eqn:D_alpha_general_c2_ape}
\end{align}
Plug Eq.~\eqref{eqn:c2_in_c1} into Eq.~\eqref{eqn:D_alpha_general_c2_ape}, 
\begin{align}
    D_{\alpha}(\pi(\bs\theta|\mathbf{Y})\|\pi(\bs\theta|\mathbf{Y}'))=& \log\mathbb{E}_{\bs\theta \sim \pi(\cdot|\mathbf{Y})}\left[e^{-c_1(\bs\theta)}\right]+\frac{1}{\alpha-1}\log \mathbb{E}_{\bs\theta\sim \pi(\cdot|\mathbf{Y})}\left[e^{(\alpha-1)c_1(\bs\theta)}\right]\label{eqn:D_alpha_general_ape}\\
    = & \log\Bigg[\!\left(\mathbb{E}_{\bs\theta \sim \pi(\cdot|\mathbf{Y})}\left[e^{-c_1(\bs\theta)}\right]\right)\left(\mathbb{E}_{\bs\theta\sim \pi(\cdot|\mathbf{Y})}\left[e^{(\alpha-1)\cdot c_1(\bs\theta)}\right]\right)^{\frac{1}{\alpha-1}}\Bigg].
\end{align}
\end{proof}

\section{Proof of Corollary~\ref{coro:uniform}}\label{ape:uniform}
\begin{proof}
Assume $c_1(\bs\theta)$ is uniformly bounded, i.e. there exist finite constants $a, b \in \mathbb R$ with $a<b$ such that $\forall \bs\theta\in\bs\Theta, c_1(\bs\theta) \in [a, b]$. Thus $-c_1(\bs\theta)\in[-b,-a]$ and per monotonicity. 
\begin{align}
c_2= \log\mathbb{E}_{\bs\theta \sim \pi(\cdot|\mathbf{Y})}\left[e^{-c_1(\bs\theta)}\right]&\in [-b, -a]\\
\frac{1}{\alpha-1}\log \mathbb{E}_{\bs\theta\sim \pi(\cdot|\mathbf{Y})}\left[e^{(\alpha-1) c_1(\bs\theta)}\right] &\in [a, b].
\end{align}
Taken together with the decomposition in Eq.~\eqref{eqn:D_alpha_general_c2_ape}, we obtain 
\begin{align}
    D_{\alpha}(\pi(\bs\theta|\mathbf{Y})||\pi(\bs\theta|\mathbf{Y}')) &=c_2 +\frac{1}{\alpha-1}\log \mathbb{E}_{\bs\theta\sim \pi(\cdot|\mathbf{Y})}\left[e^{(\alpha-1)\cdot c_1(\bs\theta)}\right]\le b-a.\label{eqn:b-a}
\end{align}
Note that to attain the \textbf{equality} in Eq.~\eqref{eqn:b-a} of the upper bound  $b-a$ would require simultaneously $c_1(\bs\theta) = a$ (so that $c_2 = -a$) and $c_1(\bs\theta) = b$ (so that $\frac{1}{\alpha-1}\log \mathbb{E}_{\bs\theta\sim \pi(\cdot|\mathbf{Y})}\left[e^{(\alpha-1) c_1(\bs\theta)}\right] = b$) almost surely, which is impossible unless $a=b$. Therefore, this bound is not tight and unattainable.


To obtain an attainable upper bound of $D_{\alpha}(\pi(\bs\theta|\mathbf{Y})||\pi(\bs\theta|\mathbf{Y}'))$, it suffices to maximize the two exponential moments of $c_1(\bs\theta)$ involved in Eq.~\eqref{eqn:b-a}, subject to the constraint $c_1(\bs\theta)\in [a, b]$. This maximization can be achieved using convex-order arguments, as in \citep{shaked2007stochastic}. 
WLOG, let $X$ be any random variable uniformly bounded with $[a,b]$, and $X^*$ denote a binary random variable supported on $\{a,b\}$ as below, with the same mean as $X$:
\begin{align}\label{eqn:X*}
X^* =
\begin{cases}
a, & \text{with probability } 1-p\\
b, & \text{with probability } p
\end{cases};
\end{align}
that is, $\mathbb{E}(X)=\mathbb{E}(X^*)=a(1-p)+bp$ for some $p\in [0,1]$.
For any convex function $\phi$ and any $x\in [a,b]$, convexity implies the chord bound: 
\begin{align}
    \phi(x)\ \le\ \lambda(x)\phi(b)+(1-\lambda(x))\phi(a) \quad \forall x \in [a,b]
\end{align}
Let $\lambda(x)=\frac{x-a}{b-a}$ and take expectation of $\phi(x)$ over its distribution, we have:
\begin{align}
    \mathbb E[\phi(X)] \le p\phi(b)+ (1-p)\phi(a)=\mathbb E[\phi(X^*)].
\end{align}

\begin{lemma}[Theorem 3.A.7 in \cite{shaked2007stochastic}]
    Let $X$ and $Y$ be two random variables with the same finite means. Then $X \le_{\mathrm{cx}} Y$ if and only if,
    \begin{equation}
        \mathbb{E}(\phi(X))\le \mathbb{E}(\phi(Y)) \qquad \text{for any convex function $\phi$}.
    \end{equation}
\end{lemma}
Hence, per the Lemma above, we have $X \le_{\mathrm{cx}} X^*$. Since $X$ can be any random variable uniformly bounded with $[a,b]$, $X^*$ is maximal in convex order among all such
bounded random variables. 

Also, since the exponential function $e^x$ is convex, and the R\'enyi divergence functional $D_\alpha$ in Eq.~\eqref{eqn:D_alpha_general} is fully determined by the exponential moments of $c_1(\bs\theta)$, it follows that $D_\alpha$ is monotone with respect to the convex order. Setting $X = c_1(\bs\theta)$, we therefore obtain
$c_1(\bs\theta)\le_{\mathrm{cx}} X^*$,  which implies that the $D_{\alpha}(\pi(\bs\theta|\mathbf{Y})||\pi(\bs\theta|\mathbf{Y}'))$ is maximized when $c_1(\bs\theta)$ coincides with the endpoint-supported binary random variable $X^*$ as defined in Eq.~\eqref{eqn:X*}. Then per Eq.~\eqref{eqn:D_alpha_general_ape}
\begin{align}
    &D_{\alpha}(\pi(\bs\theta|\mathbf{Y})||\pi(\bs\theta|\mathbf{Y}'))\notag\\
    =\;& \log\mathbb{E}_{\bs\theta \sim \pi(\cdot|\mathbf{Y})}\left[e^{-c_1(\bs\theta)}\right]+\frac{1}{\alpha-1}\log \mathbb{E}_{\bs\theta\sim \pi(\cdot|\mathbf{Y})}\left[e^{(\alpha-1)c_1(\bs\theta)}\right]\notag\\
    \le\;&  \log\mathbb{E}\left[e^{-X^*}\right]+\frac{1}{\alpha-1}\log \mathbb{E}\left[e^{(\alpha-1)X^*}\right]\quad \text{per convex order dominance}\notag\\
    =\;&  \log\left((1-p)e^{-a}+pe^{-b}\right)
    + \frac{1}{\alpha-1}\log\!\left((1-p)e^{(\alpha-1)a}+pe^{(\alpha-1)b}\right)\notag\\
    =\;&\log\!\left(1-p+pe^{-(b-a)}\right)
    + \frac{1}{\alpha-1}\log\!\left(1-p+pe^{(\alpha-1)(b-a)}\right)\label{eqn:D_alpha_p}\\
    \triangleq\;& D_{\alpha}(p)
\end{align}
For a fixed pair of neighboring datasets $\Y$ and $\Y'$, in order to align the means of $c_1(\bs\theta)$ and $X^*$, one should require,
\begin{align}
    &\mathbb{E}_{\bs\theta\sim \pi(\cdot|\Y)}\big[c_1(\bs\theta)\big] = a(1-p)+bp\notag\\
    \quad\Rightarrow\;&  \hat{p} = \frac{\mathbb{E}_{\bs\theta\sim \pi(\cdot|\Y)}\big[c_1(\bs\theta)\big]-a}{b-a},
\end{align}
However, since \(\mathbb E_{\bs\theta\sim\pi(\cdot|\Y)}[c_1(\bs\theta)]\) may take any value in $[a,b]$ across neighboring dataset pairs, the associated $\hat{p}$ ranges over $[0,1]$. Thus, an upper bound on $D_{\alpha}(\pi(\bs\theta|\mathbf{Y})||\pi(\bs\theta|\mathbf{Y}'))$ that is valid over all neighboring datasets need to be obtained by maximizing $D_{\alpha}(p)$ in Eq.~\eqref{eqn:D_alpha_p} over $p\in [0,1]$.

Let $\Delta = b-a$ and define $A= e^{-\Delta}\in (0,1)$, $B= e^{(\alpha-1)\Delta}>1$.  To maximize $D_\alpha(p)$ over $p\in[0,1]$,
take the first derivative on both sides with respect to $p$:
\begin{align}
    D'_{\alpha}(p) = \frac{dD_{\alpha}(p) }{dp} = \frac{A-1}{1-p+pA} + \frac{1}{\alpha-1}\cdot\frac{B-1}{1-p+pB}.
\end{align}
Setting $D'_{\alpha}(p)=0$, we have
\begin{align}
 p^*&=\arg\max_{p\in [0,1]} D_{\alpha}(p)\notag\\
&=\frac{(B\!-\!1)\!-\!(\alpha-1)(1-A)}{\alpha(1-A)(B-1)}\notag\\
&=\frac{ \big(e^{(\alpha-1)\Delta}-1\big)\! -\! (\alpha\!-\!1)\big(1-e^{-\Delta}\big)}
{\alpha\big(1-e^{-\Delta}\big)\big(e^{(\alpha-1)\Delta}-1\big)}.
\end{align}
Therefore,
\begin{align}
    D_{\alpha} &\le \log\left(1-p^*+p^*e^{-\Delta}\right)
    + \frac{1}{\alpha-1}\log\left(1-p^*+p^*e^{(\alpha-1)\Delta}\right)\notag\\
    & =  \log\left(1-\frac{(B\!-\!1)\!-\!(\alpha-1)(1-A)}{\alpha(1-A)(B-1)}+\frac{(B\!-\!1)\!-\!(\alpha-1)(1-A)}{\alpha(1-A)(B-1)}A\right)\notag\\
    & \;\;\;\;+ \frac{1}{\alpha-1}\log\left(1-\frac{(B\!-\!1)\!-\!(\alpha-1)(1-A)}{\alpha(1-A)(B-1)}+\frac{(B\!-\!1)\!-\!(\alpha-1)(1-A)}{\alpha(1-A)(B-1)}B\right)\notag\\
    & =  \log\left(\frac{\alpha\cancel{(1-A)}(B-1)-\cancel{(1-A)}(B\!-\!1)\!+\!(\alpha-1)\cancel{(1-A)}\!-\!(\alpha-1)\cancel{(1-A)}A}{\alpha\cancel{(1-A)}(B-1)}\right)\notag\\
    & \;\;\;\;+ \frac{1}{\alpha-1}\log\left(\frac{\alpha(1-A)\cancel{(B-1)}-\cancel{(B-1)}+B\cancel{(B-1)}\!-\!(\alpha-1)(1-A)\cancel{(B-1)}}{\alpha(1-A)\cancel{(B-1)}}\right)\notag\\
    &= \log\left(\frac{(\alpha-1)(B-A)}{\alpha(B-1)}\right)
    +\frac{1}{\alpha-1}\log\!\left(\frac{B-A}{\alpha(1-A)}\right)\triangleq \varepsilon_{\mathrm{ub}}(\alpha, \Delta).\label{eqn:eps_ub}
\end{align}
\end{proof}
\paragraph{Reverse direction.}
If $c_1(\bs\theta)\in[a,b]$ for all $\bs\theta\in\Theta$, then $\tilde c_1(\bs\theta)=-c_1(\bs\theta)\in[-b,-a]$ for all $\bs\theta\in\Theta$, an interval of the same width $\Delta=b-a$. The argument above uses only the support width of the log-likelihood ratio and holds under an arbitrary posterior. Applying it to $\tilde c_1(\bs\theta)$ under $Q=\pi(\bs\theta|\Y')$ via Eq.~\eqref{eqn:D_alpha_reverse} yields $D_\alpha(Q\|P)\le\varepsilon_{\mathrm{ub}}(\alpha,\Delta)$. Hence $\max\{D_\alpha(P\|Q),D_\alpha(Q\|P)\}\le\varepsilon_{\mathrm{ub}}(\alpha,\Delta)$, and the $(\alpha,\varepsilon_{\mathrm{ub}}(\alpha,\Delta))$-RDP guarantee holds in both directions.

\begin{claim}
    For $\forall \alpha > 1$, and for $\forall \Delta_2 \ge \Delta_1\ge0$, we have $0 \le \varepsilon_{\mathrm{ub}}(\alpha, \Delta_2)- \varepsilon_{\mathrm{ub}}(\alpha, \Delta_1)\le \Delta_2-\Delta_1$.
\end{claim}
\begin{proof}
Take the first derivative of Eq.~\eqref{eqn:D_alpha_p} with respect to $\Delta$ 
\begin{align}
\partial_\Delta D_\alpha(p,\Delta)&=-\frac{p e^{-\Delta}}{1-p+pe^{-\Delta}}+\frac{p e^{(\alpha-1)\Delta}}{1-p+pe^{(\alpha-1)\Delta}}= h\left(e^{(\alpha-1)\Delta}\right) - h\left(e^{-\Delta}\right),
\end{align}
where $h(u)=\frac{pu}{1-p+pu}$ for $u>0$. Since
\begin{align}
    \frac{\partial h(u)}{\partial u} = \frac{p(1-p+pu)-p^2u}{(1-p+pu)^2} = \frac{p(1-p)}{(1-p+pu)^2}\ge 0,
\end{align}
$h(u)$ is non-decreasing in $u$. Per $\alpha > 1$ and $\Delta \ge 0$, $e^{(\alpha-1)\Delta} \ge e^{-\Delta}$, then $\partial_\Delta D_\alpha(p,\Delta) \ge 0$ and the equality holds only in the degenerate cases $p=0, p=1$, or $\Delta = 0$.

Since the pointwise maximum preserves order, for any $\Delta_2 \ge \Delta_1$ and $\forall p \in [0,1], \forall \alpha > 1$:
\begin{align}
    &0\le D_\alpha(p,\Delta_2)-D_\alpha(p,\Delta_1)\le \Delta_2-\Delta_1\notag\\
    &\text{Taking the maximum over $p$ on both sides}\notag\\
    \Rightarrow\; & 0 \le \varepsilon_{\mathrm{ub}}(\alpha, \Delta_2)- \varepsilon_{\mathrm{ub}}(\alpha, \Delta_1)\le \Delta_2-\Delta_1.
\end{align}
We can conclude that $ \varepsilon_{\mathrm{ub}}(\alpha, \Delta)$ is $1$-Lipschitz in $\Delta$. In other words, $ \varepsilon_{\mathrm{ub}}(\alpha, \Delta)$ grows  at most linearly with slope at most 1.

\end{proof}

\section{Proof of Corollary~\ref{coro:pureDP}}\label{ape:pure}
\begin{claim}
For fixed $\Delta$, as $\alpha \rightarrow \infty$, $\varepsilon_{\mathrm{ub}}(\alpha, \Delta)=\Delta-\alpha^{-1}\log\left(\alpha(1-e^{-\Delta})\right)+\mathcal{O}\left(\alpha^{-1}\right).$
\end{claim}
\begin{proof}
 $A=e^{-\Delta} \in (0,1)$. As $\alpha \rightarrow \infty, B = e^{(\alpha-1)\Delta}\rightarrow \infty$. Thus
\begin{align}
 \frac{B-A}{B-1} = 1+\frac{1-A}{B-1} = 1+\mathcal{O}(B^{-1})=1+\mathcal{O}(e^{-(\alpha-1)\Delta})\label{eqn:1st_term_expansion}
\end{align}
Plug Eq.~\eqref{eqn:1st_term_expansion} into the first term in Eq.~\eqref{eqn:eps_ub}
\begin{align}
\log\left(\frac{(\alpha-1)(B-A)}{\alpha(B-1)}\right) =& \log\left(\frac{\alpha-1}{\alpha}\right) + \log\left(\frac{B-A}{B-1}\right)\notag\\
=& \log\left(1-\alpha^{-1}\right) + \log\left(1+\mathcal{O}(e^{-(\alpha-1)\Delta})\right)\notag\\
=& -\alpha^{-1}+\mathcal{O}(\alpha^{-2})+\mathcal{O}\left(e^{-(\alpha-1)\Delta}\right)\label{eqn:2nd_term}\quad \text{as $\alpha\rightarrow\infty$}
\end{align}
Since $B-A = e^{(\alpha-1)\Delta}-e^{-\Delta} = e^{-\Delta}\big(e^{\alpha\Delta}-1\big)$, 
\begin{align}
\frac{1}{\alpha-1}\log\!\left(\frac{B-A}{\alpha(1-A)}\right) =& \frac{1}{\alpha-1}\Big[\log\!\left(B-A\right)-\log(\alpha(1-A))\Big]\notag\\
=& \frac{1}{\alpha-1}\Big[-\Delta+\log\left(e^{\alpha\Delta}-1\right)-\log(\alpha(1-A))\Big]\notag\\
=& \frac{1}{\alpha-1}\Big[-\Delta+\alpha\Delta+\log\left(1-e^{-\alpha\Delta}\right)-\log(\alpha(1-A))\Big]\notag\\
=& \Delta - \frac{1}{\alpha-1}\log(\alpha(1-A)) + \frac{\log\left(1-e^{-\alpha\Delta}\right)}{\alpha-1}\label{eqn:1st_term}
\end{align}
Plug Eqs.~\eqref{eqn:1st_term} and \eqref{eqn:2nd_term} in Eq.~\eqref{eqn:eps_ub}, then 
    \begin{align}
        \varepsilon_{\mathrm{ub}}(\alpha, \Delta)  &= \frac{1}{\alpha-1}\log\!\left(\frac{B-A}{\alpha(1-A)}\right)+\log\left(\frac{(\alpha-1)(B-A)}{\alpha(B-1)}\right) \notag\\
        &= \Delta - \frac{1}{\alpha-1}\log(\alpha(1-A)) + \frac{\log\left(1-e^{-\alpha\Delta}\right)}{\alpha-1}-\alpha^{-1}+\mathcal{O}(\alpha^{-2})+\underbrace{\text{\normalsize $\mathcal{O}\left(e^{-(\alpha-1)\Delta}\right)$}}_{\text{dominated by $\alpha^{-2}$}}\notag\\
        &\quad \text{since } \frac{1}{\alpha-1} = \frac{1}{\alpha} + \mathcal{O}(\alpha^{-2})\text{ and } \log\left(1-e^{-\alpha\Delta}\right) \approx e^{-\alpha\Delta}\quad\text{for sufficiently large } \alpha \notag\\
        &=\Delta - \left(\frac{1}{\alpha}+\mathcal{O}(\alpha^{-2})\right)\cdot\log(\alpha(1-A))+\mathcal{O}\left(\frac{e^{-\alpha\Delta}}{\alpha}\right)-\alpha^{-1}+\mathcal{O}(\alpha^{-2})\notag\\
        &\quad \text{since } \frac{e^{-\alpha\Delta}}{\alpha}\le\frac{1}{\alpha^2}
        \le\frac{\log\alpha}{\alpha^2}\le \frac{1}{\alpha}\le\frac{\log\alpha}{\alpha}
        \quad\text{for sufficiently large }\alpha\notag\\
        &=\Delta -\frac{1}{\alpha}\log(\alpha(1-e^{-\Delta}))+\mathcal{O}\left(\alpha^{-1}\right)
    \end{align}
Similar to Appendix~\ref{ape:uniform}, $\tilde c_1(\bs\theta)\in[-b,-a]$ under $Q$, we also have $D_\alpha(Q\|P)\le b-a=\Delta$ for every $\alpha>1$. Hence $\max\{D_\infty(P\|Q),D_\infty(Q\|P)\}\le\Delta$, and the pure $\Delta$-DP conclusion of Corollary~5 is two-sided.
\end{proof}

\section{Proof of Corollary~\ref{coro:subgaussian}}\label{ape:subgaussian}
\begin{proof}
\emph{Step 1: a fixed ordered pair.}
Fix an ordered neighboring pair $(\Y,\Y')$, and let $P=\pi(\cdot|\Y)$, $Q=\pi(\cdot|\Y')$. Suppose $c_1(\bs\theta)=\log\{\ell(\bs\theta;\Y)/\ell(\bs\theta;\Y')\}$ is sub-Gaussian under $P$, i.e., there exists $\sigma^2_{\Y,\Y'}\in(0,\infty)$ such that for all $t\in\mathbb R$,
\begin{equation}\label{eqn:subG_condition}
    \log\mathbb{E}_{\bs\theta\sim P}\!\left[e^{t(c_1(\bs\theta)-\mu_{\Y,\Y'})}\right]\le \frac{t^2\sigma^2_{\Y,\Y'}}{2},
    \qquad \mu_{\Y,\Y'}\triangleq\mathbb{E}_{\bs\theta\sim P}[c_1(\bs\theta)].
\end{equation}
Setting $t=\alpha-1$ and $t=-1$ in Eq.~\eqref{eqn:subG_condition}, respectively, bounds the two exponential moments in Eq.~\eqref{eqn:D_alpha_general_ape}:
\begin{align}
    \log \mathbb{E}_{\bs\theta\sim P}\big[e^{(\alpha-1)c_1(\bs\theta)}\big]&\le (\alpha-1)\mu_{\Y,\Y'} + \frac{(\alpha-1)^2\sigma^2_{\Y,\Y'}}{2},\notag\\
    c_2=\log \mathbb{E}_{\bs\theta\sim P}\big[e^{-c_1(\bs\theta)}\big] &\le -\mu_{\Y,\Y'}+\frac{\sigma^2_{\Y,\Y'}}{2}.\notag
\end{align}
Plugging both into Eq.~\eqref{eqn:D_alpha_general_c2_ape},
\begin{align}
    D_{\alpha}(P\|Q) &= c_2 +\frac{1}{\alpha-1}\log \mathbb{E}_{\bs\theta\sim P}\big[e^{(\alpha-1)c_1(\bs\theta)}\big]\notag\\
    &\le -\mu_{\Y,\Y'}+\frac{\sigma^2_{\Y,\Y'}}{2}+\mu_{\Y,\Y'} + \frac{(\alpha-1)\sigma^2_{\Y,\Y'}}{2}
    = \frac{\alpha}{2}\,\sigma^2_{\Y,\Y'}. \label{eqn:subG_pairwise}
\end{align}

\emph{Step 2: uniform, two-sided bound.}
Applying Step~1 to the reversed pair $(\Y',\Y)$ gives $D_\alpha(Q\|P)\le\frac{\alpha}{2}\sigma^2_{\Y',\Y}$, where, by definition, $\sigma^2_{\Y',\Y}$ is the variance proxy of $\tilde c_1(\bs\theta)=-c_1(\bs\theta)$ under $Q$. Since $(\Y',\Y)$ is an ordered neighboring pair whenever $(\Y,\Y')$ is, both $\sigma^2_{\Y,\Y'}$ and $\sigma^2_{\Y',\Y}$ are bounded by
\begin{equation}
\bar\sigma^2\triangleq\sup_{(\Y,\Y'):\,d(\Y,\Y')=1}\sigma^2_{\Y,\Y'},
\end{equation}
where the supremum is over all ordered neighboring pairs in the admissible data domain. Hence, by Eq.~\eqref{eqn:subG_pairwise},
\begin{equation}
\sup_{(\Y,\Y'):\,d(\Y,\Y')=1}\max\{D_\alpha(P\|Q),D_\alpha(Q\|P)\}\le\frac{\alpha\bar\sigma^2}{2}\triangleq\varepsilon_{\mathrm{subGaus}}(\alpha,\bar\sigma^2),
\end{equation}
which completes the proof.
\end{proof}

\section{Proof of Corollary~\ref{coro:subexponential}}\label{ape:subexp}
\begin{proof}
If assume $c_1(\bs\theta)$ is sub-Exponential under $\pi(\cdot|\Y)$, i.e. there exist
$\nu_{\Y,\Y'}^2\in(0,\infty)$ and $b_{\Y,\Y'}\in(0,\infty)$ such that for all $t\in\mathbb{R}$ satisfying
$|t|<1/b_{\Y,\Y'}$, it holds that
\begin{equation}\label{eqn:subexp_con}
    \log\mathbb{E}_{\bs\theta\sim P}\!\left(e^{t(c_1(\bs\theta)-\mu_{\Y,\Y'})}\right)\le \frac{t^2\nu^2_{\Y,\Y'}}{2(1-b_{\Y,\Y'}|t|)};
\end{equation}
Then, by setting $t$ as $-1$ and $\alpha-1$ in Eq.~\eqref{eqn:subexp_con} respectively, the required exponential moments of $c_1(\bs\theta)$ under $\pi(\cdot|\Y)$ in Eq.~\eqref{eqn:D_alpha_general_ape} can be derived as:
\begin{align}
    &\text{when $t=\alpha-1$, we have: }\notag\\
    &\qquad \log \mathbb{E}_{\bs\theta\sim \pi(\cdot|\Y)}\left(e^{(\alpha-1) \cdot c_1(\bs\theta)}\right) \le (\alpha-1) \mu_{\Y,\Y'} + \frac{(\alpha-1)^2\nu^2_{\Y,\Y'}}{2(1-b_{\Y,\Y'}(\alpha-1))}, \\
    &\text{when $t=-1$, we have: }\notag\\
    &\qquad c_2=\log \mathbb{E}_{\bs\theta\sim \pi(\cdot| \Y)}[e^{-c_1(\bs\theta)}]\le -\mu_{\Y,\Y'}+\frac{\nu_{\Y,\Y'}^2}{2(1-b_{\Y,\Y'})}.
\end{align}
Plugging the above two into Eq.~\eqref{eqn:D_alpha_general_c2_ape}, under $b_{\Y,\Y'}<1$ and $b_{\Y,\Y'}(\alpha-1)<1$,
\begin{align}
    D_{\alpha}(\pi(\bs\theta|\Y)\|\pi(\bs\theta|\Y')) 
    &= c_2 +\frac{1}{\alpha-1}\log \mathbb{E}_{\bs\theta\sim \pi(\cdot|\Y)}\left[e^{(\alpha-1)\cdot c_1(\bs\theta)}\right]\notag\\
    &\le -\mu_{\Y,\Y'}+\frac{\nu_{\Y,\Y'}^2}{2(1\!-\!b_{\Y,\Y'})}+\frac{1}{\alpha\!-\!1}\Big[(\alpha\!-\!1) \mu_{\Y,\Y'} + \frac{(\alpha\!-\!1)^2\nu^2_{\Y,\Y'}}{2(1-b_{\Y,\Y'}(\alpha-1))}\Big]\notag\\
    &= \frac{\nu_{\Y,\Y'}^2}{2(1-b_{\Y,\Y'})}+ \frac{(\alpha-1)\nu^2_{\Y,\Y'}}{2(1-b_{\Y,\Y'}(\alpha-1))}\notag\\
    &= \frac{\nu_{\Y,\Y'}^2}{2}\cdot\frac{1-b_{\Y,\Y'}(\alpha-1)+(\alpha-1)(1-b_{\Y,\Y'})}{(1-b_{\Y,\Y'})(1-b_{\Y,\Y'}(\alpha-1))}\notag\\
    &= \frac{\nu_{\Y,\Y'}^2}{2}
        \frac{\alpha(1-2b_{\Y,\Y'})+2b_{\Y,\Y'}}{(1-b_{\Y,\Y'})\bigl(1-b_{\Y,\Y'}(\alpha-1)\bigr)} = \varepsilon_{\text{subExp}}(\alpha, b_{\Y,\Y'}, \nu^2_{\Y,\Y'}). \label{eqn:subexp_pairwise}
\end{align}
Applying the above to the reversed pair $(\Y',\Y)$, i.e., to $\tilde c_1(\bs\theta)=-c_1(\bs\theta)$ under $\pi(\cdot|\Y')$, gives $D_{\alpha}(\pi(\bs\theta|\Y')\|\pi(\bs\theta|\Y))\le\varepsilon_{\text{subExp}}(\alpha, b_{\Y',\Y}, \nu^2_{\Y',\Y})$, provided $b_{\Y',\Y}<\min\{(\alpha-1)^{-1},1\}$. Since $(\Y',\Y)$ is an ordered neighboring pair whenever $(\Y,\Y')$ is, if $b_{\Y,\Y'}<\min\{(\alpha-1)^{-1},1\}$ for all ordered neighboring pairs, then
\begin{align}
\max\{D_\alpha(P\|Q),D_\alpha(Q\|P)\}\le\sup_{(\Y,\Y'):\,d(\Y,\Y')=1}\varepsilon_{\text{subExp}}(\alpha, b_{\Y,\Y'}, \nu^2_{\Y,\Y'})
\end{align}
uniformly over all neighboring pairs, which completes the proof.
\end{proof}
\section{Proof of Example~\ref{eg:categorical} for Categorical data}\label{ape:multinomial}

\begin{proof}
Let $y_i \in \{1,\ldots,K\} $ denote a categorical outcome with $K$ categories. Let $\bs\theta=(\theta_1,\dots,\theta_K)$, where $\sum_{j=1}^K \theta_j = 1$ and  $\theta_j=\Pr(y_i=j)$. Assume $y_i|\bs\theta\sim\text{Categorical}(\bs\theta)$ independently for $i=1,\ldots,n$, and the likelihood of the dataset $Y = \{y_1, \ldots, y_n\}$ is $\ell (\bs\theta; \Y) = \prod_{j=1}^K\theta_j^{s_j(Y)}$, where $s_j(Y) = \sum_{i=1}^n \mathbf{1}\{y_i=j\}$ is the category counts.

\subsection{bounded neighboring}
WLOG, two neighboring datasets $Y$ and $Y'$ are both of size $n$ and differ in exactly one observation. Specifically, there exist distinct indices $j\neq k$ and $j, k \in \{1, \ldots, K\}$ such that
$s_j(Y)=s_j(Y')+1,
s_k(Y)=s_k(Y')-1$, and $s_\ell(Y)=s_\ell(Y')$ for all $\ell\notin\{j,k\}$.
\begin{align}
    &c_1(\bs\theta) = \log\frac{\ell(\bs{\theta};Y)}{\ell(\bs{\theta};Y')} = \sum_{\ell=1}^K \big(s_\ell(Y)-s_\ell(Y')\big)\log\theta_\ell = \log\frac{\theta_j}{\theta_k}\\
    \Rightarrow\; &D^{(jk)}_{\alpha, \text{bounded}}\!\big(\pi(\cdot| Y)\,\|\,\pi(\cdot| Y')\big)
    =\log \mathbb{E}_{\theta\sim\pi(\cdot| Y)}\!\left[\frac{\theta_k}{\theta_j}\right]+
    \frac{1}{\alpha-1}\left(
    \log \mathbb{E}_{\theta\sim\pi(\cdot| Y)}\!\left[\left(\frac{\theta_j}{\theta_k}\right)^{\alpha-1}\right]
    \right).\label{eqn:D_alpha_multinomial_bounded}
\end{align}
Finally, to obtain an RDP guarantee that holds for all neighboring datasets under bounded neighboring, one takes the maximum over all ordered pairs $(j,k)$ corresponding to a single-record change:
\begin{equation}
    D_{\alpha, \text{bounded}}\!\big(\pi(\cdot|Y)\,\|\,\pi(\cdot| Y')\big) = \max_{j,k\in \{1, \ldots, K\}, j\ne k}\Big\{D^{(jk)}_{\alpha, \text{bounded}}\big(\pi(\cdot|Y)\,\|\,\pi(\cdot| Y')\big)\Big\},
\end{equation}
Since the maximum ranges over all ordered category pairs $(j,k)$ and all datasets $Y$ in the domain, the reverse ordering $(Y',Y)$, which is itself a bounded-neighboring pair with $j$ and $k$ exchanged and posterior $\pi(\cdot|Y')$, is already included.
\paragraph{Dirichlet prior.} Assume a Dirichlet prior:
$\pi(\bs\theta)\!=\!\mathrm{Dir}(\alpha_1,\dots,\alpha_K).$ Then
$\bs\theta|Y \!\sim\! \mathrm{Dir}(\beta_1,\dots,\beta_K)$, where $\beta_j=\alpha_j+s_j(Y)$ for $j \in \{1, \ldots, K\}.$ We can calculate the moment below:
\begin{align}
    &\mathbb{E}_{\bs\theta\sim\pi(\cdot| Y)}\!\left[\frac{\theta_k}{\theta_j}\right]
    =\frac{\beta_k}{\beta_j-1} = \frac{\alpha_k + s_k(Y)}{\alpha_j + s_j(Y)-1}, \text{ if $\beta_j > 1$}\label{eqn:dir_c2_bounded}\\
    &\mathbb{E}_{\boldsymbol{\theta}\sim\pi(\cdot| Y)}\!\left[\left(\frac{\theta_j}{\theta_k}\right)^{\alpha-1}\right] =
    \frac{\Gamma(\beta_j+\alpha-1)}{\Gamma(\beta_j)}
    \cdot
    \frac{\Gamma(\beta_k-\alpha+1)}{\Gamma(\beta_k)}, \text{ if $\beta_k > \alpha-1$}
    \label{eqn:dir_c1_bounded}
\end{align}
Therefore, plug Eqns.~\eqref{eqn:dir_c1_bounded}\eqref{eqn:dir_c2_bounded}  into Eq~\eqref{eqn:D_alpha_multinomial_bounded}, we can derive the closed-form expression for R\'enyi Divergence:
\begin{align}
D^{(jk)}_{\alpha, \text{bounded}}\!\big(\pi(\cdot|Y)\,\|\,\pi(\cdot| Y')\big)
= \log\frac{\beta_k}{\beta_j-1}+
\frac{1}{\alpha-1}\left[
\log\left(
\frac{\Gamma(\beta_j-1+\alpha)}{\Gamma(\beta_j)}
\cdot
\frac{\Gamma(\beta_k+1-\alpha)}{\Gamma(\beta_k)}
\right)
\right].\notag
\end{align}
Finally, with $\alpha < 1+\min_k \beta_k$,
\begin{equation}
    D_{\alpha, \text{bounded}}\!\big(\pi(\cdot|Y)\,\|\,\pi(\cdot| Y')\big) = \max_{j,k\in \{1, \ldots, K\}, j\ne k}\Big\{D^{(jk)}_{\alpha, \text{bounded}}\big(\pi(\cdot|Y)\,\|\,\pi(\cdot| Y')\big)\Big\}.
\end{equation}

\paragraph{Special case.} If $\exists \;\eta \in (0,1/2)$ such that $\forall j \in \{1, \ldots, K\}, \theta_j \in [\eta, 1-\eta]$, then$|c_1(\bs\theta)|\le \log(\frac{1-\eta}{\eta})$.

\subsection{unbounded neighboring}
WLOG, let dataset $Y$ has size $n$ and dataset $Y'$ is of size $n-1$ by removing the $n$-th record $y_n\in \{1, \ldots, K\}$. Specifically, there exist one index $j\in \{1, \ldots, K\}$ such that
$s_j(Y)=s_j(Y')+1$, and $s_\ell(Y)=s_\ell(Y')$ for all $\ell\ne j$.
\begin{align}
    &c_1(\bs\theta) = \log\frac{\ell(\bs{\theta};Y)}{\ell(\bs{\theta};Y')} = \sum_{\ell=1}^K \big(s_\ell(Y)-s_\ell(Y')\big)\log\theta_\ell = \log\theta_j\\
    \Rightarrow\; &D^{(j)}_{\alpha, \text{unbounded}}\!\big(\pi(\cdot| Y)\,\|\,\pi(\cdot| Y')\big)
    =\log \mathbb{E}_{\theta\sim\pi(\cdot| Y)}\!\left[\theta_j^{-1}\right]+
    \frac{1}{\alpha-1}\left(
    \log \mathbb{E}_{\theta\sim\pi(\cdot| Y)}\!\left[\theta_j^{\alpha-1}\right]
    \right).\label{eqn:D_alpha_multinomial}
\end{align}
For the reverse direction, $\tilde c_1(\bs\theta)=-c_1(\bs\theta)=-\log\theta_j$, and by Eq.~\eqref{eqn:D_alpha_reverse},
\begin{align}
    D^{(j)}_{\alpha, \text{unbounded}}\!\big(\pi(\cdot| Y')\,\|\,\pi(\cdot| Y)\big)
    =\log \mathbb{E}_{\bs\theta\sim\pi(\cdot| Y')}\!\left[\theta_j\right]+
    \frac{1}{\alpha-1}
    \log \mathbb{E}_{\bs\theta\sim\pi(\cdot| Y')}\!\left[\theta_j^{-(\alpha-1)}\right].\label{eqn:D_alpha_multinomial_rev}
\end{align}

Finally, to obtain an RDP guarantee that holds for all neighboring datasets under unbounded neighboring, one takes the maximum over all $j$ :
\begin{align*}
    D_{\alpha, \text{unbounded}}\!\big(\pi(\cdot|Y)\,\|\,\pi(\cdot| Y')\big) = \max_{j\in \{1, \ldots, K\}}\max\Big\{&D^{(j)}_{\alpha, \text{unbounded}}\!\big(\pi(\cdot|Y)\,\|\,\pi(\cdot| Y')\big),\\
    &D^{(j)}_{\alpha, \text{unbounded}}\!\big(\pi(\cdot|Y')\,\|\,\pi(\cdot| Y)\big)\Big\},
\end{align*}

\paragraph{Dirichlet prior.} Assume a Dirichlet prior:
$\pi(\bs\theta)=\mathrm{Dir}(\alpha_1,\dots,\alpha_K).$ Then
$\bs\theta|Y \sim \mathrm{Dir}(\beta_1,\dots,\beta_K)$, where $\beta_j=\alpha_j+s_j(Y)$ for $j \in \{1, \ldots, K\}.$ And the marginal $\theta_j|Y\sim\text{Beta}(\beta_j, \sum_{\ell=1}^K\beta_\ell-\beta_j)$, based on which, we can calculate the moments below:
\begin{align}
    &\mathbb{E}_{\bs\theta\sim\pi(\cdot| Y)}\left[\theta_j^{-1}\right]
    =\frac{\sum_{\ell=1}^K\beta_\ell-1}{\beta_j-1} \text{ if $\beta_j > 1$}\label{eqn:dir_c2}\\
    &\mathbb{E}_{\bs\theta\sim\pi(\cdot| Y)}\!\left[\theta_j^{\alpha-1}\right] =
    \frac{\Gamma(\beta_j+\alpha-1)}{\Gamma(\beta_j)}
    \cdot
    \frac{\Gamma(\sum_{\ell=1}^K\beta_\ell)}{\Gamma(\sum_{\ell=1}^K\beta_\ell+\alpha-1)}.
    \label{eqn:dir_c1}
\end{align}
Therefore, plug Eqns.~\eqref{eqn:dir_c1}\eqref{eqn:dir_c2}  into Eq~\eqref{eqn:D_alpha_multinomial}, we can derive the closed-form expression for R\'enyi Divergence:
{\small
\begin{align}
D^{(j)}_{\alpha, \text{unbounded}}\!\big(\pi(\cdot|Y)\,\|\,\pi(\cdot| Y')\big)
= \log \frac{\sum_{\ell=1}^K\beta_\ell-1}{\beta_j-1}+
\frac{1}{\alpha-1}\log\left[\frac{\Gamma(\beta_j+\alpha-1)}{\Gamma(\beta_j)}
    \cdot
    \frac{\Gamma(\sum_{\ell=1}^K\beta_\ell)}{\Gamma(\sum_{\ell=1}^K\beta_\ell+\alpha-1)}
\right].
\end{align}}
Since $s_j(Y)\ge1$, $\beta_j\ge\alpha_j+1>1$, so this direction is finite for every $\alpha>1$.
For the reverse direction, $\theta_j|Y'\sim\text{Beta}(\beta_j-1, \sum_{\ell=1}^K\beta_\ell-\beta_j)$, when $1<\alpha<\beta_j$
{\small
\begin{align}
D^{(j)}_{\alpha, \text{unbounded}}\!\big(\pi(\cdot|Y')\,\|\,\pi(\cdot| Y)\big)
= \log\frac{\beta_j-1}{\sum_{\ell=1}^K\beta_\ell-1}+
\frac{1}{\alpha-1}\log\left[\frac{\Gamma(\beta_j-\alpha)}{\Gamma(\beta_j-1)}
    \cdot
    \frac{\Gamma(\sum_{\ell=1}^K\beta_\ell-1)}{\Gamma(\sum_{\ell=1}^K\beta_\ell-\alpha)}\right],
\end{align}}
but $=\infty$ for $\alpha\ge\beta_j$. Since $\beta_j\ge\alpha_j+1$ with equality when $s_j(Y)=1$, the two-sided divergence is finite uniformly over all unbounded neighboring pairs if and only if $\alpha<1+\min_j\alpha_j$; hence OPS with a Dirichlet prior does not satisfy pure DP under unbounded neighboring.

\paragraph{Special Case.} If $\exists \;\eta \in (0,1/2)$ such that $\forall j \in \{1, \ldots, K\}, \theta_j \in [\eta, 1-\eta]$, then $|c_1(\bs\theta)|\le \log(\frac{1}{\eta})$. In the reverse direction, $\tilde c_1(\bs\theta)=-\log\theta_j$ has a range of the same width, so Corollary~\ref{coro:uniform} applies in both directions.

\end{proof}

\subsection{Proof of Bernoulli data}\label{ape:bernoulli}
\begin{proof}
Assume observed dataset $Y = \{y_i\}_{i=1}^n$, where $y_i|\theta \overset{\text{i.i.d}}{\sim}\text{Bernoulli}(\theta)$ with $\theta=\Pr(y_i=1)$ and $1-\theta=\Pr(y_i=0)$.  

\subsubsection{bounded neighboring}
WLOG, suppose neighboring datasets $Y,Y'$, both of size $n$, differ in their $n$-th record $y_n$ and $y'_n$. Then the log-likelihood ratio can be derived as
\begin{align}
    &\ell(\theta;Y)=\theta^{y_n+\sum_{i=1}^{n-1} y_i}(1-\theta)^{n-y_n-\sum_{i=1}^{n-1} y_i}\notag\\
    &\ell(\theta;Y')=\theta^{y'_n+\sum_{i=1}^{n-1} y_i}(1-\theta)^{n-y'_n-\sum_{i=1}^{n-1} y_i}\notag\\
    \Rightarrow \;&c_1(\theta)=\log\frac{\ell(\theta;Y)}{\ell(\theta;Y')}
    =(y_n-y'_n)\log\left(\frac{\theta}{1-\theta}\right) = \begin{cases}
        \log\left(\frac{\theta}{1-\theta}\right) & \text{if $y_n -y'_n = 1$}\\
        -\log\left(\frac{\theta}{1-\theta}\right) & \text{if $y_n -y'_n = -1$}
    \end{cases}
\end{align}
If $y_n - y'_n = 1$ with $\mathbb{E}_{\bs\theta \sim \pi(\cdot|\mathbf{Y})}\!\left[\frac{1-\theta}{\theta}\right]< \infty$ and $\mathbb{E}_{\theta\sim\pi(\cdot| Y)}\!\left[\left(\frac{\theta}{1-\theta}\right)^{\alpha-1}\right]<\infty$:
\begin{align}
D_{\alpha, \text{bounded}}^{(+)}\big(\pi(\cdot| Y)\,\|\,\pi(\cdot| Y')\big)
= \log\mathbb{E}_{\theta \sim \pi(\cdot|Y)}\!\left[\frac{1\!-\!\theta}{\theta}\right] + 
\frac{1}{\alpha\!-\!1}\log \mathbb{E}_{\theta\sim\pi(\cdot| Y)}\!\left[\left(\frac{\theta}{1\!-\!\theta}\right)^{\alpha-1}\right].\notag
\end{align}
If $y_n - y'_n = -1$ with $\mathbb{E}_{\theta \sim \pi(\cdot|Y)}\!\left[\frac{\theta}{1-\theta}\right]< \infty$ and $\mathbb{E}_{\theta\sim\pi(\cdot| Y)}\!\left[\left(\frac{1-\theta}{\theta}\right)^{\alpha-1}\right]<\infty$:
\begin{align}
D_{\alpha, \text{bounded}}^{(-)}\big(\pi(\cdot| Y)\,\|\,\pi(\cdot| Y')\big)
= \log\mathbb{E}_{\theta \sim \pi(\cdot|Y)}\left[\frac{\theta}{1\!-\!\theta}\right] + 
\frac{1}{\alpha-1}\log \mathbb{E}_{\theta\sim\pi(\cdot| Y)}\!\left[\left(\frac{1\!-\!\theta}{\theta}\right)^{\alpha-1}\right].\notag
\end{align}
Without additional structure on the posterior, the worst-case R\'enyi divergence over bounded neighboring is given by
\begin{equation}
    D_{\alpha, \text{bounded}}\big(\pi(\cdot| Y)\,\|\,\pi(\cdot| Y')\big)\le\max\{D_{\alpha, \text{bounded}}^{(+)},\,D_{\alpha, \text{bounded}}^{(-)}\}\notag
\end{equation}
The reverse ordering $(Y',Y)$ is itself a bounded-neighboring pair with $y'_n-y_n=\mp1$ and posterior $\pi(\cdot|Y')$, i.e., it is of the form $D^{(\mp)}_{\alpha,\text{bounded}}$ evaluated at the dataset $Y'$. Hence taking the maximum over $D^{(+)}$ and $D^{(-)}$ and over all datasets in the domain already covers the reverse direction.

\paragraph{Special case.} If $\theta\in [\eta, 1-\eta]$ for some constant $\eta \in (0,1/2)$. Since $\log(\frac{\theta}{1-\theta})$ is monotonically increasing, $|c_1(\theta)|\le \log(\frac{1-\eta}{\eta})$ uniformly. 

\paragraph{Beta prior.} If $\pi(\theta) = \text{Beta}(a, b)$, and define $s = \sum_{i=1}^n y_i$,  then $\pi(\theta|Y) = \text{Beta}(A, B)$, where $A = a+s$, $B = b+n-s$. \\
If $y_n - y'_n = 1$ and  $A > 1, B > \alpha-1$
\begin{align}
&\mathbb{E}_{\theta\sim\pi(\cdot| Y)}\!\left[\frac{1-\theta}{\theta}\right]
=\frac{B}{A-1}\notag\\
&\mathbb{E}_{\theta\sim\pi(\cdot| Y)}\!\left[\left(\frac{\theta}{1-\theta}\right)^{\alpha-1}\right]
=
\frac{\text{Beta}(A-1+\alpha,B+1-\alpha)}{\text{Beta}(A,B)} = \frac{\Gamma(A+\alpha-1)}{\Gamma(A)}\cdot
\frac{\Gamma(B-(\alpha-1))}{\Gamma(B)}\notag\\
\Rightarrow \; &D_{\alpha, \text{bounded}}^{(+)}\big(\pi(\cdot| Y)\,\|\,\pi(\cdot| Y')\big)
= \log\left[\frac{B}{A-1}\right] + \frac{1}{\alpha-1}
\log \left[\frac{\Gamma(A+\alpha-1)}{\Gamma(A)}\cdot
\frac{\Gamma(B-(\alpha-1))}{\Gamma(B)}\right].
\end{align}
If $y_n - y'_n = -1$, and $B>1, A>\alpha-1$
\begin{align}
    &\mathbb{E}_{\theta\sim\pi(\cdot| Y)}\!\left[\frac{\theta}{1-\theta}\right]
    =\frac{A}{B-1}\notag\\
    &\mathbb{E}_{\theta\sim\pi(\cdot| Y)}\!\left[\left(\frac{1-\theta}{\theta}\right)^{\alpha-1}\right]=
    \frac{\text{Beta}(A+1-\alpha,B-1+\alpha)}{\text{Beta}(A,B)}= \frac{\Gamma(A-(\alpha-1))}{\Gamma(A)}\cdot
    \frac{\Gamma(B+(\alpha-1))}{\Gamma(B)}\notag\\
    \Rightarrow \;& D_{\alpha, \text{bounded}}^{(-)}\big(\pi(\cdot| Y)\,\|\,\pi(\cdot| Y')\big)
    = \log\left[\frac{A}{B-1}\right] + \frac{1}{\alpha-1}
    \log \left[\frac{\Gamma(A-(\alpha-1))}{\Gamma(A)}\cdot
    \frac{\Gamma(B+(\alpha-1))}{\Gamma(B)}\right] \notag\\
\end{align}

\begin{remark}
    To ensure finite Beta-moment, $\alpha < \alpha_{\max} = 1 + \min\{a+s,\; b+n-s\}$.
\end{remark}

\subsubsection{unbounded neighboring}
WLOG, suppose neighboring datasets $Y,Y'$, where $Y$ is of size $n$ and $Y'$ is of size $n-1$ by removing the $n$-th record $y_n$. Then similarly, the log-likelihood ratio can be derived as
\begin{align}
    &\ell(\theta;Y)=\theta^{y_n+\sum_{i=1}^{n-1} y_i}(1-\theta)^{n-y_n-\sum_{i=1}^{n-1} y_i}\notag\\
    &\ell(\theta;Y')=\theta^{\sum_{i=1}^{n-1} y_i}(1-\theta)^{n-1-\sum_{i=1}^{n-1} y_i}\notag\\
    \Rightarrow \;&c_1(\theta)=\log\frac{\ell(\theta;Y)}{\ell(\theta;Y')}
    =y_n\log\theta+(1-y_n)\log(1-\theta) = \begin{cases}
        \log\theta & \text{if $y_n= 1$}\\
        \log\left(1-\theta\right) & \text{if $y_n = 0$}
    \end{cases}
\end{align}
If $y_n =1$,
\begin{align}
D_{\alpha, \text{unbounded}}^{(1)}\big(\pi(\cdot| Y)\|\pi(\cdot| Y')\big)
= \log\mathbb{E}_{\bs\theta \sim \pi(\cdot|Y)}\left[\theta^{-1}\right]+ 
\frac{1}{\alpha-1}\log \mathbb{E}_{\theta\sim\pi(\cdot| Y)}\!\left[\theta^{\alpha-1}\right].\notag
\end{align}
If $y_n = 0$,
\begin{align}
D_{\alpha, \text{unbounded}}^{(0)}\big(\pi(\cdot| Y)\,\|\,\pi(\cdot| Y')\big)
= \log\mathbb{E}_{\bs\theta \sim \pi(\cdot|Y)}\left[(1-\theta)^{-1}\right]+ 
\frac{1}{\alpha-1}\log \mathbb{E}_{\theta\sim\pi(\cdot| Y)}\!\left[\left(1-\theta\right)^{\alpha-1}\right].\notag
\end{align}
For the reverse direction, $\tilde c_1(\theta)=-c_1(\theta)$ and the expectations are taken under $\pi(\cdot|Y')$:
\begin{align}
D_{\alpha, \text{unbounded}}^{(1)}\big(\pi(\cdot| Y')\|\pi(\cdot| Y)\big)
&= \log\mathbb{E}_{\theta \sim \pi(\cdot|Y')}\left[\theta\right]+ 
\frac{1}{\alpha-1}\log \mathbb{E}_{\theta\sim\pi(\cdot| Y')}\!\left[\theta^{-(\alpha-1)}\right],\notag\\
D_{\alpha, \text{unbounded}}^{(0)}\big(\pi(\cdot| Y')\|\pi(\cdot| Y)\big)
&= \log\mathbb{E}_{\theta \sim \pi(\cdot|Y')}\left[1-\theta\right]+ 
\frac{1}{\alpha-1}\log \mathbb{E}_{\theta\sim\pi(\cdot| Y')}\!\left[(1-\theta)^{-(\alpha-1)}\right].\notag
\end{align}
Without additional structure on the posterior, the worst-case R\'enyi divergence over unbounded neighboring is given by
\begin{equation}
    D_{\alpha, \text{unbounded}}\le\max_{k\in\{0,1\}}\max\Big\{D_{\alpha, \text{unbounded}}^{(k)}\big(\pi(\cdot| Y)\|\pi(\cdot| Y')\big),\,D_{\alpha, \text{unbounded}}^{(k)}\big(\pi(\cdot| Y')\|\pi(\cdot| Y)\big)\Big\}\notag
\end{equation}
\textbf{Special case:} If $\theta\in [\eta, 1-\eta]$ for some $\eta \in (0,1/2)$. $|c_1(\theta)|\le \max\{|\log\eta|, |\log(1-\eta)|\} = -\log\eta$ uniformly. In the reverse direction, $\tilde c_1(\theta)=-c_1(\theta)$ satisfies the same bound, so Corollary~\ref{coro:uniform} applies in both directions.

\paragraph{Beta prior.} If $\pi(\theta) = \text{Beta}(a, b)$, and define $s = \sum_{i=1}^n y_i$,  then $\pi(\theta|Y) = \text{Beta}(A, B)$, where $A = a+s$, $B = b+n-s$. Then \\
If $y_n =1$ and  $A > 1$,
\begin{align}
    &\mathbb{E}_{\theta\sim\pi(\cdot| Y)}\!\left[\theta^{-1}\right]
    =\frac{A+B-1}{A-1}; \\
    &\mathbb{E}_{\theta\sim\pi(\cdot| Y)}\!\left[\theta^{\alpha-1}\right]
    =
    \frac{\text{Beta}(A-1+\alpha,B)}{\text{Beta}(A,B)}= \frac{\Gamma(A+\alpha-1)\Gamma(A+B)}{\Gamma(A)\,\Gamma(A+B+\alpha-1)}\\
    \Rightarrow \; &D_{\alpha, \text{unbounded}}^{(1)}\big(\pi(\cdot| Y)\|\pi(\cdot| Y')\big)
= \log\left(\frac{A+B-1}{A-1}\right)+ 
\frac{1}{\alpha-1}\log \left(\frac{\Gamma(A+\alpha-1)\Gamma(A+B)}{\Gamma(A)\,\Gamma(A+B+\alpha-1)}\right).
\end{align}
If $y_n=0$, and $B>1$,
\begin{align}
    &\mathbb{E}_{\theta\sim\pi(\cdot| Y)}\!\left[(1-\theta)^{-1}\right]
    =\frac{A+B-1}{B-1};\\
    &\mathbb{E}_{\theta\sim\pi(\cdot| Y)}\!\left[\left(1-\theta\right)^{\alpha-1}\right]
    =
    \frac{\text{Beta}(A,B-1+\alpha)}{\text{Beta}(A,B)}=\frac{\Gamma(A+B)\Gamma(B+\alpha-1)}{\Gamma(A+B+\alpha-1)\Gamma(B)}\\
    \Rightarrow \; &D_{\alpha, \text{unbounded}}^{(0)}\big(\pi(\cdot| Y)\|\pi(\cdot| Y')\big)
    = \log\left(\frac{A+B-1}{B-1}\right)+ 
    \frac{1}{\alpha-1}\log \left(\frac{\Gamma(B+\alpha-1)\Gamma(A+B)}{\Gamma(B)\,\Gamma(A+B+\alpha-1)}\right).
\end{align}
For the reverse direction, if $y_n=1$, then $\pi(\theta|Y')=\text{Beta}(A-1,B)$ and if $1<\alpha<A$,
\begin{align}
D_{\alpha, \text{unbounded}}^{(1)}\big(\pi(\cdot| Y')\|\pi(\cdot| Y)\big)
= \log\left(\frac{A-1}{A+B-1}\right)+ 
\frac{1}{\alpha-1}\log \left(\frac{\Gamma(A-\alpha)\Gamma(A+B-1)}{\Gamma(A-1)\,\Gamma(A+B-\alpha)}\right);
\end{align}
if $y_n=0$, then $\pi(\theta|Y')=\text{Beta}(A,B-1)$ and if $1<\alpha<B$.
\begin{align}
D_{\alpha, \text{unbounded}}^{(0)}\big(\pi(\cdot| Y')\|\pi(\cdot| Y)\big)
= \log\left(\frac{B-1}{A+B-1}\right)+ 
\frac{1}{\alpha-1}\log \left(\frac{\Gamma(B-\alpha)\Gamma(A+B-1)}{\Gamma(B-1)\,\Gamma(A+B-\alpha)}\right).
\end{align}
Both equal $\infty$ when $\alpha$ exceeds the stated range.
\begin{remark}
    The forward direction under unbounded neighboring imposes no upper bound on the R\'enyi order $\alpha$, but the reverse direction is finite only for $\alpha<A$ (if $y_n=1$) or $\alpha<B$ (if $y_n=0$). Since $A=a+s\ge a+1$ when $y_n=1$ and $B=b+n-s\ge b+1$ when $y_n=0$, with equality attainable, the two-sided R\'enyi divergence is finite uniformly over all unbounded neighboring pairs if and only if $\alpha<1+\min\{a,b\}$. In particular, OPS with a Beta prior does not satisfy pure DP under unbounded neighboring.
\end{remark}

\end{proof}

\section{Proof of Proposition~\ref{thm:GLM} for Generalized Linear Models}\label{ape:GLM}
\begin{proof}
We consider generalized linear models (GLMs) with canonical link. Each record is $(\x_i,y_i)$, where $\x_i\in\mathbb R^d$ and
$y_i\in\mathcal Y\subseteq\mathbb R$.
The dataset is $\Y=\{(\x_i,y_i)\}_{i=1}^n$, and the model parameter is
$\bs\beta\in\mathbb R^d$. Conditioned on $\x_i$ and $\bs\beta$, the response $y_i$ follows an exponential family distribution as below:
\begin{equation}
p(y_i|\x_i,\bs\beta)=
\exp\!\left(y_i\,\x_i^\top\bs\beta-\psi(\x_i^\top\bs\beta)+
h(y_i)
\right),
\end{equation}
where $\psi(\cdot)$ is the log-partition function and $h(\cdot)$ does not depend on $\bs\beta$. Then the log-likelihood is
\begin{align}
    &\log p(y_i|\x_i,\bs\beta)=y_i\x_i^\top\bs\beta-\psi(\x_i^\top\bs\beta)+h(y_i)\\
    \Rightarrow\; & \ell(\bs\beta|\Y)=\sum_{i=1}^n \log p(y_i|\x_i,\bs\beta)=\sum_{i=1}^n\left(y_i\,\x_i^\top\bs\beta-\psi(\x_i^\top\bs\beta)\right)+\sum_{i=1}^n h(y_i). 
\end{align}
Suppose there exists a finite constant $L>0$ such that for every \((\x_i,y_i)\) and every \(\bs\beta\), $\left\|\nabla_{\bs\beta}\log p(y_i| \x_i,\bs\beta)\right\|\le L$. If imposing a Gaussian prior on $\bs\beta$, i.e. $\pi(\bs\beta)= \mathcal N(0,\tau^2 \mathbf{I}_d)$, then the posterior distribution can be derived as
\begin{equation}
\pi(\bs\beta|\Y)\propto\exp\!\left(\sum_{i=1}^n\bigl(
y_i\,\x_i^\top\bs\beta-\psi(\x_i^\top\bs\beta)\bigr)-
\frac{\|\bs\beta\|_2^2}{2\tau^2}\right).
\end{equation}

\subsection{bounded neighboring}
Let $\Y$ and $\Y'$ differ in exactly one record, i.e. $\z_n=(\x_n,y_n)$ in $\Y$ is replaced by $\z'_n=(\x'_n,y'_n)$ in $\Y'$, then
\begin{align}
    c_1(\bs\beta)=\log\frac{\ell(\bs\beta;\Y)}{\ell(\bs\beta;\Y')}
    &=\log p(y_n|\x_n,\bs\beta) - \log p(y'_n|\x'_n,\bs\beta)\notag\\
    &= y_n\x_n^\top\bs\beta-\psi(\x_n^\top\bs\beta)+h(y_n) - \left(y'_n\x_n'^{\top}\bs\beta-\psi(\x_n'^{\top}\bs\beta)+h(y'_n)\right)\notag\\
    &= y_n\x_n^\top\bs\beta-y'_n\x_n'^{\top}\bs\beta-\left(\psi(\x_n^\top\bs\beta)-\psi(\x_n'^{\top}\bs\beta)\right)+h(y_n) - h(y'_n)\label{eqn:c1_GLM_bounded}
\end{align}
First, we prove $c_1(\bs\beta)$ is Lipschitz.
\begin{align}
    \nabla_{\bs\beta} c_1(\bs\beta)
    &= y_n \x_n - y_n' \x_n' - \left(\psi'(\x_n^\top\bs\beta)\x_n - \psi'(\x_n'^\top\bs\beta)\x'_n\right)\\
    \Rightarrow \|\nabla_{\bs\beta} c_1(\bs\beta)\|
    &= \Big\|y_n \x_n - y_n' \x_n' - \left(\psi'(\x_n^\top\bs\beta)\x_n - \psi'(\x_n'^\top\bs\beta)\x'_n\right)\Big\|\notag\\
    &= \Big\|\left(y_n-\psi'(\x_n^\top\bs\beta)\right) \x_n - \left( y_n' - \psi'(\x_n'^\top\bs\beta)\right) \x_n' \Big\|\notag\\
    & \le |y_n-\psi'(\x_n^\top\bs\beta)|\|\x_n\| + |y_n' - \psi'(\x_n'^\top\bs\beta)| \|\x_n'\|\le 2L
\end{align}
Hence, $c_1(\bs\beta)$ is Lipschitz continuous with constant $2L$. Then, we continue to prove the log-concavity of the posterior $\pi(\bs\beta|\Y)$.
\begin{align}
    \log \pi(\bs\beta|\Y) &= \sum_{i=1}^n\bigl(
    y_i\,\x_i^\top\bs\beta-\psi(\x_i^\top\bs\beta)\bigr)-
    \frac{\|\bs\beta\|_2^2}{2\tau^2} + \text{constant}\\
    \Rightarrow -\nabla^2_{\bs\beta} \log \pi(\bs\beta|\Y) &=\sum_{i=1}^n \psi''(\x_i^\top\bs\beta)\x_i \x_i^\top
    +\frac{1}{\tau^2}\mathbf{I}_d \succeq \frac{1}{\tau^2}\mathbf{I}_d \quad \text{per convexity of $\psi(\cdot)$}.
\end{align}
So $\pi(\bs\beta|\Y)$ is $m$-strongly log-concave with $m=\tau^{-2}$.

\begin{lemma}[Log-Sobolev Inequality \citep{ledoux2001concentration}]
Let $\pi$ be a probability measure on $\mathbb R^d$ with density
$\pi(x)\propto e^{-U(x)}$, where $U$ is twice continuously differentiable and
$m$-strongly convex, i.e.,
$\nabla^2 U(x)\succeq m I_d$ for all $x$.
Then $\pi$ satisfies a logarithmic Sobolev inequality (LSI) with constant $1/m$:
\[
\mathrm{Ent}_\pi(g^2)\le\frac{2}{m}\,\mathbb E_\pi\!\left[\|\nabla g\|^2\right],
\quad
\text{for all smooth } g.
\]
\end{lemma}

\begin{lemma}[Herbst's Theorem \citep{ledoux2001concentration}]
Let $\pi$ be a probability measure satisfying a LSI
with constant $C_{\mathrm{LSI}}$.
Then for any $L$-Lipschitz function $f$,
\[
\log \mathbb E_\pi\!\left[
\exp\!\big(t(f-\mathbb E_\pi f)\big)
\right]\le
\frac{C_{\mathrm{LSI}}\,L^2\,t^2}{2},
\quad
\forall t\in\mathbb R.
\]
In particular, $f$ is sub-Gaussian under $\pi$ with variance proxy
$C_{\mathrm{LSI}}L^2$.
\end{lemma}
Per above lemmas, $c_1(\bs\beta)$ is sub-Gaussian under $ \pi(\cdot|\Y)$ with
\begin{align}
\sigma_{\Y,\Y'}^2\le \frac{(2L)^2}{m} = 4L^2\tau^2.
\end{align}
The same argument applies to the reversed pair $(\Y',\Y)$: $\tilde c_1(\bs\beta)=-c_1(\bs\beta)$ has the same Lipschitz constant $2L$, and $\pi(\bs\beta|\Y')$ is also $\tau^{-2}$-strongly log-concave since $\Y'$ also has $n$ records. Hence $\sigma^2_{\Y',\Y}\le4L^2\tau^2$, and $\bar\sigma^2\le4L^2\tau^2$.
Plug into Corollary~\ref{coro:subgaussian}
\begin{align}
\max\{D_{\alpha}(\pi(\bs\beta|\Y)\|\pi(\bs\beta|\Y')),\,D_{\alpha}(\pi(\bs\beta|\Y')\|\pi(\bs\beta|\Y))\} \le \frac{\alpha\bar{\sigma}^2}{2}\le 2\alpha L^2\tau^2
\end{align}

\subsection{unbounded neighboring}
Let $\Y$ has size $n$ and $\Y'$ is of size $n-1$ by removing the $n$-th record $\z_n=(\x_n,y_n)$ in $\Y$. Then
\begin{align}
    c_1(\bs\beta)=\log\frac{\ell(\bs\beta;\Y)}{\ell(\bs\beta;\Y')}
    &=\log p(y_n|\x_n,\bs\beta)= y_n\x_n^\top\bs\beta-\psi(\x_n^\top\bs\beta)+h(y_n).\label{eqn:c1_GLM_unbounded}
\end{align}
Similarly, we prove $c_1(\bs\beta)$ is Lipschitz.
\begin{align}
    \nabla_{\bs\beta} c_1(\bs\beta)
    &= y_n \x_n - \psi'(\x_n^\top\bs\beta)\x_n\notag\\
    \Rightarrow \|\nabla_{\bs\beta} c_1(\bs\beta)\|
    &= \Big\|\left(y_n - \psi'(\x_n^\top\bs\beta)\right)\x_n\Big\|\le L.
\end{align}
Combined with log-concavity, $c_1(\bs\beta)$ is sub-Gaussian under $ \pi(\cdot|\Y)$ with
\begin{align}
    \sigma_{\Y, \Y'}^2\le \frac{L^2}{m}
    \le L^2\tau^2.
\end{align}
For the reversed pair $(\Y',\Y)$, $\tilde c_1(\bs\beta)=-\log p(y_n|\x_n,\bs\beta)$ has the same gradient norm bound $L$, and
\begin{align}
-\nabla^2_{\bs\beta}\log\pi(\bs\beta|\Y')=\sum_{i=1}^{n-1}\psi''(\x_i^\top\bs\beta)\x_i\x_i^\top+\frac{1}{\tau^2}\mathbf I_d\succeq\frac{1}{\tau^2}\mathbf I_d,
\end{align}
where the empty sum equals $0$ when $n=1$. Hence $\pi(\bs\beta|\Y')$ is also $\tau^{-2}$-strongly log-concave, $\sigma^2_{\Y',\Y}\le L^2\tau^2$, and $\bar\sigma^2\le L^2\tau^2$.
Plug into Corollary~\ref{coro:subgaussian}
\begin{align}
    \max\{D_{\alpha}(\pi(\bs\beta|\Y)\|\pi(\bs\beta|\Y')),\,D_{\alpha}(\pi(\bs\beta|\Y')\|\pi(\bs\beta|\Y))\} &\le \frac{\alpha\bar{\sigma}^2}{2}\le \frac{\alpha L^2\tau^2}{2}.
\end{align}
\end{proof}
\subsection{Proof of Example~\ref{eg:logistic} for Logistic Regression}
\subsubsection{bounded neighboring}
\begin{proof}
For logistic regression we have $\psi(t) = \log(1+e^t)$, and $h(\cdot)=0$, plug into Eq.~\eqref{eqn:c1_GLM_bounded}:
\begin{align}
    c_1(\bs\beta)&= y_n\x_n^\top\bs\beta-y'_n\x_n'^{\top}\bs\beta-\left(\log(1+e^{\x_n^\top\bs\beta})-\log(1+e^{\x_n'^\top\bs\beta})\right)
\end{align}
First, we prove $c_1(\bs\beta)$ is Lipschitz. Recall $\|\x\|\le R$ and let $\sigma(t) = \frac{1}{1+e^{-t}}$, given $|\sigma(\cdot)| < 1$ and $\frac{d \log(1+e^t)}{dt} = \sigma(t)$, we can have
\begin{align}
    \nabla_{\bs\beta} c_1(\bs\beta)
    &= y_n \x_n - y_n' \x_n'
    -\Big(\sigma(\x_n^\top\bs\beta)\x_n - \sigma(\x_n'^\top\bs\beta)\x_n'\Big)\notag\\
    & = \begin{cases}
    (y_n-y'_n)\x_n&\text{ if }\x_n = \x'_n \text{ and } y_n \ne y'_n\\
    y_n \x_n - y_n \x_n'
    -\Big(\sigma(\x_n^\top\bs\beta)\x_n - \sigma(\x_n'^\top\bs\beta)\x_n'\Big) & \text{ if }\x_n \ne \x'_n \text{ and } y_n = y'_n\\
    y_n \x_n - y_n' \x_n'
    -\Big(\sigma(\x_n^\top\bs\beta)\x_n - \sigma(\x_n'^\top\bs\beta)\x_n'\Big) & \text{ if }\x_n \ne \x'_n \text{ and } y_n \ne y'_n\\
    0  & \text{ if }\x_n = \x'_n \text{ and } y_n = y'_n\\
    \end{cases}.\notag
\end{align}
Therefore, when $\x_n = \x'_n$, the worst case is indeed flipping the label, then $\|\nabla_{\bs\beta}  c_1(\bs\beta)\| = \|(y_n-y'_n)\x_n\|\le \|\x_n\|\le R$. Otherwise,
\begin{align}
    \|\nabla_{\bs\beta}  c_1(\bs\beta)\|
    &\le \|y_n \x_n - y_n' \x_n'
    -\sigma(\x_n^\top\bs\beta)\x_n + \sigma(\x_n'^\top\bs\beta)\x_n'\|\notag\\
    &\le \|(y_n-\sigma(\x_n^\top\bs\beta)) \x_n - (y_n'-\sigma(\x_n'^\top\bs\beta)) \x_n'\|\notag\\
    &\le \|(y_n-\sigma(\x_n^\top\bs\beta)) \x_n\| +\| (y_n'-\sigma(\x_n'^\top\bs\beta)) \x_n'\|\notag\\
    &\le\underbrace{|y_n-\sigma(\x_n^\top\bs\beta)|}_{\le 1}\cdot\| \x_n\| +\underbrace{| y_n'-\sigma(\x_n'^\top\bs\beta)|}_{\le 1}\cdot\| \x_n'\|\le 2R.
\end{align}
Now we have proved that $c_1(\bs\beta)$ is Lipschitz continuous with constant $L=R$ or $L=2R$. Then, we similarly prove the log-concavity of the posterior $\pi(\cdot|\Y)$.
\begin{align}
    \log \pi(\bs\beta|\Y) &= \sum_{i=1}^n\bigl(
    y_i\,\x_i^\top\bs\beta-\log(1+e^{\x_i^\top\bs\beta})\bigr)-
    \frac{\|\bs\beta\|_2^2}{2\tau^2} + \text{constant}\notag\\
    \Rightarrow -\nabla^2_{\bs\beta} \log \pi(\bs\beta|\Y) &=\sum_{i=1}^n \sigma(\x_i^\top\bs\beta)(1-\sigma(\x_i^\top\bs\beta))\x_i \x_i^\top
    +\frac{1}{\tau^2}\mathbf{I}_d \succeq \frac{1}{\tau^2}\mathbf{I}_d \text{ for $\forall \bs\beta$ }.
\end{align}
So $\pi(\bs\beta|\Y)$ is $m$-strongly log-concave with $m=\tau^{-2}$. Therefore,
\begin{align}
    \sigma_{\Y,\Y'}^2\le \frac{L^2}{m}\le
    \begin{cases}
        R^2\tau^2&\text{ if }\x_n=\x'_n\\
        (2R)^2\tau^2=4R^2\tau^2 &\text{ if } \x_n \ne \x'_n
    \end{cases}.
\end{align}
For the reversed pair $(\Y',\Y)$, the Lipschitz constants $R$ (if $\x_n=\x'_n$) and $2R$ (if $\x_n\ne\x'_n$) are symmetric in the two records, and $\pi(\bs\beta|\Y')$ is also $\tau^{-2}$-strongly log-concave, so the same bounds hold for $\sigma^2_{\Y',\Y}$.
Plug into Corollary~\ref{coro:subgaussian}
\begin{align}
    \max\{D_{\alpha}(\pi(\bs\beta|\Y)\|\pi(\bs\beta|\Y')),\,D_{\alpha}(\pi(\bs\beta|\Y')\|\pi(\bs\beta|\Y))\}&\le \frac{\alpha\bar{\sigma}^2}{2}\le   \begin{cases}
        \frac{1}{2}\alpha R^2\tau^2&\text{ if }\x_n=\x'_n\\
        2\alpha R^2\tau^2 &\text{ if } \x_n \ne \x'_n
    \end{cases}.
\end{align}
where $\bar\sigma^2$ is taken over the corresponding class of neighboring pairs.

For comparison, Theorem 16 in \citep{geumlek2017renyi} gives $(\alpha,2\alpha R^2\tau^2)$-RDP for the replacement case with $\x_n=\x'_n$; our bound for this case is $\alpha R^2\tau^2/2$.
\end{proof}

\subsubsection{unbounded neighboring}
\begin{proof}
Now we have $\psi(t) = \log(1+e^t)$, and $h(\cdot)=0$.
Similarly, we prove $c_1(\bs\beta)$ is Lipschitz first as what we did in bounded neighboring. Recall $\|\x\|\le R$ and let $\sigma(t) = \frac{1}{1+e^{-t}}$, given $y\in\{0,1\}, |\sigma(t)| < 1$ for $\forall t \in \mathbb R$ and $\frac{d \log(1+e^t)}{dt} = \sigma(t)$, we have
\begin{align}
    &\nabla_{\bs\beta} c_1(\bs\beta)
    = y_n \x_n 
    -\sigma(\x_n^\top\bs\beta)\x_n\quad \text{ per Eq.~\eqref{eqn:c1_GLM_unbounded}}\notag\\
    \Rightarrow\;& \|\nabla_{\bs\beta}  c_1(\bs\beta)\|= \|(y_n-\sigma(\x_n^\top\bs\beta))\x_n\| \le \underbrace{|y_n-\sigma(\x_n^\top\bs\beta)|}_{\le 1}\|\x_n\|\le R
\end{align}
Now we have proved that $c_1(\bs\beta)$ is Lipschitz continuous with constant $L=R$. Given $\pi(\bs\beta|\Y)$ is $m$-strongly log-concave with $m=\tau^{-2}$ and the Lemmas, $c_1(\bs\beta)$ is sub-Gaussian under $ \pi(\cdot|\Y)$ with
\begin{align}
    \sigma_{\Y, \Y'}^2\le \frac{L^2}{m}\le R^2\tau^2
\end{align}
For the reversed pair $(\Y',\Y)$, $\|\nabla_{\bs\beta}\tilde c_1(\bs\beta)\|=|y_n-\sigma(\x_n^\top\bs\beta)|\,\|\x_n\|\le R$, and $\pi(\bs\beta|\Y')$ is $\tau^{-2}$-strongly log-concave by the argument for general GLMs above. Hence $\sigma^2_{\Y',\Y}\le R^2\tau^2$ and $\bar\sigma^2\le R^2\tau^2$.
Plug into Corollary~\ref{coro:subgaussian}
\begin{align}
\max\{D_{\alpha}(\pi(\bs\beta|\Y)\|\pi(\bs\beta|\Y')),\,D_{\alpha}(\pi(\bs\beta|\Y')\|\pi(\bs\beta|\Y))\}&\le \frac{\alpha\bar{\sigma}^2}{2}\le \frac{\alpha R^2\tau^2}{2}
\end{align}
\end{proof}

\section{Proof of Proposition~\ref{thm:GG} for Gaussian-Gaussian conjugate pair}\label{ape:GG_known}
\begin{proof}
Assume observed dataset $Y = \{y_i\}_{i=1}^n$, where $y_i|\theta \overset{\text{i.i.d}}{\sim}\mathcal{N}(\theta, \sigma^2)$ and $\sigma^2$ is known. If imposing a Gaussian prior on $\theta$, i.e. $\pi(\theta)= \mathcal N(\mu_0,\tau^2)$, then per conjugacy, the posterior $\theta|Y \sim \mathcal{N}(\mu_Y, v_Y^2)$, where 
\begin{align}\label{eqn:GG_post_para}
v^2_Y=\left(\frac{n}{\sigma^2}+\frac{1}{\tau^2}\right)^{-1},
\qquad
\mu_Y=v^2_Y\left(\frac{1}{\sigma^2}\sum_{i=1}^n y_i+\frac{\mu_0}{\tau^2}\right).
\end{align}

\subsection{bounded neighboring}
WLOG, suppose neighboring datasets $Y,Y'$, both of size $n$, differ in their $n$-th record $y_n$ and $y'_n$. Then the log-likelihood ratio $c_1(\theta)$ can be derived as
\begin{align}
    \ell(\theta;Y)&=(2\pi\sigma^2)^{-n/2}\exp\left({-\frac{(y_n-\theta)^2}{2\sigma^2}-\sum_{i=1}^{n-1}\frac{(y_i-\theta)^2}{2\sigma^2}}\right);\\
    \ell(\theta;Y')&=(2\pi\sigma^2)^{-n/2}\exp\left({-\frac{(y'_n-\theta)^2}{2\sigma^2}-\sum_{i=1}^{n-1}\frac{(y_i-\theta)^2}{2\sigma^2}}\right)\\
    \Rightarrow \;c_1(\theta)&=\log\frac{\ell(\theta;Y)}{\ell(\theta;Y')}
    =\frac{(y'_n-\theta)^2-(y_n-\theta)^2}{2\sigma^2}\notag\\
    &= \frac{(y_n'^2-y_n^2)-2\theta(y_n'-y_n)}{2\sigma^2} \triangleq K+\frac{\Delta y}{\sigma^2}\theta,
\end{align}
where $K = \frac{(y_n'^2-y_n^2)}{2\sigma^2}$ and $\Delta y=y_n-y'_n$. Per the affine invariance of Gaussian distributions
\begin{align}
    c_1(\theta)\sim \mathcal{N}\left(K + \frac{\Delta y}{\sigma^2}\mu_Y, \left(\frac{\Delta y}{\sigma^2}\right)^2 v^2_Y\right).
\end{align}
Since the moment generating function (MGF) for $\mathcal{N}(\mu, \sigma^2)$ is $\exp({t\mu +\sigma ^{2}t^{2}/2})$,
\begin{align}
    &\log\mathbb{E}_{\theta\sim \pi(\theta|Y)}\left(e^{t(c_1(\theta)-\mathbb{E}_{\theta\sim \pi(\theta|Y)}\left[c_1(\theta)\right])}\right)\notag\\
    =\;&
    \frac{t^2 \left(\Delta y/\sigma^2\right)^2 v^2_Y}{2} = \frac{t^2}{2}\left(\frac{(y_n-y'_n)^2}{\sigma^4}\left(\frac{n}{\sigma^2}+\frac{1}{\tau^2}\right)^{-1}\right)\triangleq \frac{t^2\sigma_{Y, Y'}^2}{2}.
\end{align}
Then we conclude that $c_1(\theta)$ is sub-Gaussian with $\sigma^2_{Y, Y'} = \frac{(y_n-y'_n)^2}{\sigma^4}\left(\frac{n}{\sigma^2}+\frac{1}{\tau^2}\right)^{-1}$.

To obtain $\bar{\sigma}^{2} = \sup_{(Y,Y'): d(Y, Y')=1}\sigma_{Y,Y'}^2$, we need to assume a bounded data domain, i.e. $\exists\, c\in\mathbb R, k >0$ such that $y\in[c-k\sigma,\,c+k\sigma]$ for every record, then $|y_n-y'_n|\le 2k\sigma$, then
\begin{align}
    \sigma^2_{Y, Y'}\le \frac{(2k\sigma)^2}{\sigma^4}\left(\frac{n}{\sigma^2}+\frac{1}{\tau^2}\right)^{-1} = \frac{4k^2}{n+\frac{\sigma^2}{\tau^2}} = \bar{\sigma}^{2}.\label{eqn:GG_bounded_sigma_bar}
\end{align}
For the reversed pair $(Y',Y)$, $\widetilde c_1(\theta)=-c_1(\theta)$ is Gaussian under $\pi(\theta|Y')$ with variance $(y_n-y'_n)^2v_{Y'}^2/\sigma^4$; since $Y'$ also has $n$ records, $v_{Y'}^2=v_Y^2$, so the same bound holds for $\sigma^2_{Y',Y}$.
Plug Eq.~\eqref{eqn:GG_bounded_sigma_bar} into Corollary~\ref{coro:subgaussian},
\begin{align}
\max\{D_{\alpha}(\pi(\theta|Y)\|\pi(\theta|Y')),D_{\alpha}(\pi(\theta|Y')\|\pi(\theta|Y))\} \le \frac{\alpha\bar{\sigma}^2}{2}\le \frac{2k^2\alpha}{n+\frac{\sigma^2}{\tau^2}}.
\end{align}

\subsection{unbounded neighboring}
\paragraph{Forward direction}Let $Y$ has size $n$ and $Y'$ is of size $n-1$ by removing the $n$-th record $y_n$ in $Y$. Then
\begin{align}
    c_1(\theta)=\log\frac{\ell(\theta;Y)}{\ell(\theta;Y')}
    &=\frac{-(y_n-\theta)^2}{2\sigma^2}-\frac{1}{2}\log(2\pi\sigma^2) \triangleq \frac{-Z^2}{2\sigma^2}-\frac{1}{2}\log(2\pi\sigma^2).\label{eqn:GG_unbounded_c1}
\end{align}
where $Z = \theta-y_n$. Under the posterior $\theta|Y \sim \mathcal{N}(\mu_Y, v_Y^2)$, then $Z|Y\sim \mathcal{N}(\mu_Y-y_n, v_Y^2)$, and $Z^2/v^2_Y$ follows a noncentral chi-square distribution with degree of freedom 1, that is
\begin{align}
    \frac{Z^2}{v^2_Y}\sim \chi^2_k(\lambda), \quad \text{with }k=1, \lambda = \frac{(\mu_Y-y_n)^2}{v^2_Y}.
\end{align}
We note that $c_1(\theta)$ in Eq.~\eqref{eqn:GG_unbounded_c1} doesn't fall into the sub-Gaussian regime because $c_1(\theta)$ is a negative quadratic function of a Gaussian random variable, so its exponential moment $\mathbb{E}_P[e^{t c_1(\theta)}]$ is finite only for $|t|$ in a bounded neighborhood of zero (due to the noncentral $\chi^2$ tail of $Z^2$), rather than for all $t\in\mathbb{R}$ as required by the sub-Gaussian condition. So there is no finite $\sigma^2_{Y, Y'}$.

But we can still leverage properties of $\chi^2$ distribution to explicitly calculate each term in Eq.~\eqref{eqn:D_alpha_general_c2}. First, the expectation of $Z^2$ is 
\begin{align}
    \mathbb{E}_{\theta\sim \pi(\theta|Y)}\left[Z^2\right] & = v^2_Y(k+\lambda) = v^2_Y\left(1+\frac{(\mu_Y-y_n)^2}{v^2_Y}\right) = v^2_Y+(\mu_Y-y_n)^2.\label{eqn:GG_unbounded_EZ2}
\end{align}
Then, recall the moment generating function of $\chi^2_k(\lambda)$ is $e^{\lambda t/(1-2t)}{\left(1-2t\right)}^{-k/2}$ for $t<1/2$, hence
\begin{align}
\mathbb E_{\theta\sim \pi(\theta|Y)}\big[e^{tZ^2/v^2_Y}\big]
&=\exp\left({\frac{(\mu_Y-y_n)^2}{v^2_Y}\cdot\frac{t}{1-2t}}\right){\left(1-2t\right)}^{-1/2}\notag\\
\Rightarrow\; \mathbb E_{\theta\sim \pi(\theta|Y)}\big[e^{tZ^2}\big]
&=\exp\left({\frac{(\mu_Y-y_n)^2}{v^2_Y}\cdot\frac{v^2_Yt}{1-2v^2_Yt}}\right){\left(1-2tv^2_Y\right)}^{-1/2}\notag\\
&=(1-2v_Y^2 t)^{-1/2}
\exp\left(\frac{(\mu_Y-y_n)^2t}{1-2v_Y^2 t}\right),
\qquad t<\frac{1}{2v_Y^2}.\label{eqn:GG_unbounded_mgf_Z2}
\end{align}
Now we have two approaches to explicitly upper bound $D_{\alpha}(\pi(\theta|Y)\|\pi(\theta|Y'))$. 

\noindent 1) Derive sub-exponential parameters $b_Y, \nu_Y$ for $c_1(\theta)$ and apply Corollary~\ref{coro:subexponential}. Since the derivation of $b_Y, \nu_Y$ is not tight, the resulting RDP bound is generally not tight; 

\noindent 2) Directly calculate each term in Eq.~\eqref{eqn:D_alpha_general_c2_ape} using the closed-form moment generating function in Eq.~\eqref{eqn:GG_unbounded_mgf_Z2}. This approach yields an exact expression.

\textbf{Method 1: sub-exponential parameters.\footnote{For brevity, we write $b_Y,\nu_Y$ for $b_{Y,Y'},\nu_{Y,Y'}$, and only the forward direction is illustrated here.}}
Our target is to identify $b_Y, \nu^2_Y$ such that
\begin{align}
    \log\mathbb{E}_{\theta\sim P}\left(e^{t(c_1(\theta)-\mathbb E_P\left[c_1(\theta)\right])}\right)\le \frac{t^2\nu^2_{Y}}{2(1-b_{Y}|t|)}
\end{align}
for $|t| < 1/b_Y$, with $c_1(\theta)$ defined in Eq.~\eqref{eqn:GG_unbounded_c1}.
\begin{align}
    c_1(\theta)-\mathbb E_P\left[c_1(\theta)\right] & = \frac{-Z^2}{2\sigma^2}-\frac{1}{2}\log(2\pi\sigma^2) - \mathbb E_P\left(\frac{-Z^2}{2\sigma^2}-\frac{1}{2}\log(2\pi\sigma^2)\right)\notag\\
    & = -\frac{1}{2\sigma^2}\left(Z^2-\mathbb E_P(Z^2)\right)
\end{align}
So we only need to examine $(Z^2-\mathbb E_P(Z^2))$:
\begin{align}
    &\log \mathbb{E}_P\left[e^{t(Z^2-\mathbb E_P(Z^2))}\right]\notag\\
    =\; & \log \mathbb{E}_P\left[e^{t(Z^2-v^2_Y-(\mu_Y-y_n)^2)}\right]\qquad \text{ per Eq.~\eqref{eqn:GG_unbounded_EZ2}}\notag\\
    =\; & \log \mathbb{E}_P\left[e^{tZ^2}\right]-\left(v^2_Y+(\mu_Y-y_n)^2\right)t \notag\\
    =\; & -\frac{1}{2}\log(1-2v_Y^2 t)-v^2_Y t+(\mu_Y-y_n)^2t\left(\frac{1}{1-2v_Y^2 t}-1\right)\qquad \text{ per Eq.~\eqref{eqn:GG_unbounded_mgf_Z2}}\notag\\
    =\; &  \frac{1}{2}\left[-\log(1-x)-x\right]+(\mu_Y-y_n)^2t\left(\frac{1}{1-x}-1\right) \qquad \text{let $x = 2v_Y^2t$}.\label{eqn:GG_unbounded_subexp}
\end{align}
Since Eq.~\eqref{eqn:GG_unbounded_mgf_Z2} requires $t<1/(2v_Y^2)$,  $x = 2v_Y^2t < 1$. To ensure Eq.~\eqref{eqn:GG_unbounded_subexp}$\le \frac{t^2\nu'^2_{Y}}{2(1-b'_{Y}|t|)}$ for some $b'_Y, \nu'_Y$ when $|t|<1/b'_Y$, we need some inequalities below for $|x|<1$
\begin{align}
-\log(1-x)-x &\le \frac{x^2}{2(1-|x|)}, \qquad 
\frac{1}{1-x}-1  = \frac{x}{1-x} \le \frac{|x|}{1-|x|}.
\end{align}
Plug the inequalities above into Eq.~\eqref{eqn:GG_unbounded_subexp}:
\begin{align}
    \text{Eq.~\eqref{eqn:GG_unbounded_subexp}}& \le \frac{1}{2}\cdot \frac{x^2}{2(1-|x|)} + (\mu_Y-y_n)^2|t|\cdot \frac{|x|}{1-|x|}\notag\\
    & = \frac{1}{2}\cdot \frac{4v_Y^4t^2}{2(1-2v_Y^2|t|)} + (\mu_Y-y_n)^2|t|\cdot \frac{2v_Y^2|t|}{1-2v_Y^2|t|}\notag\\
    & = \frac{v_Y^4t^2}{1-2v_Y^2|t|} + \frac{2v_Y^2(\mu_Y-y_n)^2t^2}{1-2v_Y^2|t|}\notag\\
    &= \frac{2\left(v_Y^2+2(\mu_Y-y_n)^2\right)v_Y^2t^2}{2(1-2v_Y^2|t|)}\triangleq  \frac{t^2\nu'^2_{Y}}{2(1-b'_{Y}|t|)}.
\end{align}
Therefore $b'_Y = 2v_Y^2$ and $\nu'^2_Y=2\left(v_Y^2+2(\mu_Y-y_n)^2\right)v_Y^2$. 

Recall that $b'_Y, \nu'^2_Y$ are for $(Z^2-\mathbb E_P(Z^2))$, while $c_1(\theta)-\mathbb E_P\left[c_1(\theta)\right]  = -\frac{1}{2\sigma^2}\left(Z^2-\mathbb E_P(Z^2)\right)$, hence $b_Y = b'_Y/2\sigma^2$ and $\nu^2_Y = \nu'^2_Y/4\sigma^4$. Plug in Corollary~\ref{coro:subexponential}:
\begin{align}
D_{\alpha}(P||Q)  &\le \frac{\nu_{Y}^2}{2}
    \frac{\alpha(1-2b_{Y})+2b_{Y}}{(1-b_{Y})\bigl(1-b_{Y}(\alpha-1)\bigr)}\notag\\
    & = \frac{ v_Y^2\bigl(v_Y^2+2(\mu_Y-y_n)^2\bigr)}{4\sigma^4}\frac{\alpha\!\left(1-\tfrac{2v_Y^2}{\sigma^2}\right)+\tfrac{2v_Y^2}{\sigma^2}}
    {\left(1-\tfrac{v_Y^2}{\sigma^2}\right)\left(1-\tfrac{v_Y^2}{\sigma^2}(\alpha-1)\right)}\notag\\
    & = \frac{ v_Y^2\bigl(v_Y^2+2(\mu_Y-y_n)^2\bigr)}{4\sigma^2}\cdot\frac{\alpha\sigma^2-2v_Y^2(\alpha-1)}
    {\left(\sigma^2-v^2_Y\right)\left(\sigma^2-v_Y^2(\alpha-1)\right)},
\end{align}
where $v^2_Y$ and $\mu_Y$ are defined in Eq.~\eqref{eqn:GG_post_para}.

\textbf{Method 2: exact moment calculation}
For the first term in Eq.~\eqref{eqn:D_alpha_general_c2_ape}, per the $c_1(\theta)$ defined in Eq.~\eqref{eqn:GG_unbounded_c1}, and the moment generating function of $Z^2$ derived in Eq.~\eqref{eqn:GG_unbounded_mgf_Z2}
\begin{align}
-c_1(\theta)&=\frac{Z^2}{2\sigma^2}+\frac12\log(2\pi\sigma^2)\notag\\
\mathbb E_P[e^{-c_1(\theta)}] & = \mathbb E_P\left[\exp\left(\frac{Z^2}{2\sigma^2}+\frac12\log(2\pi\sigma^2)\right)\right]\notag\\
&=(2\pi\sigma^2)^{\frac12}\mathbb E_P\left[\exp\left(\frac{Z^2}{2\sigma^2}\right)\right] \notag\\
&= (2\pi\sigma^2)^{\frac12}\left(1-2v_Y^2 \frac{1}{2\sigma^2}\right)^{-\frac12}
\exp\left(\frac{(\mu_Y-y_n)^2\frac{1}{2\sigma^2}}{1-2v_Y^2 \frac{1}{2\sigma^2}}\right)\text{ plug $t=\frac{1}{2\sigma^2}$ in Eq.~\eqref{eqn:GG_unbounded_mgf_Z2}}\notag\\
& = (2\pi\sigma^2)^{\frac12}\left(1- \frac{v_Y^2}{\sigma^2}\right)^{-\frac12}\exp\left(\frac{(\mu_Y-y_n)^2}{2(\sigma^2 -v^2_Y)}\right).
\end{align}
This requires $t=\frac{1}{2\sigma^2}<\frac{1}{2v_Y^2}$, equivalently $v_Y^2<\sigma^2$, recall $v^2_Y=\left(\frac{n}{\sigma^2}+\frac{1}{\tau^2}\right)^{-1}$, this holds for $n\ge1$ in the Gaussian-Gaussian conjugate model. Then  per Lemma~\ref{lemma:c2_in_c1}
\begin{align}
    c_2 &= \log \mathbb E_P[e^{-c_1(\theta)}] = \frac{1}{2}\log(2\pi\sigma^2)-\frac{1}{2}\log\left(1- \frac{v_Y^2}{\sigma^2}\right)+\frac{(\mu_Y-y_n)^2}{2(\sigma^2 -v^2_Y)}. \label{eqn:GG_unbounded_1st}
\end{align}
Next, for the second term in Eq.~\eqref{eqn:D_alpha_general_c2_ape}, we also use the moment generating function for $Z^2$ in  Eq.~\eqref{eqn:GG_unbounded_mgf_Z2} to exactly calculate the ($\alpha-1$) moment for $c_1(\theta)$:
\begin{align}
(\alpha-1)c_1(\theta)
&= -\frac{\alpha-1}{2\sigma^2}Z^2-\frac{\alpha-1}{2}\log(2\pi\sigma^2),\notag\\
\mathbb E_P[e^{(\alpha-1)c_1(\theta)}]  & = \mathbb E_P\left[\exp\left(-\frac{\alpha-1}{2\sigma^2}Z^2-\frac{\alpha-1}{2}\log(2\pi\sigma^2)\right)\right]\notag\\
&=(2\pi\sigma^2)^{-\frac{\alpha-1}{2}}\mathbb E_P\left[\exp\left(-\frac{\alpha-1}{2\sigma^2}Z^2\right)\right] \text{ plug $t=\frac{-(\alpha-1)}{2\sigma^2}$ in Eq.~\eqref{eqn:GG_unbounded_mgf_Z2}}\notag\\
& = (2\pi\sigma^2)^{-\frac{\alpha-1}{2}}\left(1-2v_Y^2 \left(\frac{1-\alpha}{2\sigma^2}\right)\right)^{-\frac{1}{2}}
\exp\left(\frac{(\mu_Y-y_n)^2\left(\frac{1-\alpha}{2\sigma^2}\right)}{1-2v_Y^2 \left(\frac{1-\alpha}{2\sigma^2}\right)}\right)\notag\\
& = (2\pi\sigma^2)^{-\frac{\alpha-1}{2}}\left(1+v_Y^2 \left(\frac{\alpha-1}{\sigma^2}\right)\right)^{-\frac{1}{2}}
\exp\left(\frac{(\mu_Y-y_n)^2(1-\alpha)}{2(\sigma^2+(\alpha-1)v_Y^2)}\right).
\end{align}
This requires $t=\frac{-(\alpha-1)}{2\sigma^2} < \frac{1}{2v^2_Y}$. Since $\alpha>1, \frac{-(\alpha-1)}{2\sigma^2} < 0$, so this requirement always holds.
Next, the 2nd term in Eq.~\eqref{eqn:D_alpha_general_c2_ape} can be calculated as
\begin{align}
    &\frac{1}{\alpha-1}\log\mathbb E_P\left[e^{(\alpha-1)c_1(\theta)}\right]\notag\\
    =\;&-\frac{1}{2}\log(2\pi\sigma^2)-\frac{1}{2(\alpha-1)}\log\left(1+v_Y^2 \left(\frac{\alpha-1}{\sigma^2}\right)\right)-\frac{(\mu_Y-y_n)^2}{2(\sigma^2+(\alpha-1)v_Y^2)}.\label{eqn:GG_unbounded_2nd}
\end{align}
Plug $c_2$ in Eq.~\eqref{eqn:GG_unbounded_1st} and $\frac{1}{\alpha-1}\log \mathbb{E}_{P}\left[e^{(\alpha-1)\cdot c_1(\theta)}\right]$ in Eq.~\eqref{eqn:GG_unbounded_2nd} into Eq.~\eqref{eqn:D_alpha_general_c2_ape}, we have
\begin{align}
    &D_{\alpha}(\pi(\theta|Y)||\pi(\theta|Y')) \notag\\
    =\; & c_2 +\frac{1}{\alpha-1}\log \mathbb{E}_{P}\left[e^{(\alpha-1)\cdot c_1(\theta)}\right]\notag\\
    =\;&\cancel{\frac{1}{2}\log(2\pi\sigma^2)}-\frac{1}{2}\log\left(1- \frac{v_Y^2}{\sigma^2}\right)+\frac{(\mu_Y-y_n)^2}{2(\sigma^2 -v^2_Y)} \notag\\
    &\cancel{-\frac{1}{2}\log(2\pi\sigma^2)}-\frac{1}{2(\alpha-1)}\log\left(1+v_Y^2 \left(\frac{\alpha-1}{\sigma^2}\right)\right)-\frac{(\mu_Y-y_n)^2}{2(\sigma^2+(\alpha-1)v_Y^2)}\notag\\
    =\;&-\frac{1}{2}\log\!\left(\!1- \frac{v_Y^2}{\sigma^2}\right)-\frac{1}{2(\alpha\!-\!1)}\log\!\left(\!1+v_Y^2 \left(\frac{\alpha-1}{\sigma^2}\right)\right)+\frac{(\mu_Y-y_n)^2}{2(\sigma^2 -v^2_Y)}-\frac{(\mu_Y-y_n)^2}{2(\sigma^2+(\alpha-1)v_Y^2)}.\notag
\end{align}
Recall
\begin{align}
v^2_Y=\left(\frac{n}{\sigma^2}+\frac{1}{\tau^2}\right)^{-1},
\qquad
\mu_Y=v^2_Y\left(\frac{1}{\sigma^2}\sum_{i=1}^n y_i+\frac{\mu_0}{\tau^2}\right).
\end{align}
Then define $\omega = n + \frac{\sigma^2}{\tau^2}$ and assume $y_n \in [\mu_Y-k^*\sigma, \mu_Y + k^*\sigma]$.
\begin{align}
    &D_{\alpha}(\pi(\theta|Y)||\pi(\theta|Y')) \notag\\
    =\;&-\frac{1}{2}\log\!\left(\!1- \frac{v_Y^2}{\sigma^2}\right)-\frac{1}{2(\alpha\!-\!1)}\log\!\left(\!1+v_Y^2 \left(\frac{\alpha-1}{\sigma^2}\right)\right)+\frac{(\mu_Y-y_n)^2}{2(\sigma^2 -v^2_Y)}-\frac{(\mu_Y-y_n)^2}{2(\sigma^2+(\alpha-1)v_Y^2)}\notag\\
    =\; & -\frac{1}{2}\log\left(1- \omega^{-1}\right)-\frac{1}{2(\alpha\!-\!1)}\log\!\left(\!1+\frac{\alpha-1}{\omega}\right) +\frac{(\mu_Y-y_n)^2}{2\sigma^2}\left[\frac{1}{1-\omega^{-1}}-\frac{1}{1+\frac{\alpha-1}{\omega}}\right]\notag\\
    \triangleq\; & A(\alpha, \omega) + \frac{(\mu_Y-y_n)^2}{2\sigma^2} B(\alpha, \omega)\notag\\
    \le\; & A(\alpha, \omega)+\frac{k^{*2}}{2}B(\alpha, \omega)\label{eqn:eps_GG_unbounded_ape}
\end{align}
Since $\mu_Y=\big(\frac{\sigma^2}{\tau^2}\mu_0+\sum_{i=1}^n y_i\big)/\omega$ is a convex combination of $\mu_0$ and $\{y_i\}$, if all records and $\mu_0$ lie in $[c-M,c+M]$, then $|\mu_Y-y_n|\le 2M$, i.e., $k^*=2M/\sigma$ is a data-independent choice.

\begin{claim}\label{claim:mono}
For all $\alpha>1$ and $\omega>1$, we have $A(\alpha,\omega)>0$ and $B(\alpha,\omega)>0$. Furthermore, holding $\alpha$ fixed, both $A(\alpha,\omega)$ and $B(\alpha,\omega)$ decrease monotonically as $\omega$ increases.
\end{claim}
\begin{proof}
    Assume $n \ge 1$, then $\omega = n + \sigma^2/\tau^2 > 1$. Let $x = \omega^{-1}$ and $t = \alpha-1$, then $x\in (0,1)$ and $t>0$, and
    \begin{align}
        A(\alpha, \omega) = -\frac{1}{2}\log\left(1- x\right)-\frac{1}{2t}\log\left(1+xt\right) \triangleq A(x).
    \end{align}
    Thus to prove $A(\alpha, \omega)>0$ is equivalent to prove 
    \begin{align}
        -\log\left(1- x\right)>\frac{1}{t}\log\left(1+xt\right).
    \end{align}
    Since $-\log(1-x) > x$ for $x\in (0,1)$, and $\log(1+xt) < xt\Rightarrow t^{-1}\log(1+xt) < x$,
    \begin{align}
        -\log\left(1- x\right)>x > \frac{1}{t}\log\left(1+xt\right),
    \end{align}
    which completes the proof of strictly positivity of $A(\alpha, \omega)$. Then, take derivative of $x$ on $A(x)$:
    \begin{align}
        A'(x) &= \frac{1}{2(1-x)} - \frac{1}{2(1+xt)} > 0,
    \end{align}
    per $x\in (0,1)$ and $t > 0$. Thus,
    \begin{align}
        \frac{\partial A}{\partial \omega} = A'(x)\frac{d x}{d \omega} = -\omega^2 A'(x) < 0.
    \end{align}
    Furthermore, take derivative of $x$ on $B(x) = \frac{1}{1-x}-\frac{1}{1+xt}$
    \begin{align}
        B'(x) = \frac{1}{(1-x)^2} + \frac{t}{(1+xt)^2} > 0 \quad \Rightarrow \frac{\partial B}{\partial \omega} < 0
    \end{align}
    thus, both $A(\alpha,\omega)$ and $B(\alpha,\omega)$ decrease monotonically as $\omega$ increases.
\end{proof}

\paragraph{Reverse direction}
We next analyze the reverse R\'enyi divergence $D_\alpha(\pi(\theta|Y')\|\pi(\theta|Y))$, where $Y$ has size $n$ and $Y'$ is of size $n-1$, obtained by removing the $n$-th record $y_n$ from $Y$. Define
\begin{align}
\widetilde c_1(\theta)&\triangleq\log\frac{\ell(\theta;Y')}{\ell(\theta;Y)}=-\log p(y_n|\theta)=\frac{(y_n-\theta)^2}{2\sigma^2}+\frac{1}{2}\log(2\pi\sigma^2)\triangleq\frac{Z'^2}{2\sigma^2}+\frac{1}{2}\log(2\pi\sigma^2)\notag\\
&=-c_1(\theta),\label{eqn:c1_GG_reverse}
\end{align}
where $Z'=\theta-y_n$. Under the posterior $\theta|Y'\sim\mathcal N(\mu_{Y'},v_{Y'}^2)$, $Z'|Y'\sim\mathcal N(\mu_{Y'}-y_n,v_{Y'}^2)$, and $Z'^2/v_{Y'}^2$ follows a noncentral chi-square distribution with one degree of freedom:
\begin{align}
\frac{Z'^2}{v_{Y'}^2}\sim\chi_1^2(\lambda'),\qquad\lambda'=\frac{(\mu_{Y'}-y_n)^2}{v_{Y'}^2}.\label{eqn:noncentral_chisq_reverse}
\end{align}
Recall that the MGF of $\chi_k^2(\lambda)$ is $\exp\big(\frac{\lambda t}{1-2t}\big)(1-2t)^{-k/2}$ for $t<\frac{1}{2}$. Therefore,
\begin{align}
\mathbb E_{\theta\sim\pi(\theta|Y')}\big[e^{tZ'^2}\big]=\left(1-2v_{Y'}^2t\right)^{-1/2}\exp\left(\frac{(\mu_{Y'}-y_n)^2t}{1-2v_{Y'}^2t}\right),\qquad t<\frac{1}{2v_{Y'}^2}.\label{eqn:MGF_Z2_reverse}
\end{align}
We now directly calculate the two terms in Eq.~\eqref{eqn:D_alpha_general_c2_ape} using the closed-form MGF in Eq.~\eqref{eqn:MGF_Z2_reverse}. For the first term, per Eq.~\eqref{eqn:c1_GG_reverse}, $-\widetilde c_1(\theta)=-\frac{Z'^2}{2\sigma^2}-\frac{1}{2}\log(2\pi\sigma^2)$, and hence
\begin{align}
\mathbb E_{\theta\sim\pi(\theta|Y')}\big[e^{-\widetilde c_1(\theta)}\big]&=(2\pi\sigma^2)^{-1/2}\,\mathbb E_{\theta\sim\pi(\theta|Y')}\Big[\exp\Big(-\frac{Z'^2}{2\sigma^2}\Big)\Big]\notag\\
&=(2\pi\sigma^2)^{-1/2}\Big(1+\frac{v_{Y'}^2}{\sigma^2}\Big)^{-1/2}\exp\Big(-\frac{(\mu_{Y'}-y_n)^2}{2(\sigma^2+v_{Y'}^2)}\Big),\label{eqn:first_moment_reverse}
\end{align}
where we plug $t=-1/(2\sigma^2)$ into Eq.~\eqref{eqn:MGF_Z2_reverse}. Since $t<0$, the MGF condition always holds. Therefore, per Lemma~\ref{lemma:c2_in_c1},
\begin{align}
\widetilde c_2=\log\mathbb E_{\theta\sim\pi(\theta|Y')}\big[e^{-\widetilde c_1(\theta)}\big]=-\frac{1}{2}\log(2\pi\sigma^2)-\frac{1}{2}\log\Big(1+\frac{v_{Y'}^2}{\sigma^2}\Big)-\frac{(\mu_{Y'}-y_n)^2}{2(\sigma^2+v_{Y'}^2)}.\label{eqn:c2_reverse}
\end{align}
Next, for the second term in Eq.~\eqref{eqn:D_alpha_general_c2_ape}, $(\alpha-1)\widetilde c_1(\theta)=\frac{\alpha-1}{2\sigma^2}Z'^2+\frac{\alpha-1}{2}\log(2\pi\sigma^2)$. Thus,
\begin{align}
\mathbb E_{\theta\sim\pi(\theta|Y')}\big[e^{(\alpha-1)\widetilde c_1(\theta)}\big]&=(2\pi\sigma^2)^{(\alpha-1)/2}\,\mathbb E_{\theta\sim\pi(\theta|Y')}\Big[\exp\Big(\frac{\alpha-1}{2\sigma^2}Z'^2\Big)\Big]\notag\\
&=(2\pi\sigma^2)^{(\alpha-1)/2}\Big(1-\frac{(\alpha-1)v_{Y'}^2}{\sigma^2}\Big)^{-1/2}\exp\Big(\frac{(\alpha-1)(\mu_{Y'}-y_n)^2}{2(\sigma^2-(\alpha-1)v_{Y'}^2)}\Big),\label{eqn:second_moment_reverse}
\end{align}
where we plug $t=\frac{\alpha-1}{2\sigma^2}$ into Eq.~\eqref{eqn:MGF_Z2_reverse}. This requires
\begin{align}
\frac{\alpha-1}{2\sigma^2}<\frac{1}{2v_{Y'}^2},\quad\text{or equivalently}\quad(\alpha-1)v_{Y'}^2<\sigma^2.\label{eqn:reverse_finiteness_v}
\end{align}
Consequently,
\begin{align}
\frac{1}{\alpha-1}\log\mathbb E_{\theta\sim\pi(\theta|Y')}\big[e^{(\alpha-1)\widetilde c_1(\theta)}\big]=&\frac{1}{2}\log(2\pi\sigma^2)-\frac{1}{2(\alpha-1)}\log\Big(1-\frac{(\alpha-1)v_{Y'}^2}{\sigma^2}\Big)\notag\\
&+\frac{(\mu_{Y'}-y_n)^2}{2(\sigma^2-(\alpha-1)v_{Y'}^2)}.\label{eqn:second_term_reverse}
\end{align}
Plugging Eqs.~\eqref{eqn:c2_reverse} and \eqref{eqn:second_term_reverse} into Eq.~\eqref{eqn:D_alpha_general_c2_ape}, the $\frac{1}{2}\log(2\pi\sigma^2)$ terms cancel, and we obtain
\begin{align}
D_\alpha(\pi(\theta|Y')\|\pi(\theta|Y))=&-\frac{1}{2}\log\Big(1+\frac{v_{Y'}^2}{\sigma^2}\Big)-\frac{1}{2(\alpha-1)}\log\Big(1-\frac{(\alpha-1)v_{Y'}^2}{\sigma^2}\Big)\notag\\
&+\frac{(\mu_{Y'}-y_n)^2}{2(\sigma^2-(\alpha-1)v_{Y'}^2)}-\frac{(\mu_{Y'}-y_n)^2}{2(\sigma^2+v_{Y'}^2)}.\label{eqn:Dalpha_reverse_v}
\end{align}
Recall that $v_{Y'}^2=\big(\frac{n-1}{\sigma^2}+\frac{1}{\tau^2}\big)^{-1}=\frac{\sigma^2}{\omega-1}$ with $\omega=n+\frac{\sigma^2}{\tau^2}$. Then the finiteness condition in Eq.~\eqref{eqn:reverse_finiteness_v} becomes
\begin{align}
\frac{\alpha-1}{\omega-1}<1\quad\Longleftrightarrow\quad1<\alpha<\omega.\label{eqn:reverse_finiteness_omega}
\end{align}
For $1<\alpha<\omega$, Eq.~\eqref{eqn:Dalpha_reverse_v} becomes
\begin{align}
D_\alpha(\pi(\theta|Y')\|\pi(\theta|Y))=&-\frac{1}{2}\log\Big(\frac{\omega}{\omega-1}\Big)-\frac{1}{2(\alpha-1)}\log\Big(\frac{\omega-\alpha}{\omega-1}\Big)\notag\\
&+\frac{(\mu_{Y'}-y_n)^2}{2\sigma^2}\Big[\frac{1}{1-\frac{\alpha-1}{\omega-1}}-\frac{1}{1+\frac{1}{\omega-1}}\Big].\label{eqn:Dalpha_reverse_mup}
\end{align}
The posterior means satisfy
\begin{align}
\omega\mu_{Y}=(\omega-1)\mu_{Y'}+y_n\;\Rightarrow\;\mu_{Y'}-y_n=\frac{\omega}{\omega-1}(\mu_{Y}-y_n).\label{eqn:posterior_mean_reverse_relation}
\end{align}
Therefore, for $1<\alpha<\omega$,
\begin{align}
D_\alpha(\pi(\theta|Y')\|\pi(\theta|Y))&=-\frac{1}{2}\log\Big(\frac{\omega}{\omega-1}\Big)-\frac{1}{2(\alpha-1)}\log\Big(\frac{\omega-\alpha}{\omega-1}\Big)+\frac{(\mu_{Y}-y_n)^2}{2\sigma^2}\frac{\alpha\omega}{(\omega-1)(\omega-\alpha)}\notag\\
&\triangleq\widetilde A(\alpha,\omega)+\frac{(\mu_{Y}-y_n)^2}{2\sigma^2}\widetilde B(\alpha,\omega),\label{eqn:Dalpha_reverse_final}
\end{align}
where
\begin{align}
&\widetilde A(\alpha,\omega)\triangleq-\frac{1}{2}\log\Big(\frac{\omega}{\omega-1}\Big)-\frac{1}{2(\alpha-1)}\log\Big(\frac{\omega-\alpha}{\omega-1}\Big),\notag\\
&\widetilde B(\alpha,\omega)\triangleq\frac{\alpha\omega}{(\omega-1)(\omega-\alpha)}.
\end{align}
For comparison, the corresponding quantities for the forward direction in Eq.~\eqref{eqn:eps_GG_unbounded_ape} are
\begin{align}
&A(\alpha,\omega)=-\frac{1}{2}\log\Big(\frac{\omega-1}{\omega}\Big)-\frac{1}{2(\alpha-1)}\log\Big(1+\frac{\alpha-1}{\omega}\Big),\notag\\
& B(\alpha,\omega)=\frac{\alpha\omega}{(\omega-1)(\omega+\alpha-1)}.\notag
\end{align}
If $|\mu_{Y}-y_n|\leq k^*\sigma$, then for every $1<\alpha<\omega$,
\begin{align}
D_\alpha(\pi(\theta|Y')\|\pi(\theta|Y))\leq\widetilde A(\alpha,\omega)+\frac{k^{*2}}{2}\widetilde B(\alpha,\omega).\label{eqn:Dalpha_reverse_upper}
\end{align}
For $\alpha\geq\omega$, the MGF in Eq.~\eqref{eqn:second_moment_reverse} is not finite, and therefore
\begin{align}
D_\alpha(\pi(\theta|Y')\|\pi(\theta|Y))=\infty,\quad\alpha\geq\omega;\qquad\text{in particular, }D_\infty(\pi(\theta|Y')\|\pi(\theta|Y))=\infty.\label{eqn:Dalpha_reverse_infinite}
\end{align}

\begin{claim}\label{claim:reverse-AB}
For all $1<\alpha<\omega$, $\widetilde A(\alpha,\omega)>0$ and $\widetilde B(\alpha,\omega)>0$. Holding $\alpha$ fixed, both decrease monotonically in $\omega$ on $(\alpha,\infty)$, and $\widetilde A(\alpha,\omega)\to\infty$ as $\omega\downarrow\alpha$.
\end{claim}
\begin{proof}
Let $u=(\omega-1)^{-1}$ and $t=\alpha-1>0$. The condition $1<\alpha<\omega$ is equivalent to $0<tu<1$, and $\omega/(\omega-1)=1+u$, $(\omega-\alpha)/(\omega-1)=1-tu$. Hence $\widetilde A=-\frac12\log(1+u)-\frac{1}{2t}\log(1-tu)$ and $\widetilde B=\frac{\alpha u(1+u)}{1-tu}$. Since $-\log(1-y)>y$ for $y\in(0,1)$ and $\log(1+u)<u$ for $u>0$, we have $-\frac1t\log(1-tu)>u>\log(1+u)$, so $\widetilde A>0$; $\widetilde B>0$ is immediate. Further, $\frac{\partial\widetilde A}{\partial u}=\frac12\big(\frac{1}{1-tu}-\frac{1}{1+u}\big)>0$, and $\widetilde B$ is increasing in $u$ because its numerator increases and its positive denominator decreases in $u$. Since $u$ is decreasing in $\omega$, both $\widetilde A$ and $\widetilde B$ decrease in $\omega$. Finally, $tu\uparrow1$ as $\omega\downarrow\alpha$, so $-\log(1-tu)\to\infty$.
\end{proof}

\paragraph{Two-sided guarantee.}
Combining Eqs.~\eqref{eqn:eps_GG_unbounded_ape} and \eqref{eqn:Dalpha_reverse_upper}, if $|\mu_Y-y_n|\le k^*\sigma$, then for every $1<\alpha<\omega$,
\begin{align}
&\max\{D_\alpha(\pi(\theta|Y)\|\pi(\theta|Y')),D_\alpha(\pi(\theta|Y')\|\pi(\theta|Y))\}\notag\\
\le\;&\max\Big\{A(\alpha,\omega)+\frac{k^{*2}}{2}B(\alpha,\omega),\ \widetilde A(\alpha,\omega)+\frac{k^{*2}}{2}\widetilde B(\alpha,\omega)\Big\},
\end{align}
so OPS satisfies RDP of this order only for $1<\alpha<\omega$, and does not satisfy pure DP under unbounded neighboring. Note that $\widetilde B(\alpha,\omega)>B(\alpha,\omega)$ since $\omega-\alpha<\omega+\alpha-1$. As $\omega=n+\sigma^2/\tau^2$ depends on $n$, a guarantee holding uniformly over all $n\ge1$ is obtained by evaluating at $\omega_{\min}=1+\sigma^2/\tau^2$, and is finite only for $1<\alpha<1+\sigma^2/\tau^2$.
\end{proof}
\section{Proof of Proposition~\ref{eg:LR} for Linear Regression}
\begin{proof}
Let \(\x_i \in \mathbb{R}^d\) be covariates and $y_i \in \mathbb{R}$ be response for the $i$-th individual. Given the dataset \(\Y= \{(\x_i,y_i)\}_{i=1}^n\),  let $\x \in \mathbb{R}^{n\times d}$ be the design matrix, and $\y\in \mathbb{R}^n$ be the response vector. The response follows a Gaussian distribution: $y_i|\x_i,\bs\beta \sim \mathcal N(\x_i^\top\bs\beta,\sigma^2)$. Impose a Gaussian prior on regression coefficient $\bs\beta$, i.e. $\pi(\bs\beta)= \mathcal N(0,\tau^2 \mathbf{I}_d)$, then the posterior is $\bs\beta|\Y = \mathcal N(\bs\mu_\Y,\Sigma_\Y)$, where $\Sigma_\Y=\Big(\frac{1}{\sigma^2}\x^\top \x+\frac{1}{\tau^2}\mathbf{I}_d\Big)^{-1}$, and $\boldsymbol{\mu}_\Y = \Sigma_\Y \left( \frac{1}{\sigma^2} \x^\top \mathbf{y} \right)$.
\subsection{bounded neighboring}\label{ape:LR_bounded}
Let $\Y$ and $\Y'$ differ in exactly one record, i.e. $\z_n=(\x_n,y_n)$ in $\Y$ is replaced by $\z'_n=(\x'_n,y'_n)$ in $\Y'$, then
\begin{align}
c_1(\bs\beta)=\log\frac{\ell(\bs\beta;\Y)}{\ell(\bs\beta;\Y')}
    &=\log p(y_n|\x_n,\bs\beta) - \log p(y'_n|\x'_n,\bs\beta)\notag\\
    & = -\frac{(y_n-\x_n^\top\bs\beta)^2-(y_n'-\x_n'^\top\bs\beta)^2}{2\sigma^2}.
\end{align}
\paragraph{Case 1:} If $\x_n = \x_n'$, then
\begin{align}
    c_1(\bs\beta)
    & = -\frac{(y_n^2-y_n'^2)-2(y_n-y_n')\x_n^\top\bs\beta}{2\sigma^2}.
\end{align}
Since $\bs\beta|\Y = \mathcal N(\bs\mu_\Y,\Sigma_\Y)$ and per the linear transformation of Gaussian random variable, we have $\x_n^{\top}\bs\beta|\Y \sim \mathcal{N}(\x_n^{\top}\bs\mu_\Y, \x_n^{\top}\Sigma_\Y \x_n)$. Therefore, $c_1(\bs\beta)$ under $P = \pi(\cdot|\Y)$ is Gaussian with variance
\begin{align}
    \sigma_{\Y, \Y'}^2 = \mbox{Var}_{\bs\beta\sim\pi(\cdot|\Y)}[c_1(\bs\beta)]= \frac{(y_n-y_n')^2}{\sigma^4}\,\x_n^\top\Sigma_\Y\x_n.
\end{align}
Since $\x^{\top}\x$ is positive semi-definite, $\Sigma_\Y=\Big(\frac{1}{\sigma^2}\x^\top \x+\frac{1}{\tau^2}\mathbf{I}_d\Big)^{-1}\preceq
\left(\frac{1}{\tau^2} \mathbf{I}_d\right)^{-1} = \tau^2 \mathbf{I}_d$. If additionally assume $|y_n-y_n'|\le 2k\sigma$, then
\begin{align}
\sigma_{\Y, \Y'}^2 &= \frac{(y_n-y_n')^2}{\sigma^4}\,\x_n^\top\Sigma_\Y\x_n \notag\\
&\le\frac{4k^2\sigma^2}{\sigma^4}\x_n^\top(\tau^2\mathbf{I}_d)\x_n = \frac{4k^2}{\sigma^2}\tau^2\|\x_n\|^2 \le \frac{4k^2\tau^2R^2}{\sigma^2} \triangleq \bar{\sigma}^2.
\end{align}
For the reversed pair $(\Y',\Y)$, $\tilde c_1(\bs\beta)=-c_1(\bs\beta)$ is Gaussian under $\pi(\bs\beta|\Y')$ with variance $(y_n-y'_n)^2\x_n^\top\Sigma_{\Y'}\x_n/\sigma^4$; since $\Sigma_{\Y'}\preceq\tau^2\mathbf I_d$, the same bound holds for $\sigma^2_{\Y',\Y}$.
Per Corollary~\ref{coro:subgaussian},
\begin{align}
\max\{D_\alpha(P\|Q),D_\alpha(Q\|P)\}\le \frac{\alpha\bar\sigma^2}{2}\le\frac{2\alpha k^2\tau^2 R^2}{\sigma^2}.
\end{align}

\paragraph{Case 2: } 
When $\x_n \neq \x_n'$, the log-likelihood ratio decomposes as 
$c_1(\bs\beta) = L(\bs\beta) + Q(\bs\beta)$, where
\begin{align}
L(\bs\beta) &= -\frac{(y_n^2-y_n'^2) - 
2(y_n\x_n-y_n'\x_n')^\top\bs\beta}{2\sigma^2}, \\
Q(\bs\beta) &= -\frac{\bs\beta^\top A\bs\beta}{2\sigma^2}, \quad 
A \triangleq \x_n\x_n^\top - \x_n'\x_n'^\top.
\end{align}
Since $A$ may be indefinite, $Q(\bs\beta)$ is a linear combination 
of non-central $\chi^2$ random variables with potentially negative 
coefficients under $\bs\beta|\Y \sim \mathcal{N}(\bs\mu_\Y, 
\Sigma_\Y)$, so $c_1(\bs\beta)$ is generally not sub-Gaussian. 
We show that it satisfies a sub-exponential MGF bound under 
additional boundedness assumptions, by bounding the MGF of each 
part separately and combining via Cauchy-Schwarz. Let $\mu_L = 
\mathbb{E}[L(\bs\beta)]$, $\mu_Q = \mathbb{E}[Q(\bs\beta)]$, 
and $\mu^{c_1}_\Y = \mu_L + \mu_Q$\footnote{For brevity, we write $\mu^{c_1}_\Y,\nu^2_\Y,b_\Y$ for $\mu_{\Y,\Y'},\nu^2_{\Y,\Y'},b_{\Y,\Y'}$ in Corollary~\ref{coro:subexponential}.}.

\textbf{Step 1: MGF of the linear part.}
Since $L(\bs\beta)$ is affine in $\bs\beta$ and $\bs\beta|\Y 
\sim \mathcal{N}(\bs\mu_\Y,\Sigma_\Y)$, $L(\bs\beta)$ is 
Gaussian under the posterior. Using $\Sigma_\Y \preceq 
\tau^2\mathbf{I}_d$, $|y_i| \leq M_y$, and $\|\x_i\| \leq R$:
\begin{align}
\mathrm{Var}[L(\bs\beta)] 
= \frac{(y_n\x_n-y_n'\x_n')^\top\Sigma_\Y
(y_n\x_n-y_n'\x_n')}{\sigma^4} 
&\leq \frac{\|y_n\x_n-y_n'\x_n'\|^2\tau^2}{\sigma^4} \notag\\
&\leq \frac{4M_y^2R^2\tau^2}{\sigma^4} 
\triangleq \bar{\sigma}^2_L.
\end{align}
Therefore $L(\bs\beta)$ is sub-Gaussian with variance proxy 
$\bar{\sigma}^2_L$ and for all $t \in \mathbb{R}$:
\begin{align}
\log\mathbb{E}\left[e^{t(L(\bs\beta)-\mu_L)}\right] 
\leq \frac{t^2\bar{\sigma}^2_L}{2}.
\label{eqn:mgf_L_case2}
\end{align}

\textbf{Step 2: Centered MGF of the quadratic part.}
We compute $\mu_Q$ exactly. For $\bs\beta \sim 
\mathcal{N}(\bs\mu_\Y,\Sigma_\Y)$:
\begin{align}
\mu_Q = -\frac{1}{2\sigma^2}\mathbb{E}[\bs\beta^\top A\bs\beta] 
= -\frac{1}{2\sigma^2}\left(\mathrm{tr}(A\Sigma_\Y) + 
\bs\mu_\Y^\top A\bs\mu_\Y\right).
\label{eqn:mu_Q_case2}
\end{align}
Define the symmetrized matrix $\tilde{A} \triangleq 
\Sigma_\Y^{1/2}A\Sigma_\Y^{1/2}$. By Sylvester's determinant 
identity:
\begin{align}
\det\!\left(\mathbf{I}+\frac{t}{\sigma^2}\Sigma_\Y A\right) 
= \det\!\left(\mathbf{I}+\frac{t}{\sigma^2}
\Sigma_\Y^{1/2}A\Sigma_\Y^{1/2}\right) 
= \det\!\left(\mathbf{I}+\frac{t}{\sigma^2}\tilde{A}\right),
\end{align}
so the positivity condition $\mathbf{I}+\frac{t}{\sigma^2}
\tilde{A} \succ 0$ controls both the determinant and the 
existence of the MGF, while the inverse term is retained in 
the nonsymmetric form $(\mathbf{I}+\frac{t}{\sigma^2}
\Sigma_\Y A)^{-1}$, which is algebraically equivalent. 
Since $\tilde{A}$ is symmetric:
\begin{align}
\|\tilde{A}\|_{\mathrm{op}} = 
\|\Sigma_\Y^{1/2}A\Sigma_\Y^{1/2}\|_{\mathrm{op}} 
\leq \|\Sigma_\Y\|_{\mathrm{op}}\|A\|_{\mathrm{op}} 
\leq \tau^2(\|\x_n\|^2+\|\x_n'\|^2) \leq 2\tau^2R^2,
\end{align}
so the positivity condition holds for $|t| < b_Q^{-1}$ where 
$b_Q \triangleq 2\tau^2R^2/\sigma^2$. The raw MGF of 
$Q(\bs\beta)$ is:
\begin{align}
\mathbb{E}\left[e^{tQ(\bs\beta)}\right] 
= \frac{\exp\!\left(-\dfrac{t}{2\sigma^2}\bs\mu_\Y^\top A
\left(\mathbf{I}+\dfrac{t}{\sigma^2}\Sigma_\Y A\right)^{-1}
\bs\mu_\Y\right)}
{\sqrt{\det\!\left(\mathbf{I}+\dfrac{t}{\sigma^2}
\tilde{A}\right)}}.
\label{eqn:mgf_Q_raw_case2}
\end{align}
Taking the logarithm of Eq.~\eqref{eqn:mgf_Q_raw_case2} and 
centering by $\mu_Q$ via Eq.~\eqref{eqn:mu_Q_case2}:
\begin{align}
\log\mathbb{E}\left[e^{t(Q(\bs\beta)-\mu_Q)}\right] 
&= -\frac{1}{2}\log\det\!\left(\mathbf{I}+
\frac{t}{\sigma^2}\tilde{A}\right) - \frac{t}{2\sigma^2}\bs\mu_\Y^\top A
\left(\mathbf{I}+\frac{t}{\sigma^2}\Sigma_\Y A\right)^{-1}
\bs\mu_\Y \notag\\
&\;\;\;\;+ \frac{t}{2\sigma^2}\left(\mathrm{tr}(A\Sigma_\Y) + 
\bs\mu_\Y^\top A\bs\mu_\Y\right).
\label{eqn:centered_mgf_Q_case2}
\end{align}
We bound each term in 
Eq.~\eqref{eqn:centered_mgf_Q_case2} separately.

\textit{Term 1 (log-determinant).} 
Applying the scalar inequality $-\log(1+x) \leq -x + 
x^2/(1-|x|)$ (valid for $|x|<1$) eigenvalue-by-eigenvalue 
to $\tilde{A}$, and using $\|t\tilde{A}/\sigma^2\|_{\mathrm{op}} 
\leq b_Q|t| < 1$:
\begin{align}
-\frac{1}{2}\log\det\!\left(\mathbf{I}+\frac{t}{\sigma^2}
\tilde{A}\right) 
\leq -\frac{t}{2\sigma^2}\mathrm{tr}(\tilde{A}) + 
\frac{t^2\,\mathrm{tr}(\tilde{A}^2)}{2\sigma^4(1-b_Q|t|)},
\label{eqn:logdet_bound_case2}
\end{align}
where $\mathrm{tr}(\tilde{A}) = \mathrm{tr}(\Sigma_\Y A)$.

\textit{Term 2 (quadratic numerator).}
Since $(\mathbf{I}+\frac{t}{\sigma^2}\Sigma_\Y A)^{-1}$ is 
not symmetric in general, we use operator norms directly. 
Note first that
\begin{align}
\left\|\frac{t}{\sigma^2}\Sigma_\Y A\right\|_{\mathrm{op}} 
\leq \frac{|t|}{\sigma^2}\|\Sigma_\Y\|_{\mathrm{op}}
\|A\|_{\mathrm{op}} \leq b_Q|t| < 1,
\label{eqn:op_norm_SigmaA}
\end{align}
so the resolvent bound gives
\begin{align}
\left\|\left(\mathbf{I}+\frac{t}{\sigma^2}\Sigma_\Y A
\right)^{-1}\right\|_{\mathrm{op}} \leq 
\frac{1}{1-b_Q|t|}.
\label{eqn:resolvent_bound}
\end{align}
Writing $(\mathbf{I}+\frac{t}{\sigma^2}\Sigma_\Y A)^{-1} = 
\mathbf{I} - \frac{t}{\sigma^2}\Sigma_\Y A
(\mathbf{I}+\frac{t}{\sigma^2}\Sigma_\Y A)^{-1}$ and 
applying Eq.~\eqref{eqn:resolvent_bound}:
\begin{align}
&-\frac{t}{2\sigma^2}\bs\mu_\Y^\top A
\left(\mathbf{I}+\frac{t}{\sigma^2}\Sigma_\Y A\right)^{-1}
\bs\mu_\Y \notag\\
=\;& -\frac{t}{2\sigma^2}\bs\mu_\Y^\top A\bs\mu_\Y 
+ \frac{t^2}{2\sigma^4}\bs\mu_\Y^\top A\Sigma_\Y A
\left(\mathbf{I}+\frac{t}{\sigma^2}\Sigma_\Y A\right)^{-1}
\bs\mu_\Y \notag\\
\leq\;& -\frac{t}{2\sigma^2}\bs\mu_\Y^\top A\bs\mu_\Y + 
\frac{t^2\|\bs\mu_\Y\|^2\|A\|_{\mathrm{op}}^2
\|\Sigma_\Y\|_{\mathrm{op}}}
{2\sigma^4(1-b_Q|t|)},
\label{eqn:numerator_bound_case2}
\end{align}
where the last step uses $\left|\bs\mu_\Y^\top A\Sigma_\Y A
(\mathbf{I}+\frac{t}{\sigma^2}\Sigma_\Y A)^{-1}
\bs\mu_\Y\right| \leq \|\bs\mu_\Y\|^2\|A\|_{\mathrm{op}}^2
\|\Sigma_\Y\|_{\mathrm{op}}/(1-b_Q|t|)$ via 
Eq.~\eqref{eqn:resolvent_bound}.

\textit{Term 3 (centering).}
\begin{align}
\frac{t}{2\sigma^2}\left(\mathrm{tr}(A\Sigma_\Y) + 
\bs\mu_\Y^\top A\bs\mu_\Y\right) 
= \frac{t}{2\sigma^2}\mathrm{tr}(\tilde{A}) + 
\frac{t}{2\sigma^2}\bs\mu_\Y^\top A\bs\mu_\Y.
\label{eqn:centering_case2}
\end{align}

Combining Terms 1--3, the linear terms 
$-\frac{t}{2\sigma^2}\mathrm{tr}(\tilde{A})$ and 
$-\frac{t}{2\sigma^2}\bs\mu_\Y^\top A\bs\mu_\Y$ cancel 
exactly with Eq.~\eqref{eqn:centering_case2}, giving:
\begin{align}
\log\mathbb{E}\left[e^{t(Q(\bs\beta)-\mu_Q)}\right] 
\leq \frac{t^2}{2(1-b_Q|t|)}\left(
\frac{\mathrm{tr}(\tilde{A}^2)}{\sigma^4} + 
\frac{\|\bs\mu_\Y\|^2\|A\|_{\mathrm{op}}^2
\|\Sigma_\Y\|_{\mathrm{op}}}{\sigma^4}
\right) 
\triangleq \frac{t^2\nu^2_Q}{2(1-b_Q|t|)}.
\label{eqn:mgf_Q_bound_case2}
\end{align}
Bounding each term using $\Sigma_\Y \preceq \tau^2\mathbf{I}_d$, 
$\|A\|_F^2 \leq 8R^4$, $\|A\|_{\mathrm{op}} \leq 2R^2$, and 
$\|\bs\mu_\Y\| \leq n\tau^2M_yR/\sigma^2$:
\begin{align}
\mathrm{tr}(\tilde{A}^2) = \mathrm{tr}(\Sigma_\Y A\Sigma_\Y A) 
&\leq \tau^4\|A\|_F^2 \leq 8\tau^4R^4, 
\label{eqn:tr_final_case2}\\
\|\bs\mu_\Y\|^2\|A\|_{\mathrm{op}}^2
\|\Sigma_\Y\|_{\mathrm{op}} 
&\leq \frac{n^2\tau^4M_y^2R^2}{\sigma^4}
\cdot 4R^4 \cdot \tau^2 
= \frac{4n^2\tau^6M_y^2R^6}{\sigma^4},
\label{eqn:quad_final_case2}
\end{align}
so that
\begin{align}
\nu^2_Q \leq \frac{8\tau^4R^4}{\sigma^4} + 
\frac{4n^2\tau^6M_y^2R^6}{\sigma^8}.
\label{eqn:nu_Q_case2}
\end{align}

\textbf{Step 3: Combining via Cauchy-Schwarz.}
Since $L(\bs\beta)$ and $Q(\bs\beta)$ both depend on $\bs\beta$, 
they are not independent. Applying Cauchy-Schwarz:
\begin{align}
\log\mathbb{E}\!\left[e^{t(c_1(\bs\beta)-\mu^{c_1}_\Y)}
\right] 
&\leq \frac{1}{2}\log\mathbb{E}\!\left[e^{2t(L-\mu_L)}
\right] + \frac{1}{2}\log\mathbb{E}\!\left[e^{2t(Q-\mu_Q)}
\right] \notag\\
&\leq \frac{(2t)^2\bar{\sigma}^2_L}{4} + 
\frac{(2t)^2\nu^2_Q}{4(1-b_Q|2t|)} \notag\\
&= t^2\bar{\sigma}^2_L + \frac{t^2\nu^2_Q}{1-2b_Q|t|} 
\notag\\
&\leq \frac{t^2(\bar{\sigma}^2_L+\nu^2_Q)}{1-2b_Q|t|} 
\triangleq \frac{t^2\nu^2_\Y/2}{1-b_\Y|t|},
\label{eqn:combined_case2}
\end{align}
where
\begin{align}
\nu^2_\Y &\triangleq 2(\bar{\sigma}^2_L+\nu^2_Q) \leq 
\frac{8M_y^2R^2\tau^2}{\sigma^4} + 
\frac{16\tau^4R^4}{\sigma^4} + 
\frac{8n^2\tau^6M_y^2R^6}{\sigma^8},
\label{eqn:nu_Y_case2}\\
b_\Y &\triangleq 2b_Q = \frac{4\tau^2R^2}{\sigma^2}.
\label{eqn:b_Y_case2}
\end{align}
This establishes that $c_1(\bs\beta)$ satisfies a 
sub-exponential MGF bound under $\pi(\bs\beta|\Y)$ with 
parameters $(\nu^2_\Y, b_\Y)$ in the sense of 
Eq.~\eqref{eqn:subexp_con}.

For the reversed pair $(\Y',\Y)$, $\tilde c_1(\bs\beta)=-c_1(\bs\beta)=-L(\bs\beta)-Q(\bs\beta)$, i.e., $(L,Q,A)$ is replaced by $(-L,-Q,-A)$. Steps~1--3 use only $\Sigma\preceq\tau^2\mathbf I_d$, $\|\x_i\|\le R$, $|y_i|\le M_y$, $\|\bs\mu\|\le n\tau^2M_yR/\sigma^2$, $\|{-A}\|_{\mathrm{op}}=\|A\|_{\mathrm{op}}$ and $\|{-A}\|_F=\|A\|_F$, all of which also hold under $\pi(\bs\beta|\Y')$ since $\Y'$ also has $n$ records. Hence $\tilde c_1(\bs\beta)$ satisfies the same sub-exponential bound with parameters $(\nu^2_\Y,b_\Y)$ in Eqs.~\eqref{eqn:nu_Y_case2}--\eqref{eqn:b_Y_case2}.

\textbf{Step 4: Applying Corollary~\ref{coro:subexponential}.}
Corollary~\ref{coro:subexponential} requires $b_\Y < 
\min\{(\alpha-1)^{-1},1\}$. The condition $b_\Y < 1$ is an 
additional scale restriction requiring:
\begin{align}
\frac{4\tau^2R^2}{\sigma^2} < 1 
\iff \sigma^2 > 4\tau^2R^2.
\label{eqn:scale_condition}
\end{align}
Under Eq.~\eqref{eqn:scale_condition} and $\alpha < 1 + 
\sigma^2/(4\tau^2R^2)$, substituting $(\nu^2_\Y, b_\Y)$ into 
Corollary~\ref{coro:subexponential}:
\begin{align}
\max\{D_\alpha(P\|Q),D_\alpha(Q\|P)\} \leq \varepsilon_{\text{LR, Case2}}
(\alpha, b_\Y, \nu^2_\Y) \triangleq \frac{\nu^2_\Y}{2}\cdot
\frac{\alpha(1-2b_\Y)+2b_\Y}
{(1-b_\Y)\left(1-b_\Y(\alpha-1)\right)}.
\label{eqn:eps_LR_case2}
\end{align}

\begin{remark}
Three sources of looseness affect Eq.~\eqref{eqn:eps_LR_case2}. 
First, the Cauchy-Schwarz step doubles $b_Q$, restricting the 
valid range of $\alpha$. Second, the operator norm bound 
$\|\tilde{A}\|_{\mathrm{op}} \leq 2\tau^2R^2$ ignores 
data-dependent tightening of $\Sigma_\Y$ due to the likelihood. 
Third, the $n^2$-dependent term in $\nu^2_\Y$ arises from 
bounding $\|\bs\mu_\Y\|$ using only row-wise boundedness; under 
a direct bound $\|\bs\mu_\Y\| \leq C$ for some constant $C$, 
this term becomes $n$-independent. Despite this looseness, 
Eq.~\eqref{eqn:eps_LR_case2} yields an explicit finite-order 
RDP bound for OPS in the bounded-neighboring linear-regression 
setting with covariate replacement, and illustrates how the 
sub-exponential framework of Corollary~\ref{coro:subexponential} extends 
beyond the sub-Gaussian cases treated in the main text.
\end{remark}

\subsection{unbounded neighboring}
Let $\Y$ has size $n$ and $\Y'$ is of size $n-1$ by removing the $n$-th record $\z_n=(\x_n,y_n)$ in $\Y$. Then
\begin{align}
    c_1(\bs\beta)=\log\frac{\ell(\bs\beta;\Y)}{\ell(\bs\beta;\Y')}
    &=\log p(y_n|\x_n,\bs\beta)= -\frac{(y_n - \mathbf{x}_n^\top\boldsymbol{\beta})^2}{2\sigma^2} - \frac{1}{2}\log(2\pi\sigma^2).\notag
\end{align}
This expression is structurally identical to Eq.~\eqref{eqn:GG_unbounded_c1} by setting $Z = y_n - \mathbf{x}_n^\top\boldsymbol{\beta}$, where 
\begin{align}
    Z|\Y \sim \mathcal{N}\left(y_n - \mathbf{x}_n^\top\boldsymbol{\mu}_\mathbf{Y}, \mathbf{x}_n^\top \Sigma_\mathbf{Y} \mathbf{x}_n\right).\notag
\end{align}
In Proposition~\ref{thm:GG}, $\omega=\sigma^2/v_Y^2$, where $v_Y^2$ is the posterior variance of $\theta-y_n$; here the corresponding quantity is the posterior variance of $Z|\Y$. Thus, the corresponding $\omega_n$ here is $\omega_n =\frac{\sigma^2}{\mathbf{x}_n^\top \Sigma_\mathbf{Y} \mathbf{x}_n}$, plug in Eq.~\eqref{eqn:eps_GG_unbounded_ape}
\begin{align}
    D_\alpha(P\|Q) &= \underbrace{-\frac{\log(1-\omega_n^{-1})}{2} - \frac{\log(1+\frac{\alpha-1}{\omega_n})}{2(\alpha-1)}}_{A(\alpha, \omega_n)} + \frac{(\mathbf{x}_n^\top \boldsymbol{\mu}_\mathbf{Y} - y_n)^2}{2\sigma^2} \underbrace{\left[ \frac{1}{1-\omega_n^{-1}} - \frac{1}{1+\frac{\alpha-1}{\omega_n}} \right]}_{B(\alpha, \omega_n)}\notag\\
    & \le A(\alpha, \omega_n) + \frac{k^{*2}}{2}B(\alpha, \omega_n),
\end{align}
if $|\mathbf{x}_n^\top \boldsymbol{\mu}_\mathbf{Y} - y_n|\le k^*\sigma$ for some $k^*>0$.

\paragraph{Reverse direction.}
We now bound $D_\alpha(\pi(\bs\beta|\Y')\|\pi(\bs\beta|\Y))$. The posterior under $\Y'$ is $\bs\beta|\Y'\sim\mathcal N(\bs\mu_{\Y'},\Sigma_{\Y'})$ with $\Sigma_{\Y'}^{-1}=\Sigma_\Y^{-1}-\sigma^{-2}\x_n\x_n^\top$. Let $v_n'^2\triangleq\x_n^\top\Sigma_{\Y'}\x_n$ and
\begin{align}
&\tilde c_1(\bs\beta)=\log\frac{\ell(\bs\beta;\Y')}{\ell(\bs\beta;\Y)}=\frac{(y_n-\x_n^\top\bs\beta)^2}{2\sigma^2}+\frac12\log(2\pi\sigma^2),\notag\\
&Z'\triangleq y_n-\x_n^\top\bs\beta\,|\,\Y'\sim\mathcal N(y_n-\x_n^\top\bs\mu_{\Y'},\,v_n'^2).
\end{align}
This is structurally identical to Eq.~\eqref{eqn:c1_GG_reverse} with $\mu_{Y'}$ replaced by $\x_n^\top\bs\mu_{\Y'}$ and $v_{Y'}^2$ replaced by $v_n'^2$. Hence, by Eq.~\eqref{eqn:Dalpha_reverse_v}, $D_\alpha(Q\|P)<\infty$ if and only if $(\alpha-1)v_n'^2<\sigma^2$. By the Sherman--Morrison formula,
\begin{align*}
&\Sigma_\Y=\Sigma_{\Y'}-\frac{\Sigma_{\Y'}\x_n\x_n^\top\Sigma_{\Y'}}{\sigma^2+v_n'^2}\;\Rightarrow\;\x_n^\top\Sigma_\Y\x_n=\frac{\sigma^2v_n'^2}{\sigma^2+v_n'^2}\;\Rightarrow\;\omega_n=1+\frac{\sigma^2}{v_n'^2},\\ 
&\text{ i.e. }\ \frac{v_n'^2}{\sigma^2}=\frac{1}{\omega_n-1}.
\end{align*}
Moreover, $\bs\mu_\Y=\Sigma_\Y(\Sigma_{\Y'}^{-1}\bs\mu_{\Y'}+\sigma^{-2}\x_ny_n)$ and $\x_n^\top\Sigma_\Y\Sigma_{\Y'}^{-1}=\frac{\sigma^2}{\sigma^2+v_n'^2}\x_n^\top$, so
\begin{align}
\x_n^\top\bs\mu_\Y=\frac{\sigma^2}{\sigma^2+v_n'^2}\x_n^\top\bs\mu_{\Y'}+\frac{v_n'^2}{\sigma^2+v_n'^2}y_n\;\Rightarrow\;\x_n^\top\bs\mu_{\Y'}-y_n=\frac{\omega_n}{\omega_n-1}(\x_n^\top\bs\mu_\Y-y_n).
\end{align}
These coincide with $v_{Y'}^2=\sigma^2/(\omega-1)$ and Eq.~\eqref{eqn:posterior_mean_reverse_relation} with $\omega$ replaced by $\omega_n$. Repeating the derivation of Eq.~\eqref{eqn:Dalpha_reverse_final}, when $1<\alpha<\omega_n$
\begin{align}
D_\alpha(Q\|P)=\widetilde A(\alpha,\omega_n)+\frac{(\x_n^\top\bs\mu_\Y-y_n)^2}{2\sigma^2}\widetilde B(\alpha,\omega_n)\le\widetilde A(\alpha,\omega_n)+\frac{k^{*2}}{2}\widetilde B(\alpha,\omega_n),
\end{align}
and $D_\alpha(Q\|P)=\infty$ for $\alpha\ge\omega_n$.

\paragraph{Two-sided, data-independent guarantee.}
Since $\Sigma_{\Y'}\preceq\tau^2\mathbf I_d$ and $\|\x_n\|\le R$, $v_n'^2\le\tau^2R^2$, hence $\omega_n\ge\omega_{\min}\triangleq1+\sigma^2/(\tau^2R^2)$ for every admissible neighboring pair, with equality, e.g., when $n=1$ and $\|\x_1\|=R$. By the monotonicity of $A,B$ and $\widetilde A,\widetilde B$ in $\omega$ (Claim~\ref{claim:reverse-AB} and ~\ref{claim:mono}), for every $1<\alpha<\omega_{\min}$,
\begin{align}
&\max\{D_\alpha(P\|Q),D_\alpha(Q\|P)\}\notag\\
\le\;&\max\Big\{A(\alpha,\omega_{\min})+\frac{k^{*2}}{2}B(\alpha,\omega_{\min}),\ \widetilde A(\alpha,\omega_{\min})+\frac{k^{*2}}{2}\widetilde B(\alpha,\omega_{\min})\Big\},
\end{align}
whereas for $\alpha\ge\omega_{\min}$ there exist admissible neighboring pairs with $D_\alpha(Q\|P)=\infty$. In particular, OPS does not satisfy pure DP in this setting.

\end{proof}

\section*{NeurIPS Paper Checklist}

\begin{enumerate}

\item {\bf Claims}
    \item[] Question: Do the main claims made in the abstract and introduction accurately reflect the paper's contributions and scope?
    \item[] Answer: \answerYes{} 
    \item[] Justification: The abstract and introduction state the paper’s theoretical contributions

\item {\bf Limitations}
    \item[] Question: Does the paper discuss the limitations of the work performed by the authors?
    \item[] Answer: \answerYes{} 
    \item[] Justification: The paper discusses limitations in Section~\ref{sec:discuss}, including the dependence of the guarantees on model assumptions, tail conditions on the log-likelihood ratio, and the looseness of general sub-exponential bounds relative to model-specific exact calculations.

\item {\bf Theory assumptions and proofs}
    \item[] Question: For each theoretical result, does the paper provide the full set of assumptions and a complete (and correct) proof?
    \item[] Answer: \answerYes{} 
    \item[] Justification: The paper is theoretical. Each theoretical result states its assumptions, and proofs are provided either in the main text or in the appendix.

    \item {\bf Experimental result reproducibility}
    \item[] Question: Does the paper fully disclose all the information needed to reproduce the main experimental results of the paper to the extent that it affects the main claims and/or conclusions of the paper (regardless of whether the code and data are provided or not)?
    \item[] Answer: \answerNA{} 
    \item[] Justification: The paper does not report empirical experiments. The main contributions are theoretical.

\item {\bf Open access to data and code}
    \item[] Question: Does the paper provide open access to the data and code, with sufficient instructions to faithfully reproduce the main experimental results, as described in supplemental material?
    \item[] Answer: \answerNA{} 
    \item[] Justification: The paper does not use datasets or code to support experimental claims. The work is theoretical.

\item {\bf Experimental setting/details}
    \item[] Question: Does the paper specify all the training and test details (e.g., data splits, hyperparameters, how they were chosen, type of optimizer) necessary to understand the results?
    \item[] Answer: \answerNA{} 
    \item[] Justification: The paper does not include experiments. The work is theoretical.

\item {\bf Experiment statistical significance}
    \item[] Question: Does the paper report error bars suitably and correctly defined or other appropriate information about the statistical significance of the experiments?
    \item[] Answer: \answerNA{} 
    \item[] Justification: The paper does not include experiments. The work is theoretical.

\item {\bf Experiments compute resources}
    \item[] Question: For each experiment, does the paper provide sufficient information on the computer resources (type of compute workers, memory, time of execution) needed to reproduce the experiments?
    \item[] Answer: \answerNA{} 
    \item[] Justification: The paper does not include experiments. The work is theoretical.
    
\item {\bf Code of ethics}
    \item[] Question: Does the research conducted in the paper conform, in every respect, with the NeurIPS Code of Ethics \url{https://neurips.cc/public/EthicsGuidelines}?
    \item[] Answer: \answerYes{} 
    \item[] Justification: The research is theoretical and concerns privacy guarantees for Bayesian posterior sampling.

\item {\bf Broader impacts}
    \item[] Question: Does the paper discuss both potential positive societal impacts and negative societal impacts of the work performed?
    \item[] Answer: \answerYes{} 
    \item[] Justification: Potential positive impacts include improved understanding of privacy-preserving Bayesian posterior sampling. Such understanding can help design mechanisms that preserve more utility in released statistical outputs while exploiting the inherent privacy provided by posterior randomness.  potential limitations include the risk of generalizing the guarantees outside the stated assumptions.
    
\item {\bf Safeguards}
    \item[] Question: Does the paper describe safeguards that have been put in place for responsible release of data or models that have a high risk for misuse (e.g., pre-trained language models, image generators, or scraped datasets)?
    \item[] Answer: \answerNA{} 
    \item[] Justification: The paper does not release models, datasets, or tools with high misuse risk.

\item {\bf Licenses for existing assets}
    \item[] Question: Are the creators or original owners of assets (e.g., code, data, models), used in the paper, properly credited and are the license and terms of use explicitly mentioned and properly respected?
    \item[] Answer: \answerNA{} 
    \item[] Justification: The paper does not use existing datasets, code, models, or other external assets.Related Prior work is listed in the References Section.

\item {\bf New assets}
    \item[] Question: Are new assets introduced in the paper well documented and is the documentation provided alongside the assets?
    \item[] Answer: \answerNA{} 
    \item[] Justification: The paper does not introduce or release new datasets, codebases, models, benchmarks, or other assets.

\item {\bf Crowdsourcing and research with human subjects}
    \item[] Question: For crowdsourcing experiments and research with human subjects, does the paper include the full text of instructions given to participants and screenshots, if applicable, as well as details about compensation (if any)? 
    \item[] Answer: \answerNA{} 
    \item[] Justification: The paper does not involve crowdsourcing.

\item {\bf Institutional review board (IRB) approvals or equivalent for research with human subjects}
    \item[] Question: Does the paper describe potential risks incurred by study participants, whether such risks were disclosed to the subjects, and whether Institutional Review Board (IRB) approvals (or an equivalent approval/review based on the requirements of your country or institution) were obtained?
    \item[] Answer: \answerNA{} 
    \item[] Justification: The paper does not involve human-subject research.

\item {\bf Declaration of LLM usage}
    \item[] Question: Does the paper describe the usage of LLMs if it is an important, original, or non-standard component of the core methods in this research? Note that if the LLM is used only for writing, editing, or formatting purposes and does \emph{not} impact the core methodology, scientific rigor, or originality of the research, declaration is not required.
    \item[] Answer: \answerNA{} 
    \item[] Justification: The core method development does not involve LLMs as an important, original, or non-standard component.

\end{enumerate}

\end{document}